\documentclass{article}

     \PassOptionsToPackage{numbers, compress}{natbib}

\usepackage{temp}

\usepackage{graphicx}
\usepackage{amsmath}
\usepackage{amssymb}
\usepackage{multirow}
\usepackage{threeparttable}
\usepackage[belowskip=-10cm,aboveskip=-10pt]{caption}
\usepackage[belowskip=-0cm,aboveskip=0pt]{subcaption}

\usepackage[utf8]{inputenc} 
\usepackage[T1]{fontenc}    
\usepackage{hyperref}       
\usepackage{url}            
\usepackage{booktabs}       
\usepackage{amsfonts}       
\usepackage{nicefrac}       
\usepackage{microtype}      
\usepackage{xcolor}         
\usepackage{floatrow}
\newfloatcommand{capbtabbox}{table}[][\FBwidth]
\usepackage[capitalize]{cleveref}
\usepackage{wrapfig}

\usepackage{lipsum}
\usepackage{listings}
\definecolor{light-gray}{gray}{0.80}

\usepackage{blindtext}

\title{Studying inductive biases in image classification task}

\author{%
  Nana Arizumi\\
  DENSO CORPORATION\\
}

\begin{document}

\maketitle

\begin{abstract}
Recently, self-attention (SA) structures became popular in computer vision fields.  They have locally independent filters and can use large kernels, which contradicts the previously popular convolutional neural networks (CNNs).  CNNs success was attributed to the hard-coded inductive biases of locality and spatial invariance.  However, recent studies have shown that inductive biases in CNNs are too restrictive.  On the other hand, the relative position encodings, similar to depthwise (DW) convolution, are necessary for the local SA networks, which indicates that the SA structures are not entirely spatially variant.  Hence, we would like to determine which part of inductive biases contributes to the success of the local SA structures.  To do so, we introduced context-aware decomposed attention (CADA), which decomposes attention maps into multiple trainable base kernels and accumulates them using context-aware (CA) parameters.  This way, we could identify the link between the CNNs and SA networks.  We conducted ablation studies using the ResNet50 applied to the ImageNet classification task.  DW convolution could have a large locality without increasing computational costs compared to CNNs, but the accuracy saturates with larger kernels.  CADA follows this characteristic of locality.  We showed that context awareness was the crucial property; however, large local information was not necessary to construct CA parameters.  Even though no spatial invariance makes training difficult, more relaxed spatial invariance gave better accuracy than strict spatial invariance.  Also, additional strong spatial invariance through relative position encoding was preferable.   We extended these experiments to filters for downsampling and showed that locality bias is more critical for downsampling but can remove the strong locality bias using relaxed spatial invariance.

\end{abstract}

\section{Introduction}
Global self-attention (SA) structures are becoming popular among computer vision tasks \cite{dosovitskiy2020image, touvron2020training}, which have context-aware (CA) locally independent filters and can handle long-range dependencies. They are extended to non-CA MLP-based networks \cite{tolstikhin2021mlp, touvron2021resmlp}, which are trained to induce locality and spatial invariance in the earlier layers. The inductive biases of SA and MLP-based networks are much more relaxed than the previously popular convolutional neural networks (CNNs). Hard-coded inductive biases of spatial invariance and locality are believed to be essential for the remarkable success of CNNs. However, several studies \cite{elsayed2020revisiting, d2019finding} have shown that relaxing the inductive biases of CNNs can improve accuracy. Both the locally-connected and fully-connected networks without any spatial invariance are unable to train as the CNNs \cite{neyshabur2020towards, novak2018bayesian}, although they include convolutions in their parameter space. Hence, we would like to find out which part of inductive biases contributes to the success of the SA networks.

There are several reasons for the difficulty in comparing the CNNs and SA networks reported in the literature. First, many studies compare different networks using the same computation budgets or network sizes, making the comparison between components impossible. In particular, if the backbones are different, the effect of each component differs significantly. Especially the global SA networks use patches and lose local information. Because of computational complexity, the pyramid backbones of ResNets without patches are not feasible for the global SA networks, which further complicates the comparison.  Second, the recent advancement in training schemes and regularizations \cite{touvron2020training, touvron2021resmlp, chen2021vision} makes it difficult to compare the results of various reports in the literature. Even the traditional CNNs are improving using modern methods \cite{he2019bag, bello2021revisiting, wightman2021resnet}. Third, the low computational cost of SA networks is due to the multihead (MH) aggregation operation, which is the same operation as the MH depthwise (DW) computation; hence, comparing SA networks with convolution does not give a fair comparison. The SA networks should be compared with the DW convolutional networks, which have more relaxed local inductive biases than CNNs. We conducted rigorous step-by-step ablation studies to compare each component.  

\begin{figure}
\begin{floatrow}
\ffigbox{%
  \includegraphics[width=1.05\linewidth]{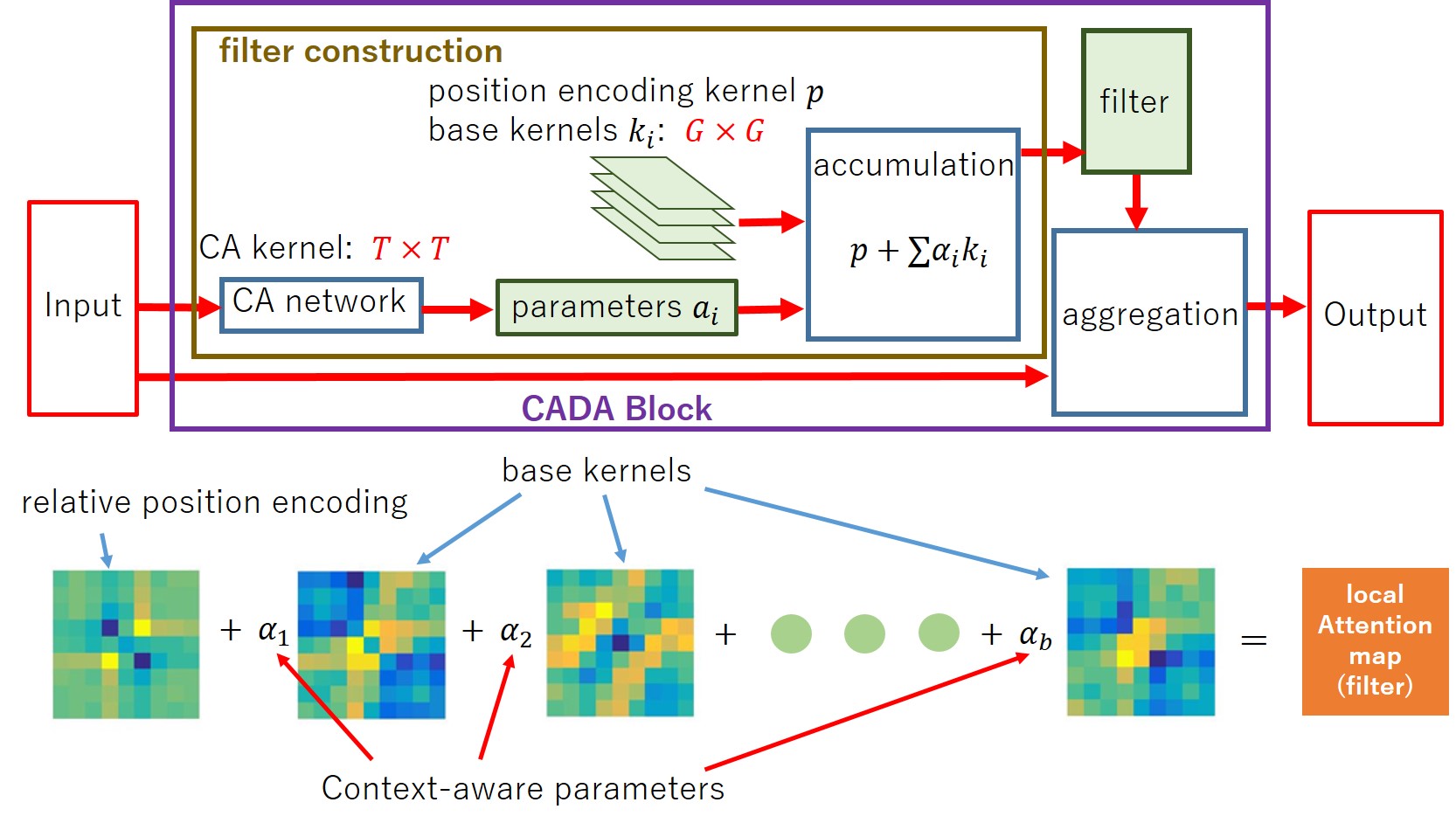}%
}{%
  \caption{Schematic of context-aware decomposed attention (CADA) structure}%
  \label{Fig:1}
}
\capbtabbox{%
\begin{tabular}{l|ll}
    \toprule
        & context & spatial\\
        & -aware & -invariant \\
    \midrule
      locally-connected & No & No \\
      convolution & No & Yes \\
      local self-attention & Yes & Somewhat \\
      CADA & Yes & Somewhat \\
      DA & No & Somewhat \\
    \bottomrule
  \end{tabular} 
}{%
  \caption{Properties of each structure }%
  \label{Tab:1}
}
\end{floatrow}
\end{figure}

The local SA networks without patch initial layer \cite{huang2019ccnet, wang2020axial, ramachandran2019stand, hu2019local, bello2019attention, zhao2020exploring} and with patch initial layer \cite{hassani2022neighborhood} give promising results.  They are easier to compare with the CNNs, so we focused on the local SA networks.  Note that this locality is different from popular window-based local Transformers, such as Swin Transformer \cite{liu2021swin}.  Their attention kernels are the same within the local window.  However, the attention kernels of the local SA networks are different depending on the location, which is discussed further in \cref{sec:global}.  Previous local SA networks required position encodings and reported relative position encoding is better than absolute position encoding \cite{bello2019attention, srinivas2021bottleneck, ramachandran2019stand, wang2020axial}.  Additional trainable computation to reshape relative position encoding is helpful \cite{hu2019local, d2021convit}.  In both cases, the position encodings are the same as the DW convolutional filters.  Using all trainable position encoding does not work \cite{hu2019local}, which is similar to the locally-connected networks.   So, the local SA structures are not entirely spatially variant.  We can assume they have inductive biases similar to DW convolutions.  To study that, we introduced context-aware decomposed attention (CADA), which decomposes attention maps with trainable base kernels.  It is the same idea as the low-rank SA structures for reducing computation costs \cite{wang2020linformer, choromanski2020rethinking, katharopoulos2020transformers}.  CADA has one trainable kernel with the same size as the other base kernels to represent the relative position encodings.  The base kernels are mixed using accumulation parameters computed through CA networks using the local CA kernel from the input feature map.  CADA has two localities, one for the base kernels and the other for the CA kernels.  Spatial invariance bias can be relaxed by adding base kernels.  The schematic of the local attention map of CADA is given in \cref{Fig:1}.  This filter construction is similar to traditional image processing, where carefully hand-crafted filters are provided, and they are mixed through the local kernel information.  The aggregation block of CADA is the same operation as the DW convolution and the aggregation of the local SA structure.  The main difference is how the attention maps/filters are constructed.  We also studied non-CA decomposed attention (DA) to show the importance of context awareness.  The properties of each structure are shown in \cref{Tab:1}.  All networks could have the same locality bias.

Recent hierarchical structures apply downsampling between the stages \cite{liu2021swin, hou2022vision, wu2021p2t, liu2022convnet}, similar to classical image processing, which implements subsampling after the low-pass filter to avoid aliasing \cite{gonzalez2009digital, szeliski2010computer}.  On the other hand, popular implementations of ResNets  \cite{gross2016training, he2019bag} have downsampling within the skip connection.  We observed that the accuracy could be boosted by applying some filters before subsampling to implement downsampling between the stages of ResNet.  We called these filters "downsampling filters" to distinguish filters for feature extraction, which we call "spatial filters."  We studied the inductive biases of both filters.  

We used the ResNet50-D \cite{he2016deep, he2019bag} backbone for ablation studies.  As the local SA networks, some of the CADA give significantly lower computation costs than the original convolution, which only means that separable convolution structures \cite{howard2017mobilenets, chollet2021xception} have lower computation costs.  The accuracy of CADA saturates as the base kernel size increases, which is consistent with the DW convolution's results \cite{tan2019mixconv}.  We showed that context awareness in CADA was crucial in inducing better results, but the larger CA kernel does not provide more information.  Relaxing the spatial invariance by adding more base kernels in CADA improves the accuracy.  Even though removing spatial invariance altogether makes it difficult to train \cite{neyshabur2020towards, novak2018bayesian}.  Additional spatial invariance through relative position encoding is also preferable.  

We also observed that the locality bias of the downsampling filters is more strict than the spatial filters.  Adding more base kernels to the downsampling filters can mitigate the local bias; however, accuracy does not improve with larger kernels, which contradicts the properties of the spatial filters.  Interestingly, the downsampling filter does not need multi-head structures; applying the same filter in all channels suffices.  The lists of preferable settings are given in \cref{Tab:properties}.



\begin{table}[t]
\begin{threeparttable}[t]
  \caption{Preferable settings.  Having several base kernels can relax spatial invariance inductive bias.  The CA kernel size represents the locality bias for context awareness, and the aggregation kernel size represents the locality bias of filters.  }
  \centering
  \begin{tabular}{l|c|c|c}
    \toprule
        & \multicolumn{2}{c|}{Accuracy}  & computational \\
         & spatial filter & downsampling filter & /space  complexity \\
    \midrule
       number of heads & many$^*$ & any & less\\
    \midrule
    number of base kernels &  \multicolumn{2}{c|}{many$^*$} & less\\
    \midrule
    context-aware (CA) kernel size&  \multicolumn{2}{c|}{more than $1\times1$ $^*$} & smaller\\
    \midrule
        aggregation kernel size & larger$^*$ & $3\times 3$ & smaller\\
    \bottomrule
  \end{tabular} 
  \label{Tab:properties}

\begin{tablenotes}
\item[*] improvement saturates
\end{tablenotes}
\end{threeparttable}
\end{table}


\section{Related work}

Several studies are extending CNNs and local SA networks toward each other.  We believe our study can generalize both the CNNs-based and local SA-based approaches.  \cref{Fig:1} shows the schematic of the CADA block.  The original local SA block gives filters straight from the CA network block using queries and keys.  Also, in the local SA structures, the input features are transformed into values before going into the aggregation block.  The original convolutional blocks do not have filter construction blocks; filters are not context-aware and trained through back-propagation.  

\textbf{Local SA-based approach:} The patchwise SAN \cite{zhao2020exploring} and involution \cite{li2021involution} do not frame their networks with the base kernels. However, their implementation provides the same general structure as our CADA block.  Both networks use $1 \times 1$ convolution with a bias to compute filters; hence, these biases represent position-encoding kernels.  The involution uses $1 \times 1$ CA kernels, and the size of the patchwise SAN CA kernels is the same as that of base kernels.  \cite{zhao2020exploring} tested various CA networks and compared the patchwise SAN  with more standard pairwise SAN, which constructs filters through the pairwise computation of each location with the position encodings.  The patchwise SAN gave better results than the pairwise SAN throughout different CA networks, including traditional local SA networks.  Hence, we did not repeat the ablation studies of exchanging the CADA block with the local SA block.

\textbf{CNN-based approach:} To reduce the hard-coded spatial invariance, LRLC \cite{elsayed2020revisiting} provides several trainable convolutional weights.   Their structures are close to ours; however, their base kernels include channel direction, leading to more restrictive inductive biases.  Because of the high computational cost of convolution, they introduced per-row and per-column weights to represent their filters.  Their construction showed that spatial invariance-induced locally-connected networks provide better results than CNNs using small datasets.  LRLC was considered both with and without CA computation and showed that context-awareness helps with less aligned data.  The CondConv \cite{yang2019condconv} and dynamic convolution \cite{chen2020dynamic} give similar structures, and they use several filters like inceptions \cite{szegedy2015going, chollet2017xception} and mix them using trainable parameters according to each spatial location.  The CA LRLC, CondConv, and dynamic convolution use full CA kernels, which are the same size as the input feature map.
 
\section{Context-aware decomposed attention (CADA) block}
\begin{figure}[!t]
  \centering
   \includegraphics[width=0.9\linewidth]{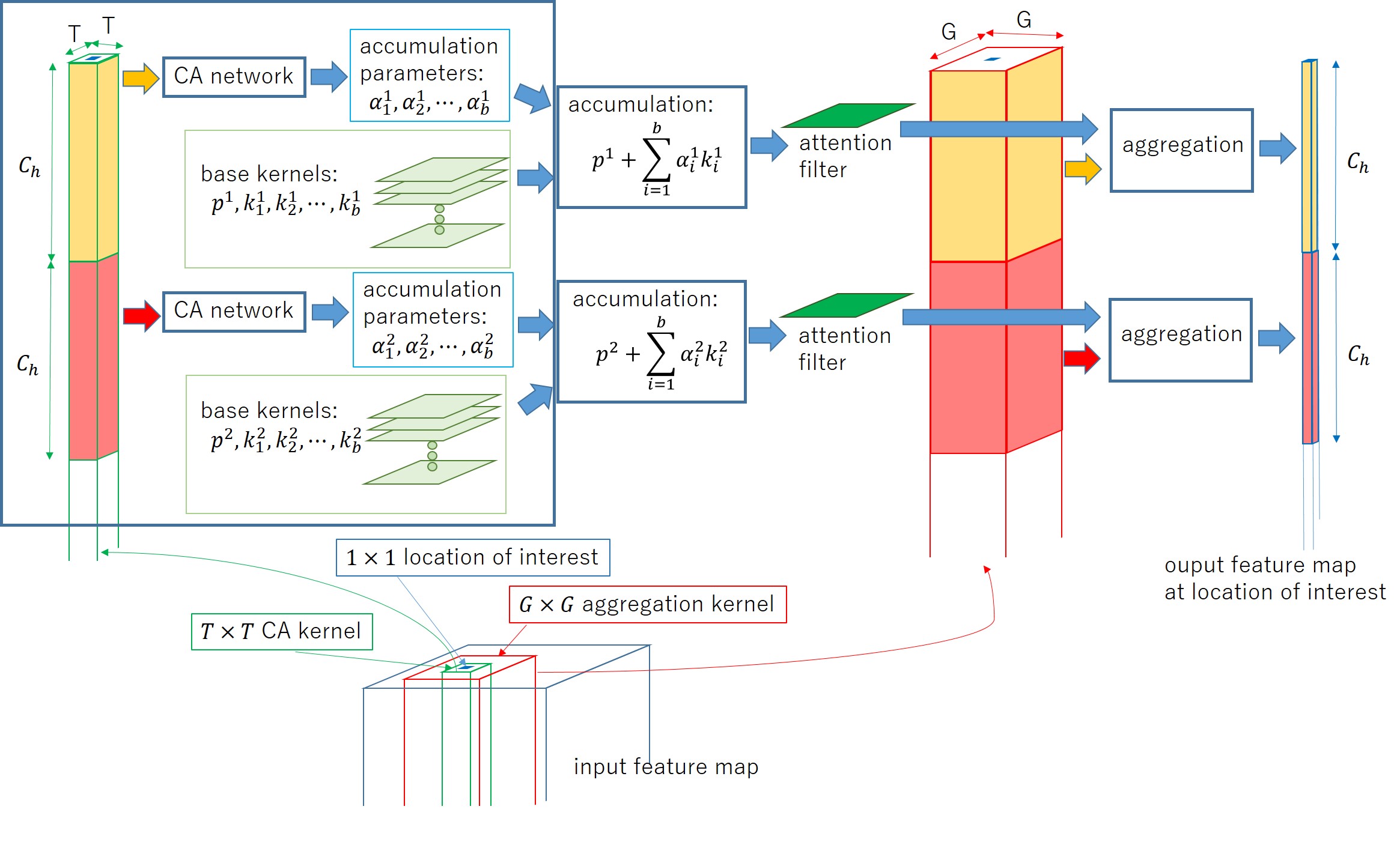}
   \caption{Context-aware decomposed attention (CADA) schematic on the location of interest}
   \vspace{-0.6cm}
   \label{fig:CADA}
\end{figure}
\begin{figure}[t]

\begin{floatrow}

\ffigbox[\FBwidth][][]{%

  \includegraphics[width=0.84\linewidth]{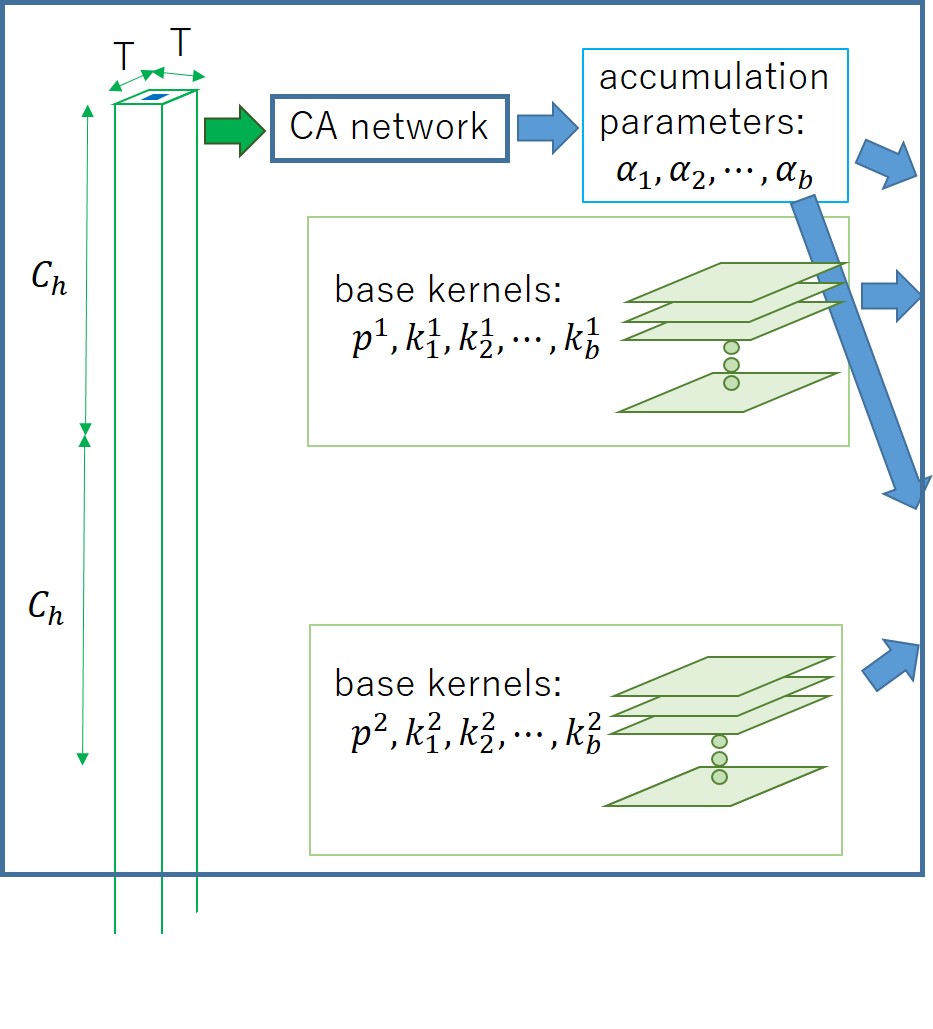}%
}{%
  \vspace{-0.2cm}
  \caption[]{Part of the context-aware decomposed attention, shared parameter (CADAsp) schematic. The rests are the same as in \cref{fig:CADA}}
  \label{fig:CADAsp}
}
\ffigbox[\FBwidth][][]{%
  \begin{minipage}[t]{0.32\linewidth}
    \centering
    \vspace{0pt}
    \includegraphics[width=\textwidth]{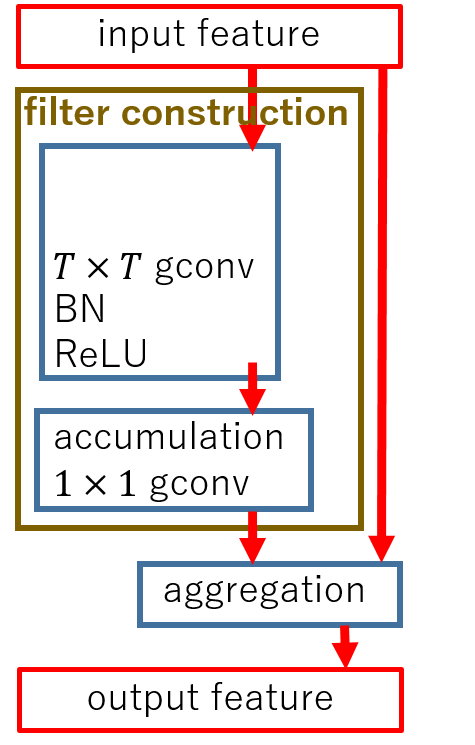}
    \subcaption{network A}
  \end{minipage}
  \begin{minipage}[t]{0.32\linewidth}
    \centering
    \vspace{0pt}
    \includegraphics[width=\textwidth]{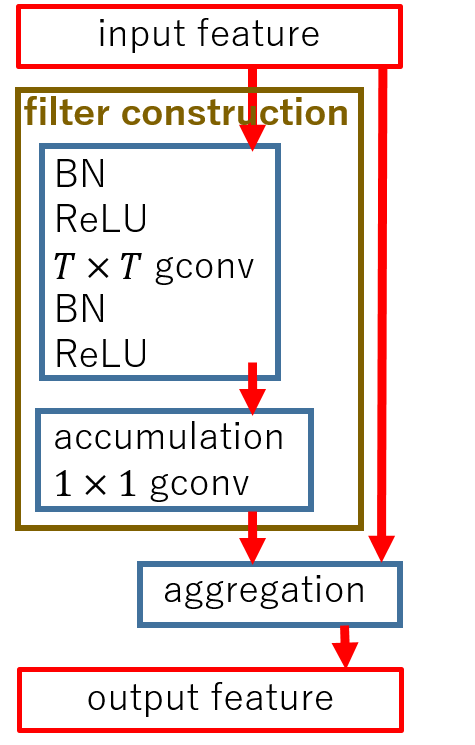}
    \subcaption{network B}
  \end{minipage}
  \begin{minipage}[t]{0.32\linewidth}
    \centering
    \vspace{0pt}
    \includegraphics[width=\textwidth]{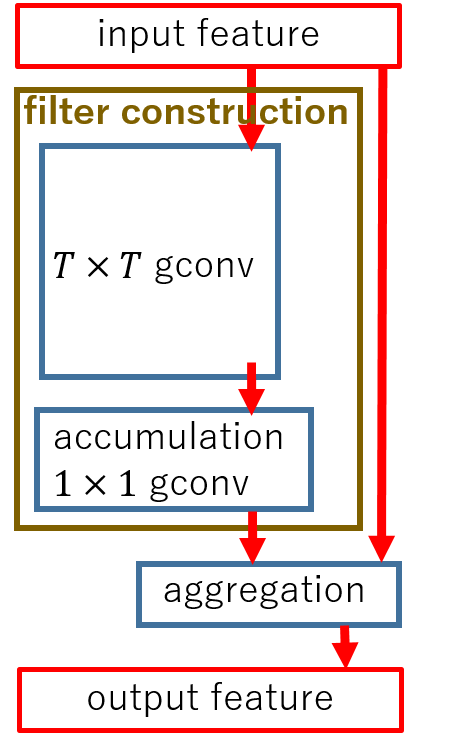}
    \subcaption{network C}
  \end{minipage}
}{%
  \caption[]{context aware (CA) networks}%
  \label{Fig:CAnetwork}
}
\end{floatrow}
\end{figure}

We decomposed $G \times G$ attention map $F^{l,h} \in \mathbb{R}^{G \times G}$  at location $l$ and head $h$ as 
\begin{equation}
F^{l,h} = p^h + \sum_{i=1}^b \alpha^{l,h}_i k^h_i,
\label{eq:CADA}
\end{equation}
where $p^h \in \mathbb{R}^{G \times G}$ is the $G \times G$ relative position encoding kernel, $k_i^h \in \mathbb{R}^{G \times G}$ is the $G \times G$ base kernel, $b$ is the number of base kernels, and $\alpha^{l,h}_i \in \mathbb{R}$ is the accumulation parameter computed through the CA network using $T \times T$ input feature map $I^{l,h}_{CA} \in \mathbb{R}^{T \times T \times C_h}$ with channel size $C_h$.  The schematic of CADA construction is shown in \cref{fig:CADA}.  The accumulation parameters $\alpha^{l,h}$ are computed for each head.  As shown in \cref{fig:CADAsp}, we also considered shared parameter (sp) cases where $\alpha^{l,h}$ are the same for all heads and computed using $T \times T$ input feature map with all channels.  We call them CADAsp.

There are several choices of CA networks.  We consider three networks, as shown in \cref{Fig:CAnetwork}.  
CADAsp can use similar CA networks; the only modification required is to exchange the group convolution (gconv) to regular convolution.

Once attention maps are constructed, the aggregation operation is the same as the aggregation method of the local MH SA networks.  We followed the patchwise SAN \cite{zhao2020exploring}\footnote{https://github.com/hszhao/SAN} implementation.  To construct attention maps, we only need $T \times T$ convolution and $1 \times 1$ convolution with grouping for CADA and without grouping for CADAsp.  The details and a few modifications of aggregation implementations are listed in \cref{sec:code}.  

To study spatial invariance, we can change the size of base kernels.  We could choose a different number of base kernels for each stage, denoting $b=(i, j, k, l)$ for four stages in ResNet50.  To study locality, we can modify the size of the CA kernels and aggregation kernels.  The number of heads is another variable, and we use $C_h$ to denote the number of channels inside each head.  




\begin{figure*}[t]
\captionsetup[subfigure]{justification=centering}
  \centering
  \hspace*{\fill}%
  \begin{subfigure}[b]{0.24\linewidth}
    \includegraphics[width=\linewidth]{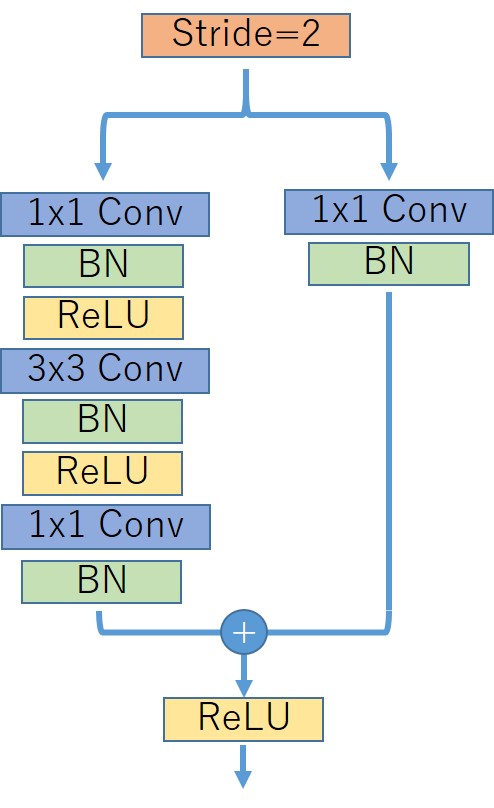}
    \caption{original ResNet \cite{he2016deep}\\ \hphantom{(b)}}\label{fig:PostOrig}
  \end{subfigure}
  \begin{subfigure}[b]{0.24\linewidth}
    \includegraphics[width=\linewidth]{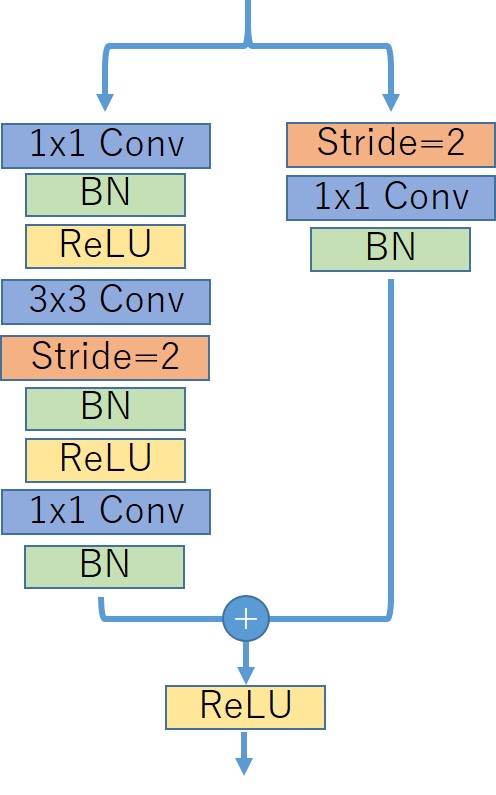}
    \caption{ResNet-B \cite{gross2016training}\\ \hphantom{(b)} (TorchVision)} \label{fig:PostResNet-B}
  \end{subfigure}
  \begin{subfigure}[b]{0.24\linewidth}
    \includegraphics[width=\linewidth]{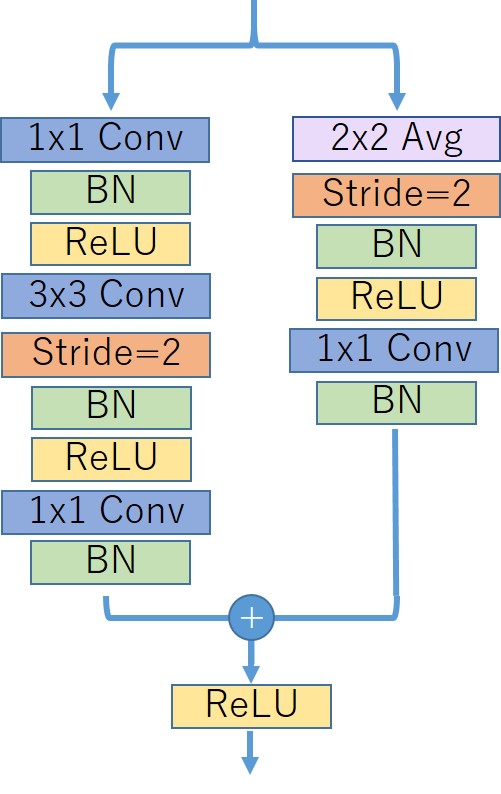}
    \caption{ResNet-D \cite{he2019bag}\\ \hphantom{(b)}} \label{fig:PostResNet-D}
  \end{subfigure}
  \begin{subfigure}[b]{0.24\linewidth}
    \includegraphics[width=\linewidth]{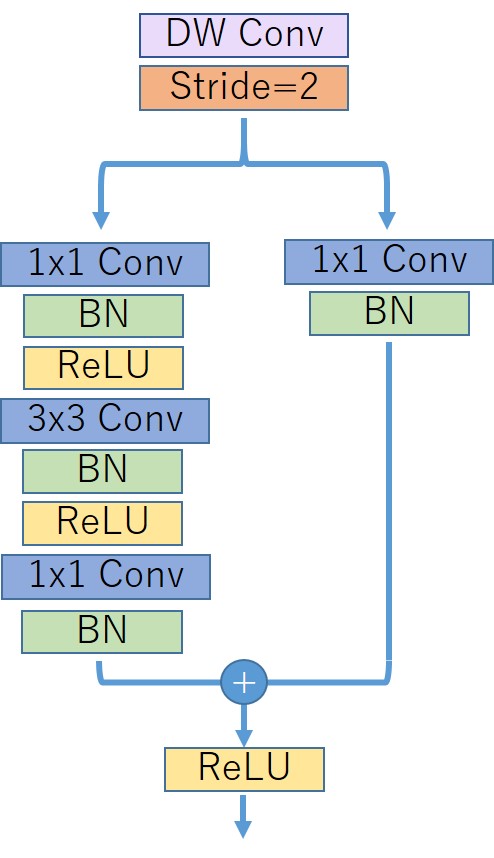}
    \caption{ResNet-E \\ \hphantom{(b)}} \label{fig:PostOurs}
  \end{subfigure}
  \hspace*{\fill}%
  \caption{Different modifications of the ResNet regarding downsampling.}
  \label{fig:ResNets}
  \label{fig:PostAct_}
\end{figure*}
\section{Different functionality of filters}\label{sec:functionalityFilters}

Neural networks downsample images into a hierarchical order, similar to the traditional multi-scale feature representations.  Even the patchwise images for Transformers and MLP-based networks use downsampling to mitigate the high computation cost of handling high-resolution images \cite{liu2021swin, hou2022vision, wu2021p2t}.  These networks usually implement downsampling by subsampling with pooling or convolution.  Subsampling may violate the Nyquist rate and generate aliasing \cite{oppenheim1997signals}.  Therefore, classical image processing applies a low-pass filter before subsampling \cite{gonzalez2009digital, szeliski2010computer}.  For neural networks, the assumption is that the networks can learn the trait without an implicit low-pass filter.  

The popular implementation of ResNets \cite{gross2016training, he2019bag} moved subsampling inside skip connection to incorporate with other filters, as shown in \cref{fig:PostResNet-B} and \cref{fig:PostResNet-D}.  On the other hand, the hierarchical networks with patchwise images apply downsampling in between stages.  ConvNext \cite{liu2022convnet} took this idea of downsampling to DW convolutional networks and gave empirically better results than the downsampling within the residual block.  We introduce ResNet-E, as shown in \cref{fig:PostOurs}, which adds a DW convolutional filter just before subsampling in the original ResNet \cite{he2016deep}. ResNet-E gave better accuracy than downsampling within the residual block, as shown in \cref{tab:post_c} and \cref{Tab:longImageNet} from Appendix.  Therefore, to implement downsampling, some filters should be given before subsampling to mitigate aliasing.  We call these filters "downsampling filters" and distinguish them from filters for feature extraction, which we call "spatial filters."
We observed that using DW convolutional filters as downsampling filters yields results comparable to or better than using implicit low-pass filters as downsampling filters; details are provided in \cref{sec:ResNet-E}.  Using ResNet-E, we could study the inductive biases of the downsampling filters.

\section{Experiments}\label{sec:ExperimentSpatial}
We used the ResNet50-D \cite{he2019bag} on the ImageNet classification task \cite{russakovsky2015imagenet} as a backbone for our ablation studies. Since ImageNet images are compressed by JPG, the high-frequency bands of the images have already been removed.  Therefore, it makes no sense to apply an extra low-pass filter in STEM.  For ResNet-E, we used the same STEM as ResNet-D except for the final max pooling so that the downsampling filters could apply before each stage.  

For training, we set the cosine learning rate to 0.1 \cite{loshchilov2016sgdr}, momentum to 0.9, and weight decay to 1e-4 \cite{he2016deep}.  We used synchronous SGD with 256 minibatch and 120 epochs on 8 V100 GPUs.  We applied basic data augmentation of random cropping to 224x224 patches, random horizontal flipping, and normalization.  We ran a few experiments with a modern training scheme and presented the results in \cref{sec:modern}.

For ablation studies, we replaced $3 \times 3$ convolution and reported maximum Top-1 validation accuracy.  For a fair comparison of the components, we also experimented with MH DW convolution.  Recently, it has been shown that removing normalization can improve the accuracy of DW convolution \cite{liu2022convnet}.   We also tested with and without batch normalization (BN) and ReLU activation before $3 \times 3$ convolution on ResNet-D. No BN and ReLU with DW convolution following $1 \times 1$ convolution is equivalent to $G \times G$ convolution decomposed into $1 \times 1$ convolution and $G \times G$ DW convolution. \cref{tab:BN/ReLU} shows the accuracy of the ImageNet classification task with and without BN and ReLU. From these results, we chose not to use BN and ReLU with DW convolution in our experiments.

\begin{table}[h]
  \caption{Accuracy of the ImageNet classification task. Each head has 8 channels.}
  \centering
  \setlength{\tabcolsep}{5pt} 
  \begin{tabular}{l|c|c|c}
    \toprule
        & $3 \times 3$ conv (original)     & $7 \times 7$ DW conv w/o head & $7 \times 7$ DW conv with head  \\
    \midrule
    BN/ReLU   & \bf{78.26}\%  & 77.18\% & 77.22\% \\
    None    & 77.39\% & \bf{78.40}\% & \bf{78.05}\% \\
    BN    & 77.58\% & \bf{78.35}\% & \bf{78.08}\% \\
    ReLU & \bf{78.26}\% & 78.01\% & 77.66\% \\
    \bottomrule
  \end{tabular}
    \vspace{-0.2cm}
  \label{tab:BN/ReLU}
\end{table}

\begin{table}[h]
  \caption{Accuracy of the ImageNet classification task using CA network A from \cref{Fig:CAnetwork} with $3 \times 3$ CA kernels, $7 \times 7$ aggregation kernels, and 8 channels in each head ($C_h = 8$).  "$b$" represents the number of base kernels in each stage.}
  \centering
  \setlength{\tabcolsep}{5pt} 
  \begin{tabular}{l|c|ccc}
    \toprule
        & with BN/ReLU     & \multicolumn{3}{c}{ without BN/ReLU }\\
        & Top-1 & Top-1 &  Params     &  FLOPs\\
    \midrule
    $b = (4, 4, 4, 4) $  & 77.89\%  & 78.64\% & 14.45M & 2.64G\\
    $b = (8, 16, 32, 64)$    & 77.85\% & 78.99\% & 15.96M & 2.86G \\

    \bottomrule
  \end{tabular}
  \label{tab:BN/ReLU_CADA}
\end{table}

We also tested CADA with and without BN and ReLU using CA network A from \cref{Fig:CAnetwork} on ResNet-D.  \cref{tab:BN/ReLU_CADA} shows the accuracy of the ImageNet classification task using CADA with $3 \times 3$ CA kernels, $7 \times 7$ aggregation kernels, and 8 channels in each head.  Both $b = (4, 4, 4, 4) $ and $b = (8, 16, 32, 64)$ cases preferred no BN and ReLU; hence we did not use BN and ReLU in our CADA experiments.

\begin{table}[h]
  \caption{Architecture profiles of the ResNet-based local self-attention (SA) networks on the ImageNet classification task with a similar training scheme as ours.}
  \centering
  \setlength{\tabcolsep}{5pt} 
  \begin{tabular}{llll}
    \toprule
    Architecture     &  Top-1   &   Params & FLOPs  \\
    \midrule
    Patchwise SAN19 \cite{zhao2020exploring}     &  78.2\% & 20.5M & 3.3G     \\
    Axial ResNet-S \cite{wang2020axial}    & 78.1\% & 12.5M & 3.3G  \\
    RedNet(involution)-50 \cite{li2021involution}    & 78.4\% & 15.5M & 2.7G   \\

    \bottomrule
  \end{tabular}
  \label{tab:LoaclSA}
\end{table}
  
Even though the construction of our attention map is quite different from other local SA-based methods, CADA can provide comparable results using a similar training scheme, as shown in \cref{tab:BN/ReLU_CADA} and \cref{tab:LoaclSA}.  Our intention was not to create the SOTA networks but to study the inductive biases.  Hence, we did not conduct an extensive hyper-parameter search. 

\begin{table}[htb]
  \caption{Accuracy of the ImageNet classification task from different CA networks in \cref{Fig:CAnetwork} and a non-CA network in \cref{Fig:DAnetwork} using $3 \times 3$ CA kernels and $7 \times 7$ aggregation kernels. The accuracy from $7 \times 7$ MH DW convolution is also given for the baseline.  "$C_h$" represents the number of channels in each head, and "$b$" represents the number of base kernels in each stage.}
  \centering
  \setlength{\tabcolsep}{4.5pt} 
  \begin{tabular}{c|cc|cc|cc|cc}
    \toprule
    & \multicolumn{4}{c|}{$C_h = 8$}      & \multicolumn{4}{c}{$C_h = 32$} \\
    \midrule
    DW conv & \multicolumn{4}{c|}{78.05\%}      & \multicolumn{4}{c}{77.98\%}  \\    
    \midrule
    & \multicolumn{2}{c}{$b = (4,4,4,4)$} & \multicolumn{2}{c|}{$b = (8,16,32,64)$} & \multicolumn{2}{c}{$b = (4,4,4,4)$} & \multicolumn{2}{c}{$b = (8,16,32,64)$} \\
    \midrule
    & CADA & CADAsp & CADA & CADAsp & CADA & CADAsp & CADA & CADAsp\\
    \midrule
    network A   & 78.64\%  & 78.69\% & 78.99\% & 78.92\%& 78.76\%  & 78.74\% & 79.04\% & 79.09\%\\
    network B   & 78.58\% & 78.54\% & 78.75\% & 79.08\% & 78.25\%  & 78.24\% & 78.99\% & 78.99\%\\
    network C   & 78.76\% & 78.82\% & 78.21\% & 78.95\% & 78.62\%  & 78.12\% & 78.74\% & 78.93\%\\
    \midrule
    & DA & DAsp & DA & DAsp & DA & DAsp & DA & DAsp\\
    \midrule
    network D   & 78.13\%  & 78.45\% & 77.84\% & 78.18\%& 77.95\%  & 77.93\% & 77.58\% & 77.32\%\\

    \bottomrule
  \end{tabular}
  \label{tab:CADAnetwork}
\end{table}

We tested with different CA networks in \cref{Fig:CAnetwork} with $3 \times 3$ CA kernels and $7 \times 7$ aggregation kernels on ResNet-D.  CADAsp use convolution instead of group convolution in CA networks.  \cref{tab:CADAnetwork} shows that CADAsp gave similar results to CADA on network A and network B.  We chose to use network A for our CADA and CADAsp experiments. 

In the following section, we first show the inductive biases of the spatial filters using ResNet-D.  Then, we provide the experimental results of the downsampling filters' biases using ResNet-E.

\subsection{Spatial filter}

\begin{figure}
\begin{floatrow}

\capbtabbox{%
  \begin{tabular}{c|cc|cc}
    \toprule

    & \multicolumn{2}{c|}{$b = (4,4,4,4)$} & \multicolumn{2}{c}{$b = (8,16,32,64)$}  \\
    \midrule
    & DA & DAsp & DA & DAsp\\
    \midrule
    D   & 78.13\%  & 78.45\% & 77.84\% & 78.18\%\\
    E   & 78.24\% & 77.95\% & 78.14\% & 78.09\% \\
    F   & 78.25\% & 78.00\% & 78.28\% & 78.12\% \\
    \bottomrule
  \end{tabular}
}{%
  \caption{Accuracy of the ImageNet classification task from different non-CA networks in \cref{Fig:DAnetwork} using $7 \times 7$ aggregation kernels, and 8 channels in each head.  "$b$" represents the number of base kernels in each stage.}%
  \label{Tab:DA}
}
\hspace{-8pt}
\ffigbox[\FBwidth][][]{%

  \begin{minipage}{0.32\linewidth}
    \includegraphics[keepaspectratio, scale=0.3]{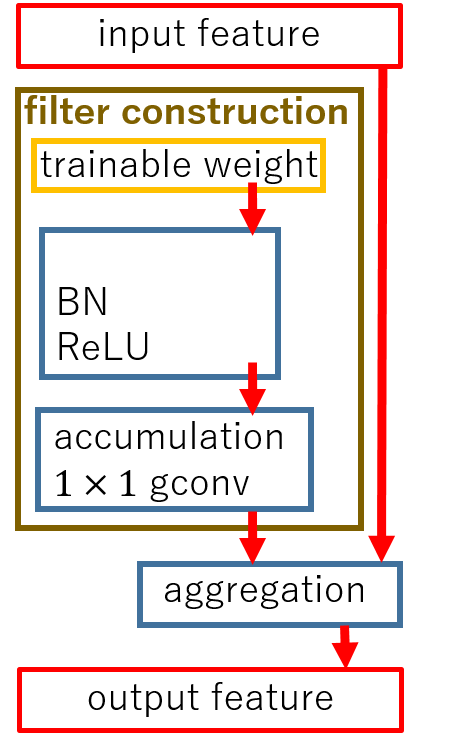}
    \subcaption{network D}
  \end{minipage}
  \begin{minipage}{0.32\linewidth}
    \centering
    \includegraphics[keepaspectratio, scale=0.3]{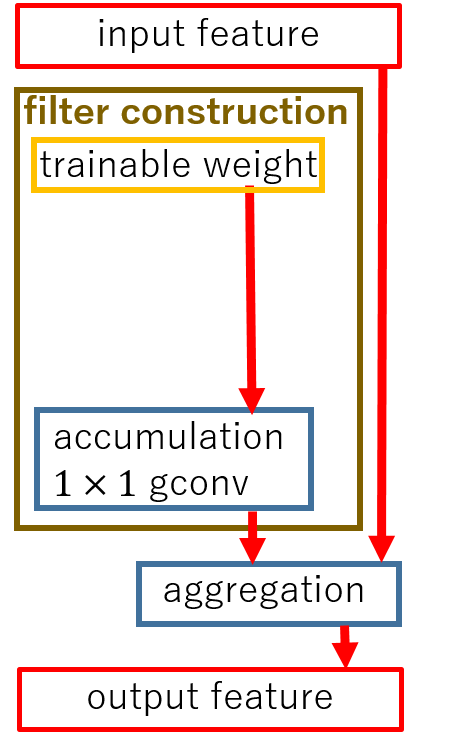}
    \subcaption{network E}
  \end{minipage}
  \begin{minipage}{0.32\linewidth}
    \centering
    \includegraphics[keepaspectratio, scale=0.3]{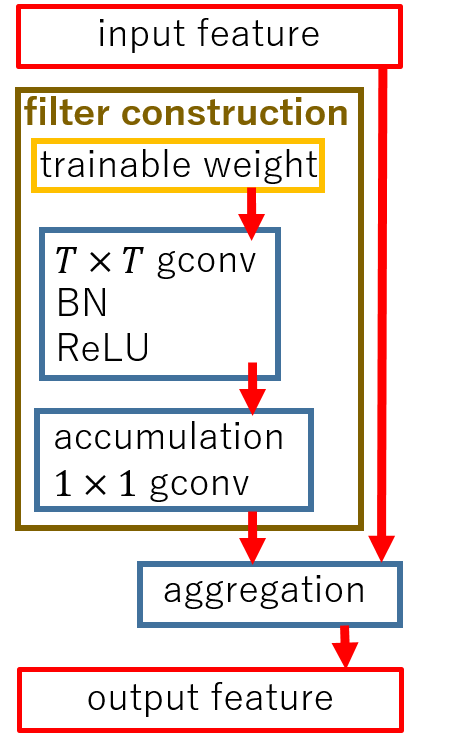}
    \subcaption{network F}
  \end{minipage}
}{%
  \caption{non-CA networks for decomposed attention (DA)}%
  \label{Fig:DAnetwork}
}
\end{floatrow}
\end{figure}

CADA has context awareness and relaxed spatial invariance bias.  On the other hand, DW convolution does not have context awareness and has a strong spatial invariance bias.  Both CADA and DW convolution can have the same locality in the aggregation kernels.  So, to ensure the importance of context awareness, we removed them from CADA and CADAsp. 

\begin{table}[h]
  \caption{Accuracy of the ImageNet classification task given by no-head DAsp using $7 \times 7$ and $9 \times 9$ aggregation kernels and the different number of base kernels represented in the second rows.}
  \centering
  \setlength{\tabcolsep}{5pt} 
  \begin{tabular}{c|ccccc|c}
    \toprule
    &\multicolumn{5}{c|}{DAsp} & DW conv\\
    \midrule
     $b=$ & 2 & 4 & 8 & 16 & 32 & \\
    \midrule
    $7\times7$   & 78.55\%  & 78.41\% & 78.13\% & 78.44\%& 78.24\%  & 78.40\%\\
    $9\times9$   & 78.32\% & 78.58\% & 78.33\% & 78.19\% & 78.31\%  & 78.13\%\\
    \bottomrule
  \end{tabular}
  \label{tab:DAnetwork}
\end{table}

As shown in \cref{Fig:DAnetwork}, we provided trainable weights instead of input features to construct accumulation parameters.  We call this construction DA and DAsp to correspond to CADA and CADAsp.  \cref{Tab:DA} shows the accuracy of the ImageNet classification task using different non-CA networks in \cref{Fig:DAnetwork}.  We chose network D for our non-CA network.  As shown in \cref{tab:CADAnetwork} and \cref{Tab:DA}, DAsp gave similar results to DA.  
We tried several DAsp networks with a different number of base kernels in each layer using $7 \times 7$ and $9 \times 9$ aggregation kernels without a head, as shown in \cref{tab:DAnetwork}.  It gave no significant difference with the DW convolution.  
We also experimented with the different number of heads using $3 \times 3$ CA kernels, $7 \times 7$ aggregation kernels, and two different sizes of base kernels, as shown in \cref{fig:Ch}.  It again showed no significant difference with the MH DW convolution.  
Hence, we can conclude that context awareness is a crucial structure, and we should study its locality bias.


\subsubsection{Multi head}
\begin{figure}[t]
  \centering
   \includegraphics[width=1.0\linewidth]{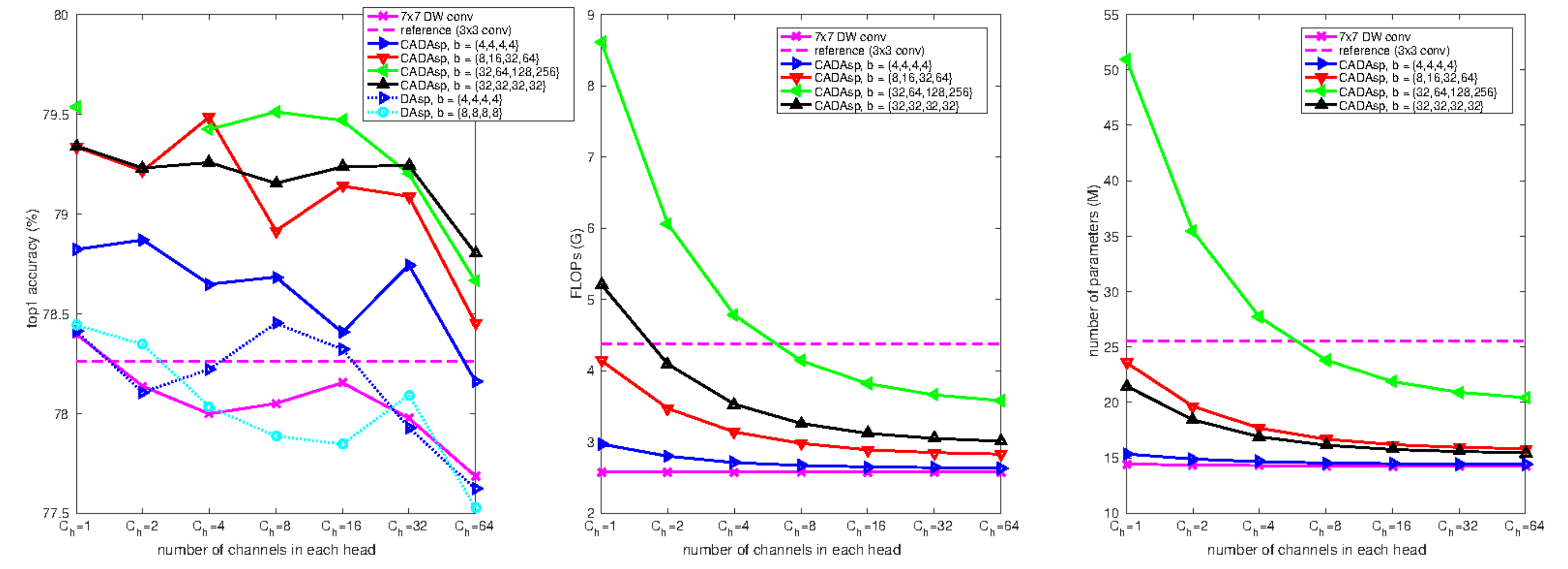}
   \caption{Left figure shows the accuracy of the Imagenet classification task, middle shows FLOPs, and right shows the number of parameters of corresponding networks depends on the size of the head using $3 \times 3$ CA kernels and $7 \times 7$ aggregation kernels. "$b$" represents the number of base kernels in each stage.}
   \label{fig:Ch}
\end{figure}

We tested CADAsp with a different number of channels in each head using $3 \times 3$ CA kernels and $7 \times 7$ aggregation kernels, as shown in \cref{fig:Ch}.  We also tested MH DW convolutional networks for comparison.  There is a trade-off between computational complexity and accuracy, but there is no strong degradation between $C_h=1$ to $C_h=32$.  Note, we could not produce a result of CADAsp with $C_h=2$ and $b=(32,64,128,256)$ due to the out-of-memory error in back-propagation.

\subsubsection{Spatial invariance} \label{sec:spatial}
\begin{figure}[t]
  \centering
  \vspace{-0.1cm}
   \includegraphics[width=1.0\linewidth]{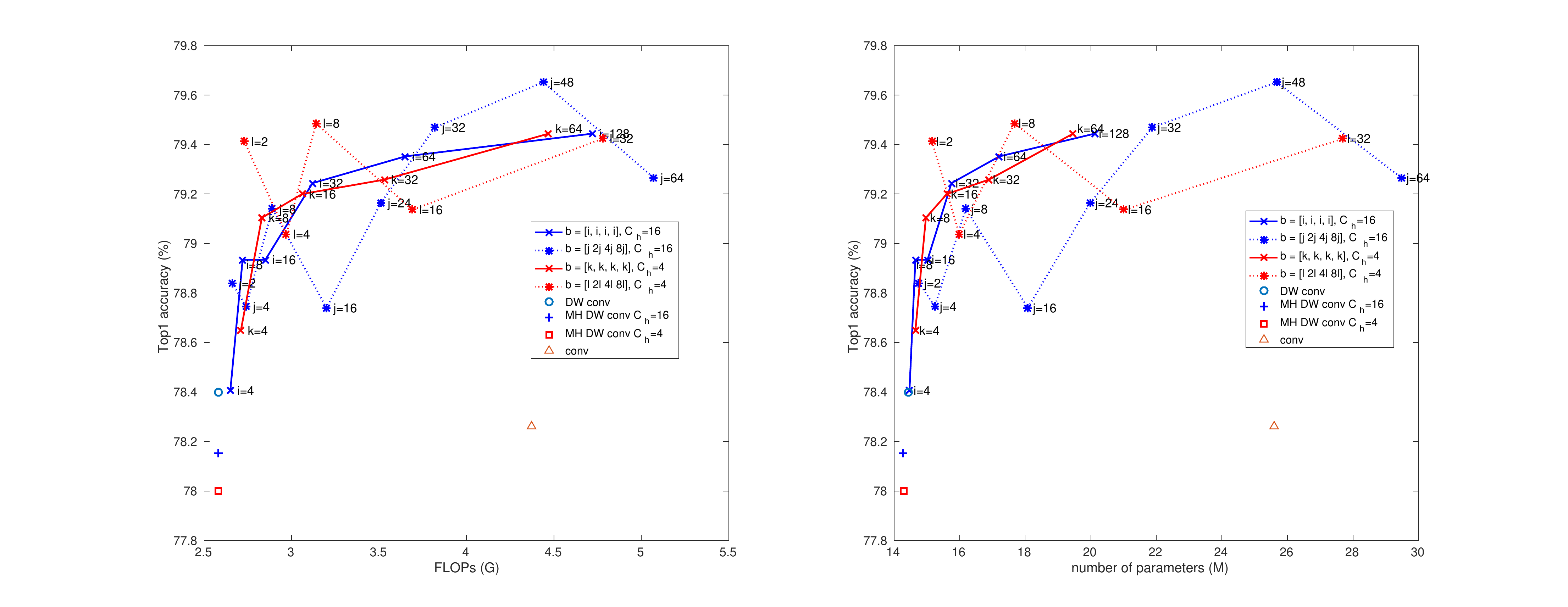}
   \vspace{-0.7cm}
   \caption{Accuracy of the ImageNet classification task depends on the size of the networks using $3 \times 3$ CA kernels and $7 \times 7$ aggregation kernels.  "$C_h$" represents the number of channels in each head, and "$b$" represents the number of base kernels in each stage.}
   \label{fig:baseNumber}
\end{figure}
To study spatial invariance, we first checked $7\times7$ locally-connected DW network, which is much more relaxed than the locally-connected network as in \cite{d2019finding}.  However, it only gave an accuracy of 76.68\%, which was significantly worse than 78.40\% of $7\times7$ DW convolution.  Hence, strong spatial invariance is the important inductive bias for locally-connected structures.

We could relax this inductive bias by adding $7\times7$ base kernels.  \cref{fig:baseNumber} shows the different number of base kernels over FLOPs and the number of parameters using $3 \times 3$ CA kernels and $7\times7$ aggregation kernels.  We tested two different ways of adding base kernels.  One was to add the same number of base kernels in each layer, $b=(i,i,i,i)$, which showed steady accuracy improvement over the size of the networks.  The other was to double every stage as the channel increases, $b=(j,2j,4j,8j)$, which was unstable; however, it still gave better accuracy compared with the DW convolution.  So, relaxing spatial invariance help improve the accuracy.

Even though many base kernels gave better accuracy, some base kernels were trained to be small and easy to be pruned.   So, the magnitude of spatial invariance cannot be determined solely on the number of base kernels.  \cref{tab:prune} shows the average number of base kernels over the heads in each layer after pruning, where we allowed 0.1\% accuracy reduction without fine-tuning and used the $L_1$ norm pruning.  The denominator shows the original number of bases.  Some of the layers had a strong correlation between base kernels as well.  Details are given in \cref{sec:kernels}.

\begin{table}[h]
  \caption{Average number of base kernels over heads in each layer after pruning.  }
  \centering
  \setlength{\tabcolsep}{4pt} 
  \begin{tabular}{ccc|cccc|cccccc|ccc}
$\frac{84}{128}$& $\frac{70}{128}$& $\frac{128}{128}$& $\frac{107}{128}$& $\frac{114}{128}$& $\frac{81}{128}$& $\frac{124}{128}$&$\frac{125}{128}$& $\frac{74}{128}$& $\frac{119}{128}$& $\frac{112}{128}$& $\frac{120}{128}$&$\frac{126}{128}$&$\frac{121}{128}$& $\frac{111}{128}$& $\frac{64}{128}$ \\
\midrule
$\frac{29}{48}$& $\frac{22}{48}$& $\frac{26}{48}$& $\frac{96}{96}$& $\frac{92}{96}$& $\frac{72}{96}$& $\frac{96}{96}$& $\frac{174}{192}$& $\frac{152}{192}$& $\frac{160}{192}$& $\frac{182}{192}$& $\frac{169}{192}$&$\frac{164}{192}$& $\frac{270}{384}$& $\frac{216}{384}$&$\frac{106}{384}$ \\
  \end{tabular}
  \label{tab:prune}
\end{table}

\begin{figure}[t]
  \centering
  \vspace{-0.4cm}
   \includegraphics[width=0.45\linewidth]{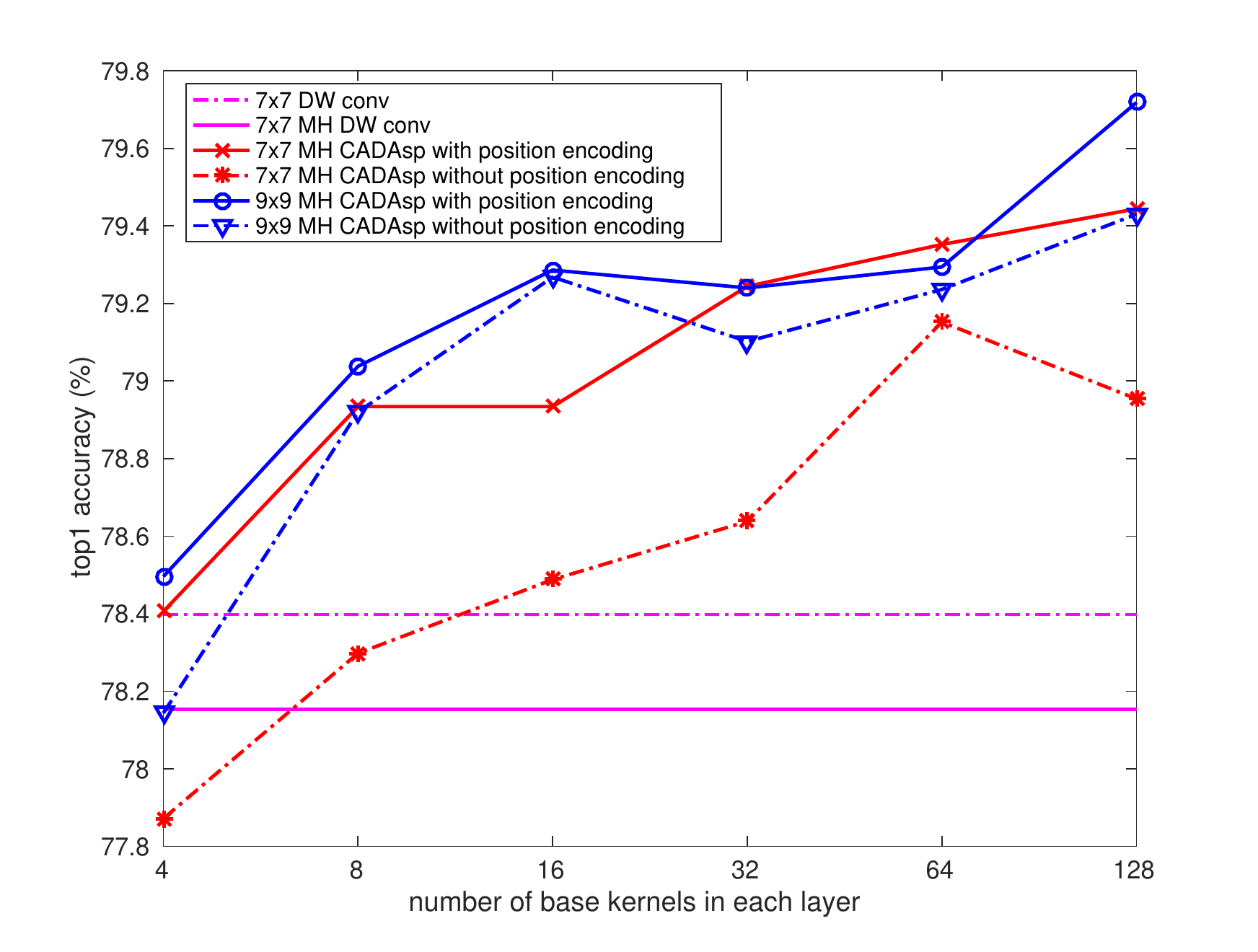}
   \vspace{-0.5cm}
   \caption{Accuracy of the ImageNet classification task depends on the size of base kernels in each layer using $3 \times 3$ CA kernels and 16 channels in each head ($C_h = 16$).\vspace{-0.4cm}}
   \label{fig:position}
\end{figure}

Our construction has a relative position encoding, which gives strong spatial invariance.  We tested with and without relative position encoding in CADAsp using $7 \times 7$ and $9 \times 9$ aggregation kernels, $3\times3$ CA kernels, and 16 channels in each head, as shown in \cref{fig:position}.  It shows that having strong spatial invariance slightly helps increase accuracy, which is in line with the same experiments using a vision transformer \cite{chen2021empirical}.  In the implementation, relative position encoding in the accumulation block is provided by adding a bias term in convolution, so additional computational cost is negligible.  

\begin{figure}[t]
  \centering
   \includegraphics[width=1.0\linewidth]{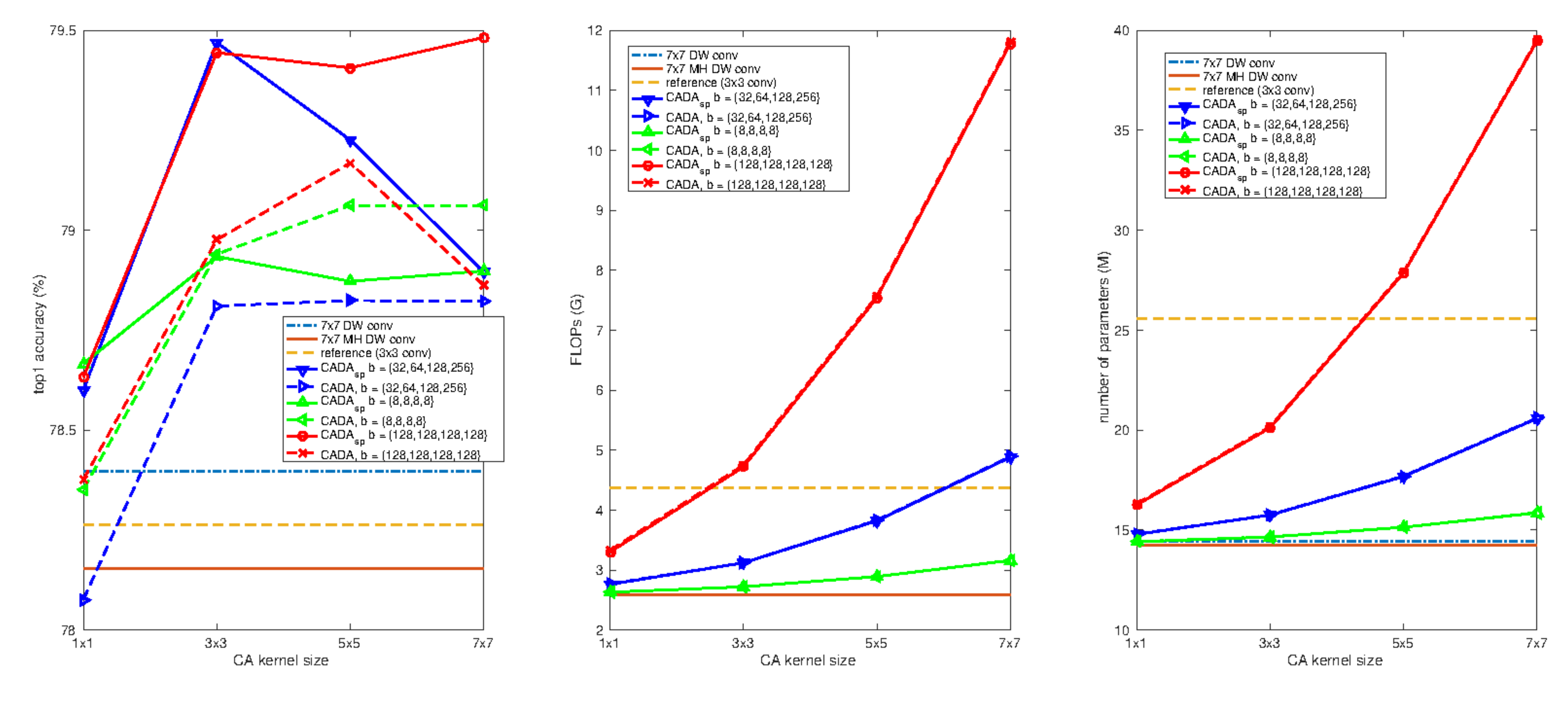}
   \vspace{-0.7cm}
   \caption{Left figure shows the accuracy of the Imagenet classification task, middle shows FLOPs, and right shows the number of parameters of corresponding networks depends on the CA kernel size using $7 \times 7$ aggregation kernels and 16 channels in each head ($C_h = 16$).  "$b$" represents the number of base kernels in each stage.
   }
   \label{fig:CAkernel}
\end{figure}
\subsubsection{Locality}

There are two localities in our construction, one for context awareness in the CA kernels and the other in the aggregation kernels, which is the same locality as the convolution.  \cref{fig:CAkernel} shows the accuracy of the ImageNet classification task depending on the size of CA kernels using $7 \times 7$ aggregation kernels and 16 channels in each head.  All layers have the same size of CA kernels.  The accuracy is quickly saturated with large CA kernels; hence, we only require local information to construct an attention map/filter for the classification task.

\begin{figure}[t]
  \centering
  \vspace{-0.2cm}
   \includegraphics[width=1.0\linewidth]{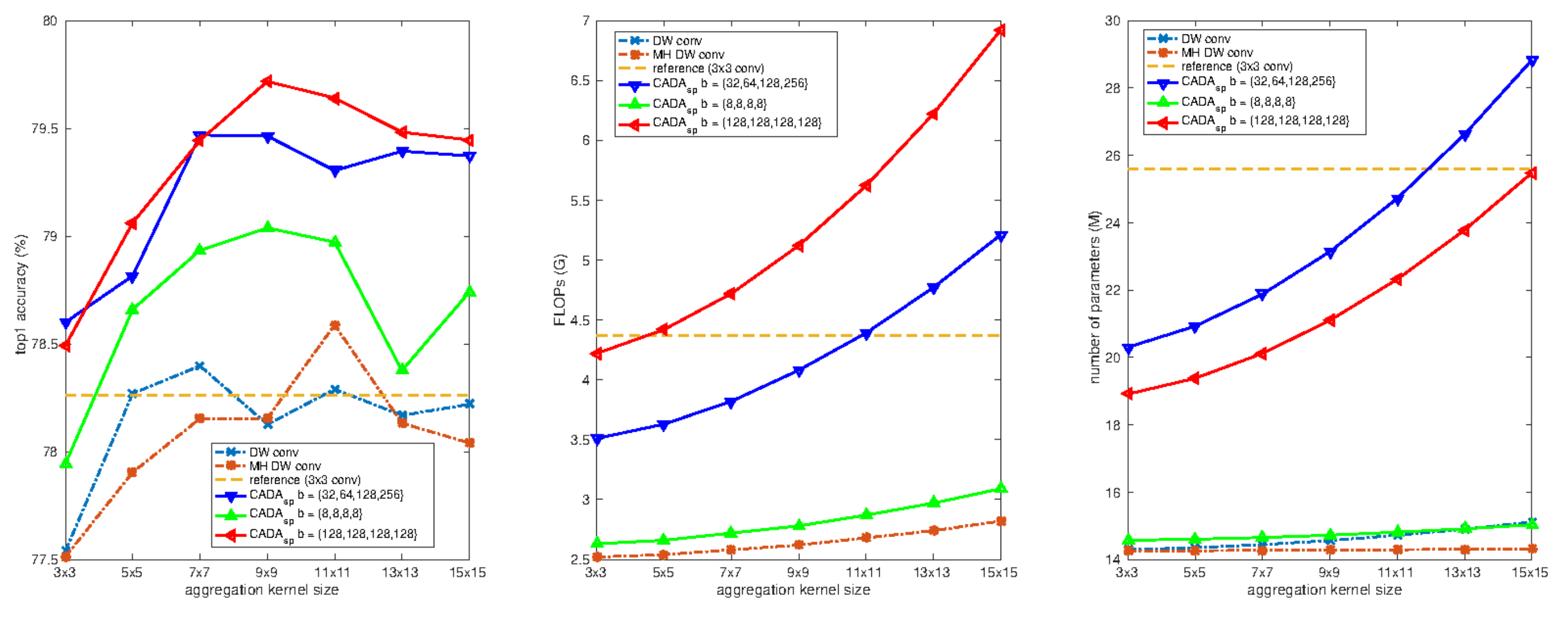}
   \caption{Left figure shows the accuracy of the Imagenet classification task, middle shows FLOPs, and right shows the number of parameters of corresponding networks depends on the aggregation kernel size using $3 \times 3$ CA kernels and 16 channels in each head ($C_h = 16$).  "$b$" represents the number of base kernels in each stage.
   \vspace{-0.6cm}}
   \label{fig:aggregation}
\end{figure}

\cref{fig:aggregation} shows the accuracy of the ImageNet classification task depending on the size of aggregation kernels using $3 \times 3$ CA kernels and 16 channels in each head.  A large base kernel provided better accuracy, but it got saturated, similar to the DW convolution's locality.  The properties of locality biases are similar among all spatial invariance; it only gets better with relaxed spatial invariance.  In this experiment, we used the same kernel sizes for all layers even though the smaller convolutional kernels are preferable in earlier layers \cite{tan2019mixconv}.  In other words, the local biases are different in each layer.   So, to have better accuracy, we should consider different kernel sizes in each layer.

\subsection{Downsampling filter}\label{sec:ExperimentDownsampling}
\begin{figure}[t]
\captionsetup[subfigure]{justification=centering}
  \centering
  \begin{subfigure}[b]{0.245\linewidth}
    \includegraphics[width=\linewidth]{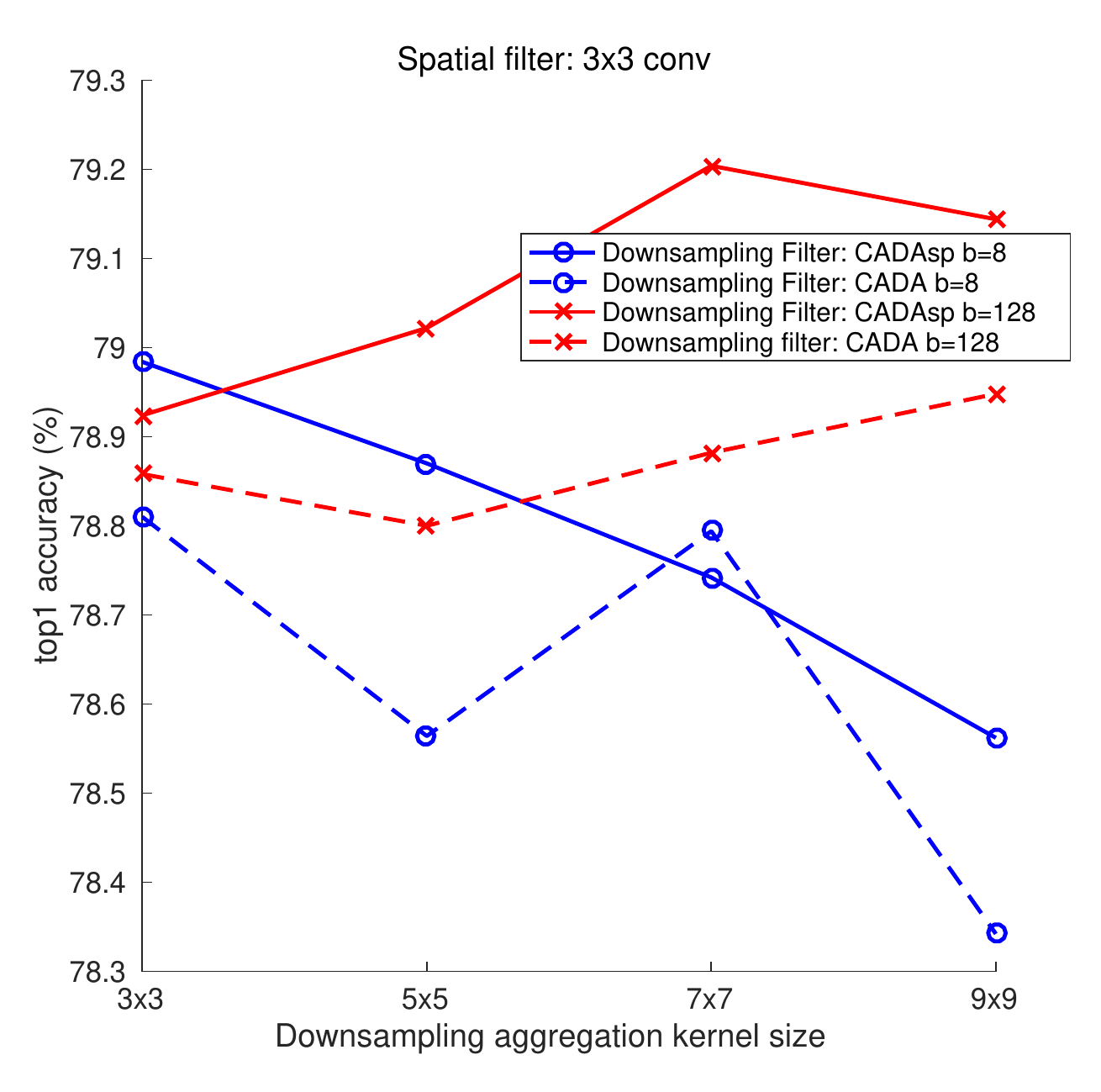}
    \caption{SF: $3 \times 3$ conv}
  \end{subfigure}
  \begin{subfigure}[b]{0.245\linewidth}
    \includegraphics[width=\linewidth]{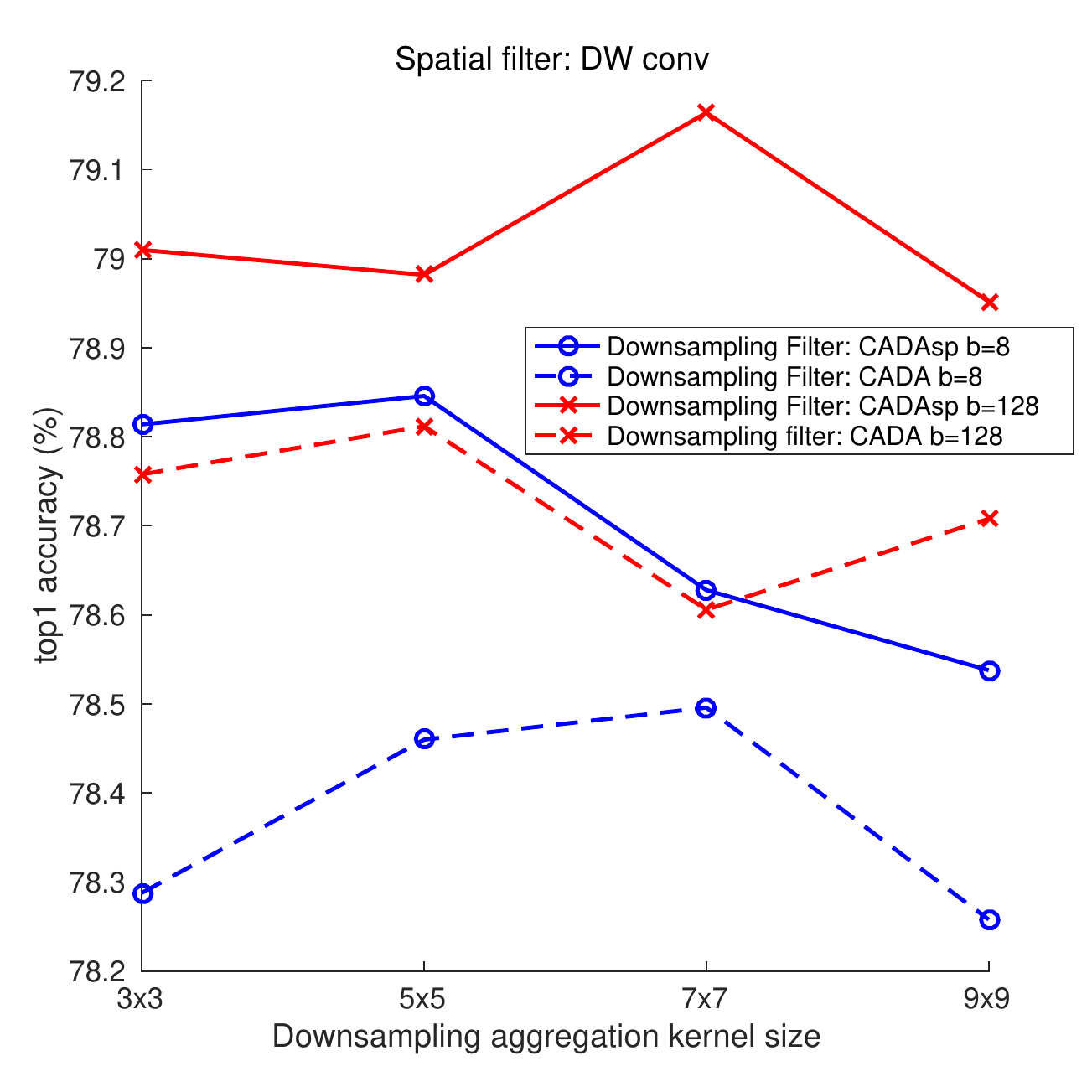}
    \caption{SF: $7 \times 7$ DW conv} 
  \end{subfigure}
  \begin{subfigure}[b]{0.245\linewidth}
    \includegraphics[width=\linewidth]{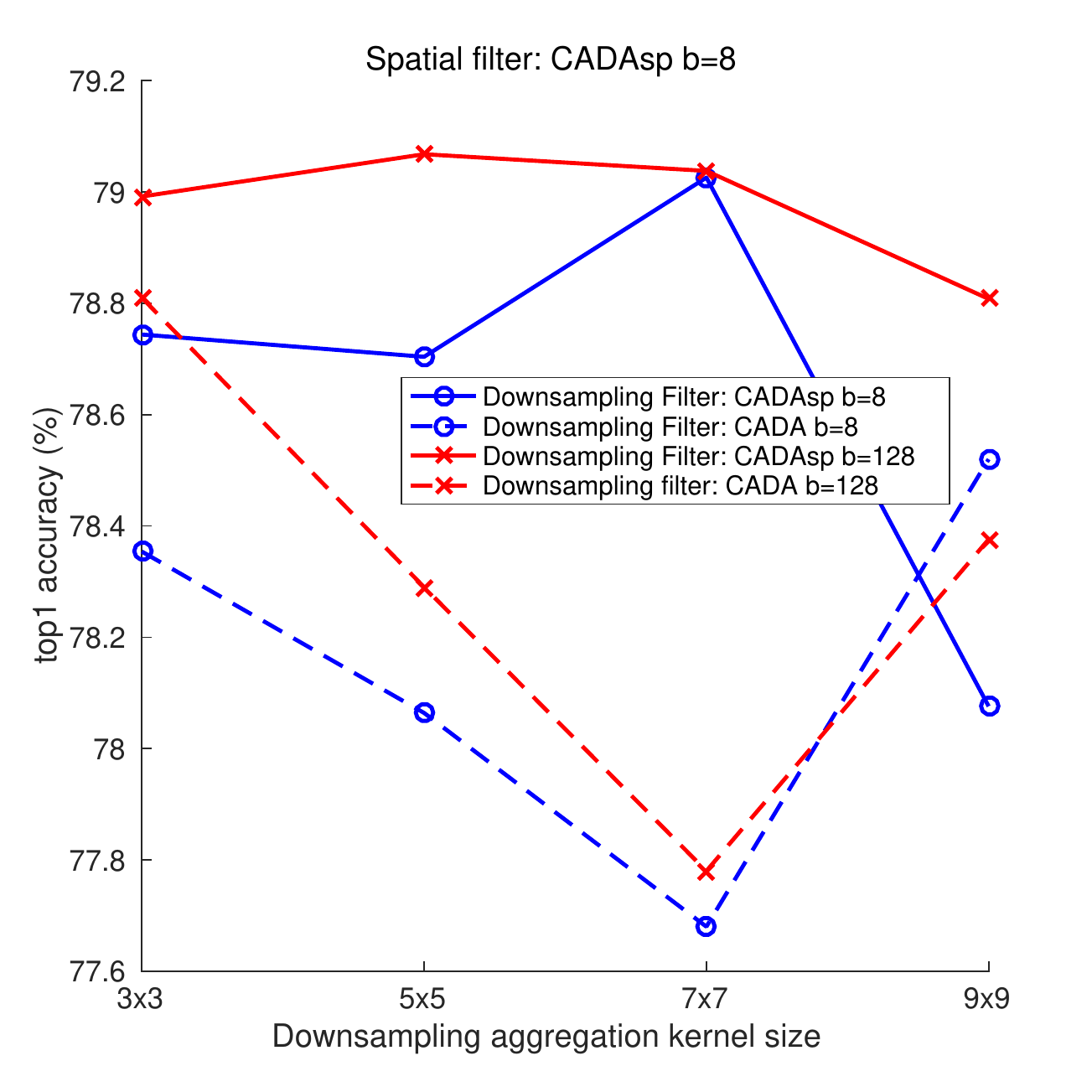}
    \caption{SF: CADAsp b=8} 
  \end{subfigure}
  \begin{subfigure}[b]{0.245\linewidth}
    \includegraphics[width=\linewidth]{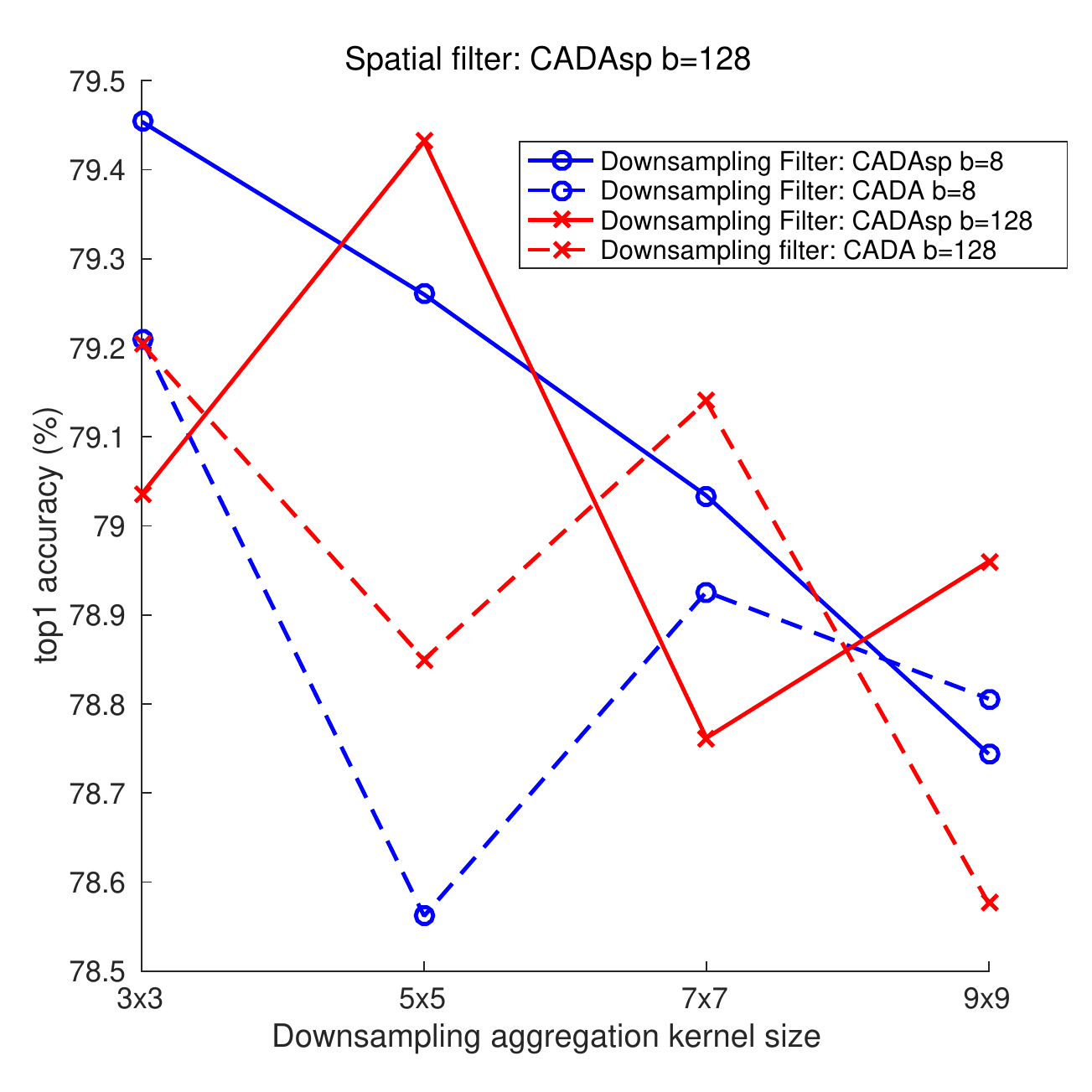}
    \caption{SF: CADAsp b=128} 
  \end{subfigure}
  \caption{Accuracy of the ImageNet classification task depends on the aggregation kernel size of the CADA and CADAsp downsampling filters using $3 \times 3$ CA kernels and 16 channels in each head.  "$b$" represents the number of base kernels.  Each figure shows experiments with different spatial filters (SF) represented in each caption.  The CADAsp SF use $3 \times 3$ CA kernels, $7 \times 7$ aggregation kernels, and 16 channels in each head.}
  \label{fig:CADA_VS_CADAsp}
\end{figure}

  We experimented with the CADA, and CADAsp downsampling filters using several spatial filters, as shown in \cref{fig:CADA_VS_CADAsp}.  CADAsp generally gave better results, so we used CADAsp to study inductive biases in this section.  Also, we chose to use the same number of base kernels for all layers.
  
  We conducted ablation studies on four downsampling filters between stages, so the additional computation cost is significantly smaller than the modification of the spatial filters.  

\subsubsection{Multi head}
\begin{figure}
\begin{floatrow}

\ffigbox[\FBwidth][][]{%
  \centering
   \includegraphics[width=0.9\linewidth]{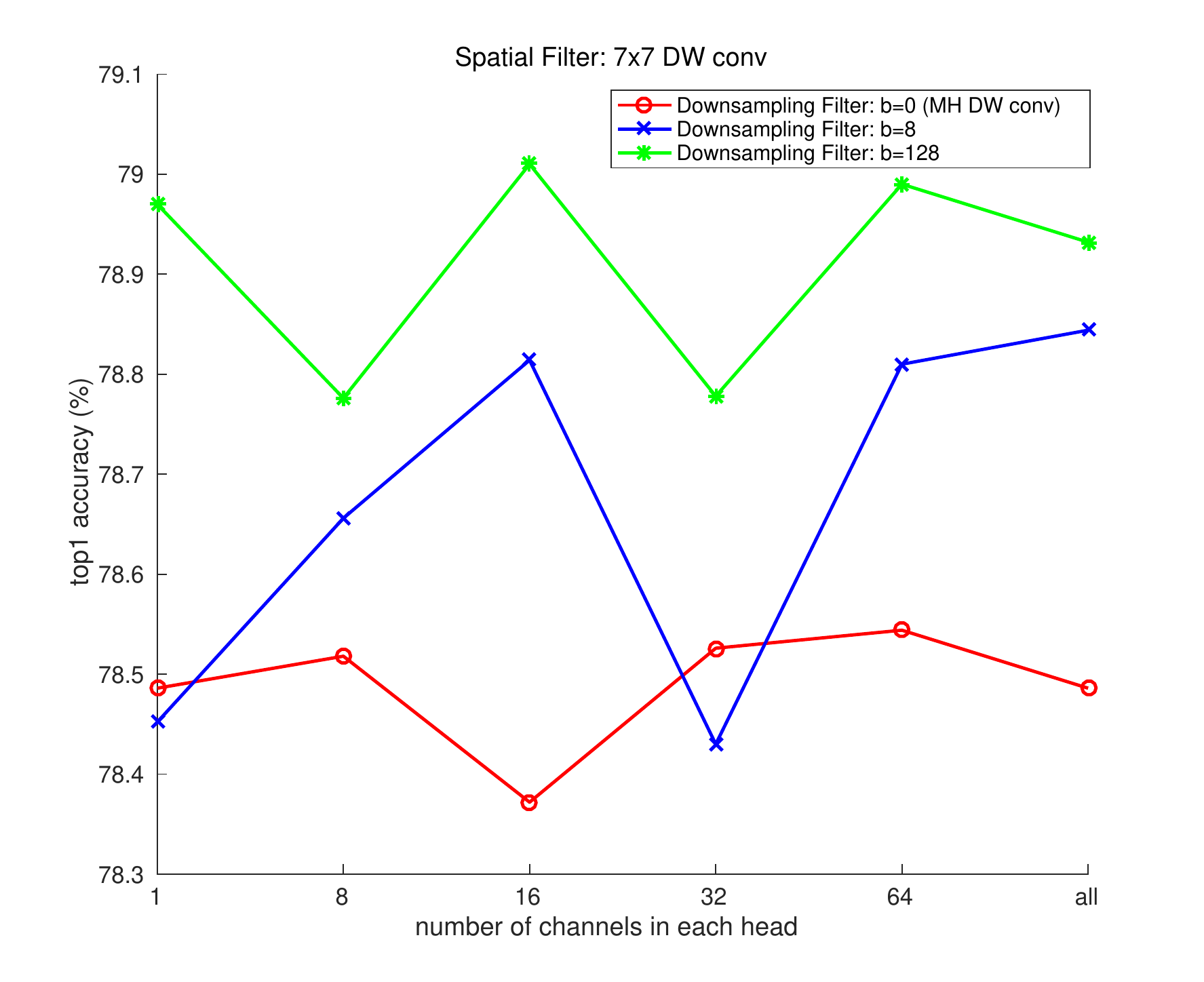}
   \label{fig:DS_head}
}{%
  \caption{Accuracy of the ImageNet classification task depends on the number of channels in each head of the $3 \times 3$ MH DW convolutional and CADAsp downsampling filter using $3 \times 3$ CA kernels and $3 \times 3$ aggregation kernels.  "$b$" represents the number of base kernels, "all" represents one head with all channels, and the spatial filters are $7 \times 7$ DW convolution.}%
  \label{Fig:DFmultiHead1}
}
\ffigbox[\FBwidth][][]{%
  \centering
   \includegraphics[width=0.9\linewidth]{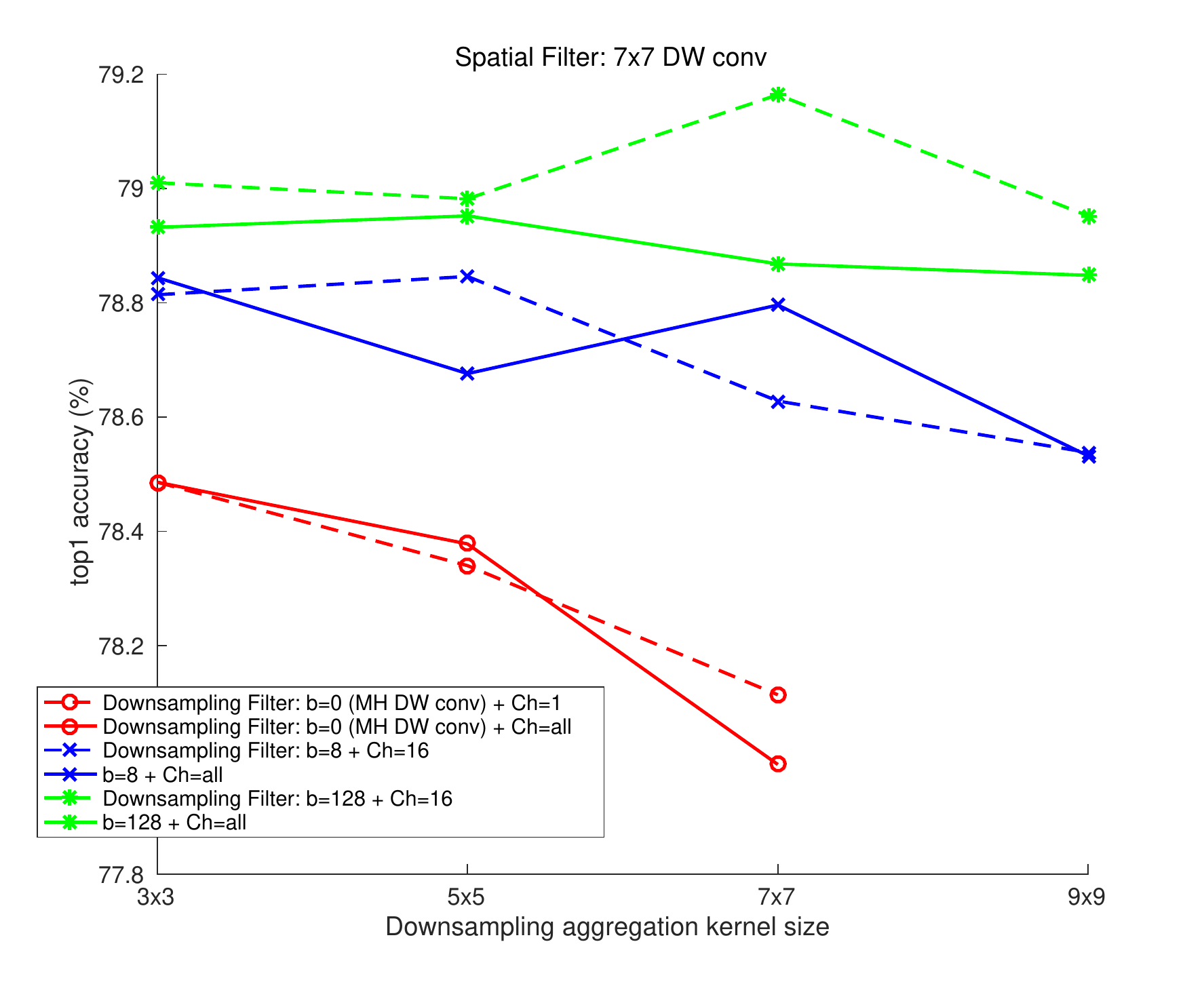}
}{%
  \caption{Accuracy of the ImageNet classification task depends on the aggregation kernel size of the $3 \times 3$ MH DW convolutional and CADAsp downsampling filters using $3 \times 3$ CA kernels.  "$b$" represents the number of base kernels.  $C_h$ represents the number of channels in each head, and "all" represents one head with all channels.  The spatial filters are $7 \times 7$ DW convolution.}%
  \label{Fig:DFmultiHead2}
}
\end{floatrow}
\end{figure}

\begin{table}[h]
  \caption{Accuracy of the ImageNet classification task for different head sizes.  The CADAsp spatial filters use $3 \times 3$ CA kernels, $7 \times 7$ aggregation kernels, and 16 channels in each head.  The DW convolutional spatial filters also use $7\times 7$ aggregation kernels.  The CADAsp downsampling filters (DF) use $3 \times 3$ CA kernels and $3 \times 3$ aggregation kernels. The DW convolutional downsampling filters also use $3\times 3$ aggregation kernels.  "$b$" represents the number of base kernels, and "all" represents one head with all channels.}
  \centering
  \setlength{\tabcolsep}{5pt} 
  \begin{tabular}{lcc|c|cc}
    \toprule
    & & & \multicolumn{3}{c}{Spatial Filters}                   \\
    & & & \multirow{2}{*}{DW conv} & \multicolumn{2}{c}{CADAsp}\\
    & & & & $b=8$ & $b=128$\\
    \midrule
    DF & $b$ & number of channels in a head & & &\\
    \midrule
    \multirow{4}{*}{CADAsp} & \multirow{2}{*}{8} & all & 78.84 & 79.13 & 78.99\\
    & & 16 & 78.81 & 78.74 & 79.45 \\
    \cmidrule(r){2-6}
     & \multirow{2}{*}{128} & all & 78.93 & 79.15 & 79.57\\
    & & 16  & 79.01 & 78.99 & 79.04 \\
    \midrule
    \multirow{2}{*}{DW conv} & & all & 78.49 & 78.67 & 79.17\\
     & & 1  & 78.49 & 78.66 & 79.17\\

    \bottomrule
  \end{tabular}
  \label{tab:DFmultiHead}
\end{table}

To study a preferable number of heads for downsampling filters, we first tested with $7 \times 7$ DW convolutional spatial filters.
\cref{Fig:DFmultiHead1} shows the accuracies of the ImageNet classification task depending on the number of channels in each head of $3 \times 3$ MH DW convolutional and CADAsp downsampling filters using $3 \times 3$ CA kernels and $3 \times 3$ aggregation kernels.  The accuracy does not decline as the number of channels increases in each head, contrary to the spatial filters, as shown in \cref{fig:Ch}.  \cref{Fig:DFmultiHead2} shows the accuracy of the ImageNet classification task depending on the aggregation kernel size of the MH DW convolutional and CADAsp downsampling filter using $3 \times 3$ CA kernels.  The accuracy is almost equivalent to single and multiple heads.  

We extended these experiments with the CADAsp spatial filters using $3 \times 3$ CA kernels, $7 \times 7$ aggregation kernels, and 16 channels in each head.  We used the same downsampling filters from prior experiments with different numbers of heads, as shown in \cref{tab:DFmultiHead}, which again shows no significant difference with single and multiple heads.  Thus, for the downsampling filters, having multiple heads is not important.  Considering the space complexity, we should choose a single head.

\subsubsection{Spatial invariance}
\begin{figure}[t]
\captionsetup[subfigure]{justification=centering}
  \centering
  \begin{subfigure}[b]{0.24\linewidth}
    \includegraphics[width=\linewidth]{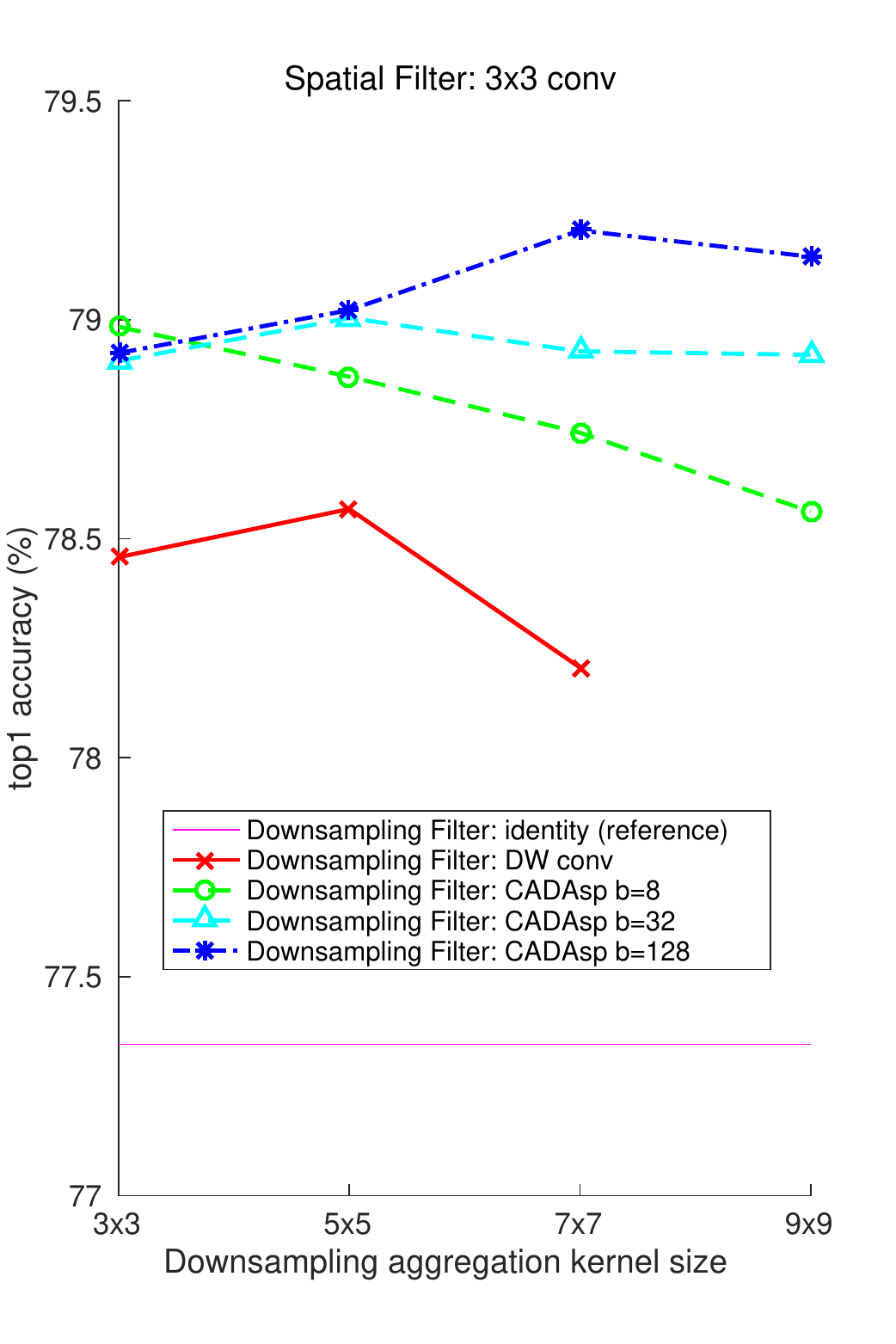}
    \caption{SF: $3 \times 3$ conv}
  \end{subfigure}
  \begin{subfigure}[b]{0.24\linewidth}
    \includegraphics[width=\linewidth]{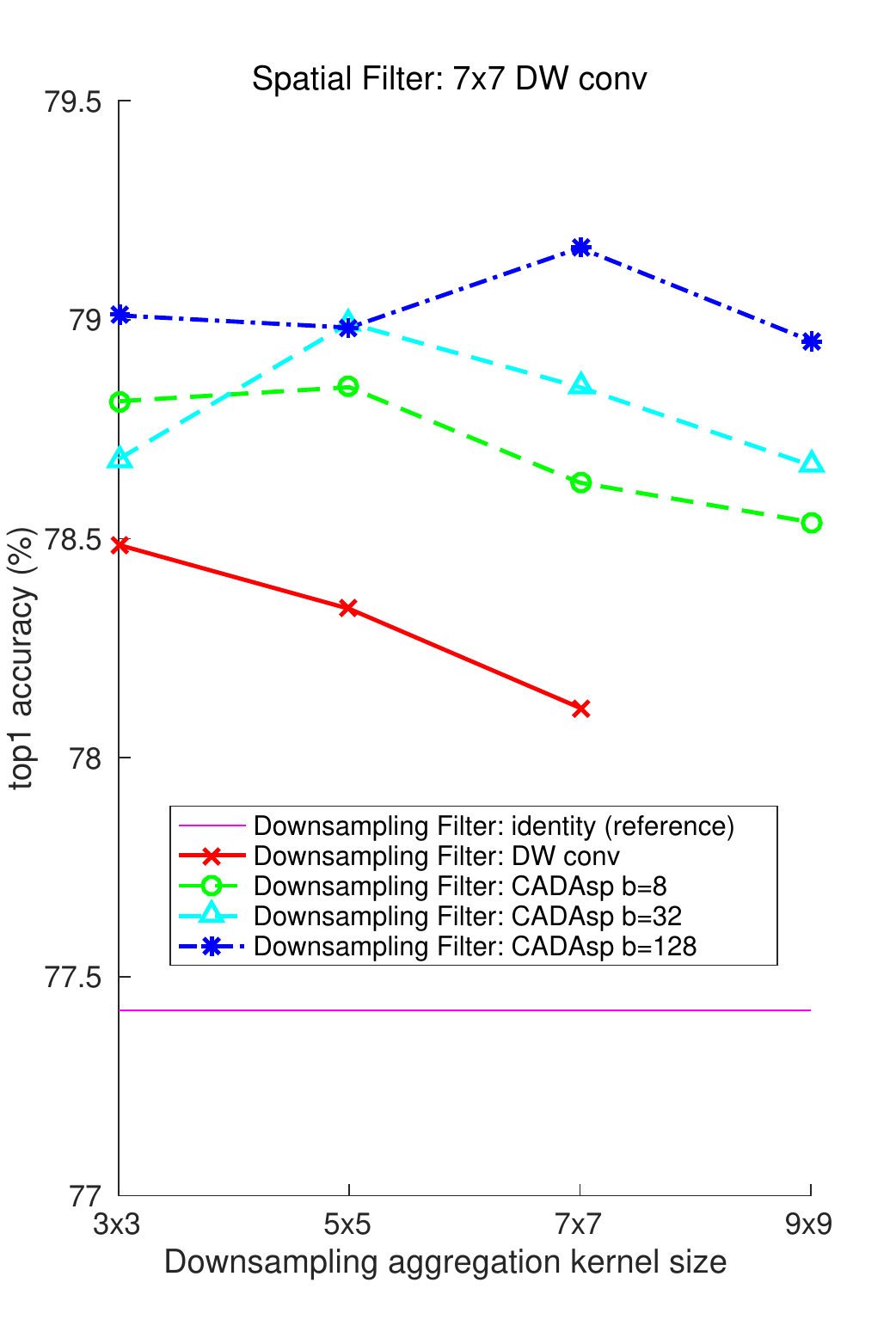}
    \caption{SF: $7 \times 7$ DW conv} 
  \end{subfigure}
  \begin{subfigure}[b]{0.24\linewidth}
    \includegraphics[width=\linewidth]{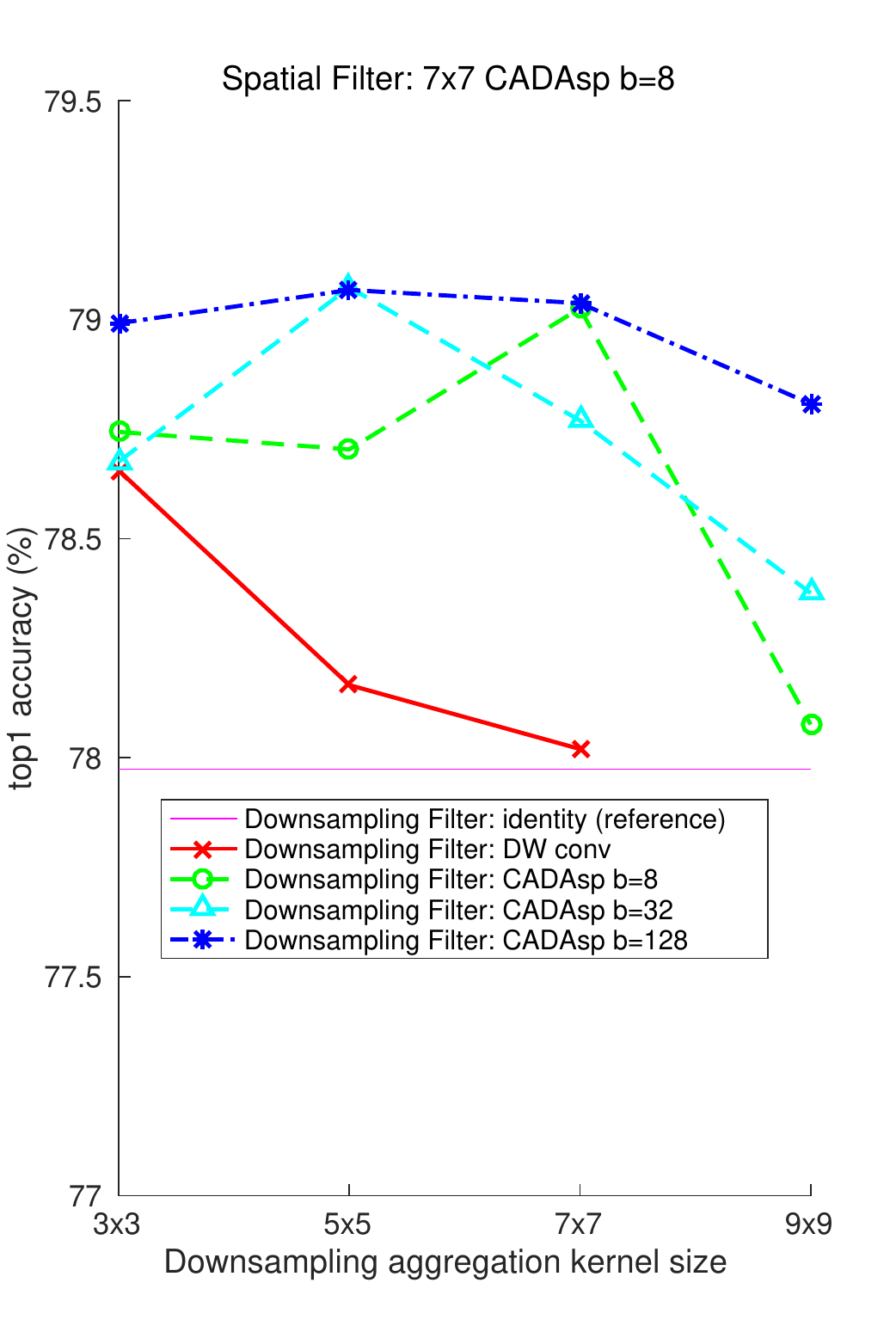}
    \caption{SF: CADAsp b=8} 
  \end{subfigure}
  \begin{subfigure}[b]{0.24\linewidth}
    \includegraphics[width=\linewidth]{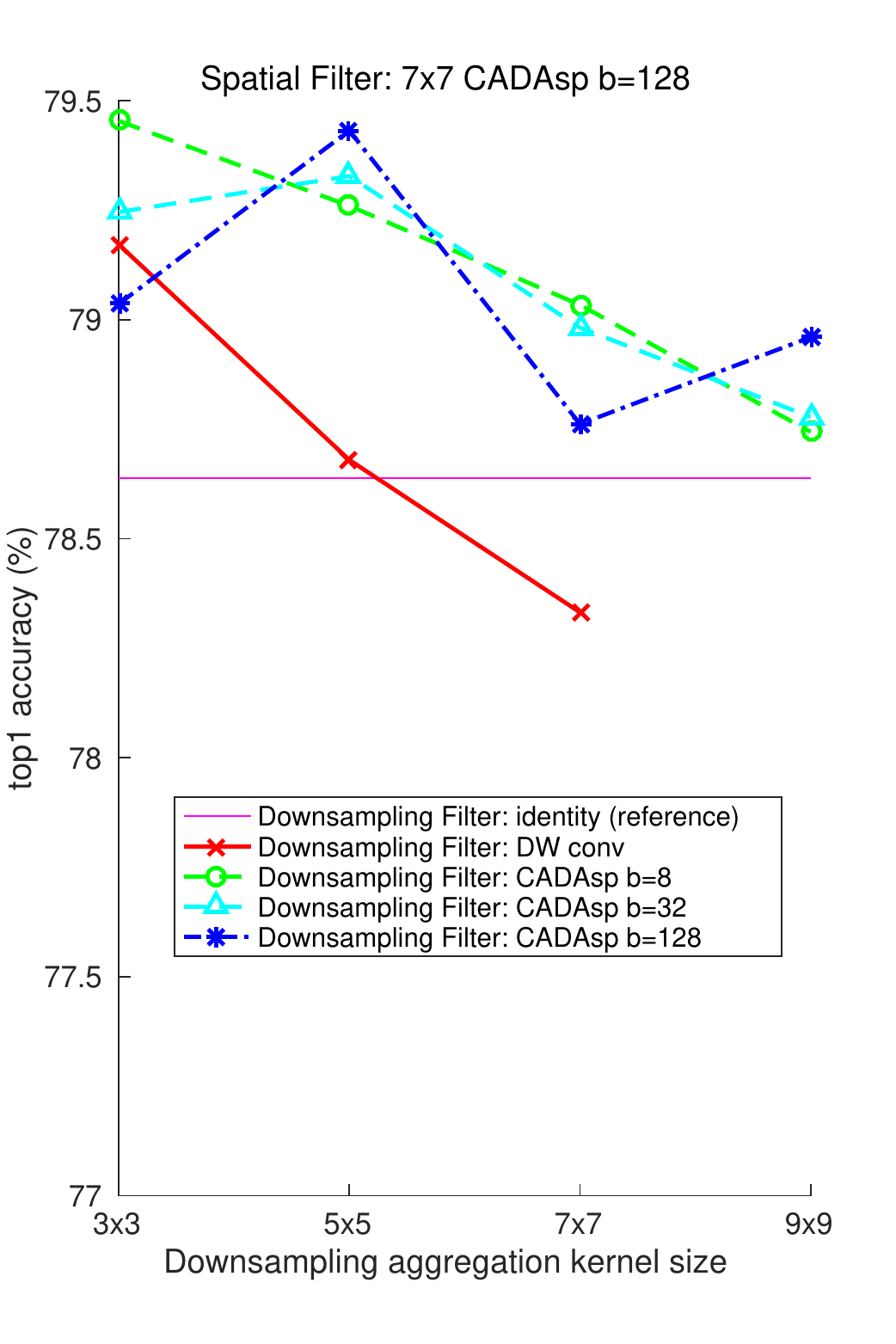}
    \caption{SF: CADAsp b=128} 
  \end{subfigure}
  \caption{Accuracy of the ImageNet classification task depends on the aggregation kernel size of the DW convolutional and CADAsp downsampling filters.  The CADAsp downsampling filters use $3 \times 3$ CA kernels and 16 channels in each head.  "b" represents the number of base kernels.   Each subfigure shows a different spatial filter (SF).  The CADAsp spatial filters use $3 \times 3$ CA kernels, $7 \times 7$ aggregation kernels, and 16 channels in each head.  }
 
  \label{fig:SF_DF}
\end{figure}

We tested spatial invariance inductive bias in the downsampling filters using the different number of base kernels in the CADAsp with different spatial filters, as shown in \cref{fig:SF_DF}.   The straight line shows the accuracy without the downsampling filters.  As expected from \cref{sec:spatial}, the accuracy can be improved by relaxing the spatial invariance bias of spatial filters.  We used the CADAsp spatial filters using $3 \times 3$ CA kernels, $7 \times 7$ aggregation kernels, 16 channels in each head, and 8 or 128 base kernels.  Any additional downsampling filters with $3 \times 3$ aggregation kernels further improved accuracy.  We used the CADAsp downsampling filters with $3 \times 3$ CA kernels and 16 channels in each head.  Relaxing spatial invariance increase accuracy in general, as shown in each subfigure.   However, the difference becomes less significant as spatial invariance in the spatial filters gets relaxed. 

\subsubsection{Locality}

\begin{figure}
\begin{floatrow}

\ffigbox[\FBwidth][][]{%
  \centering
    \includegraphics[width=0.9\linewidth]{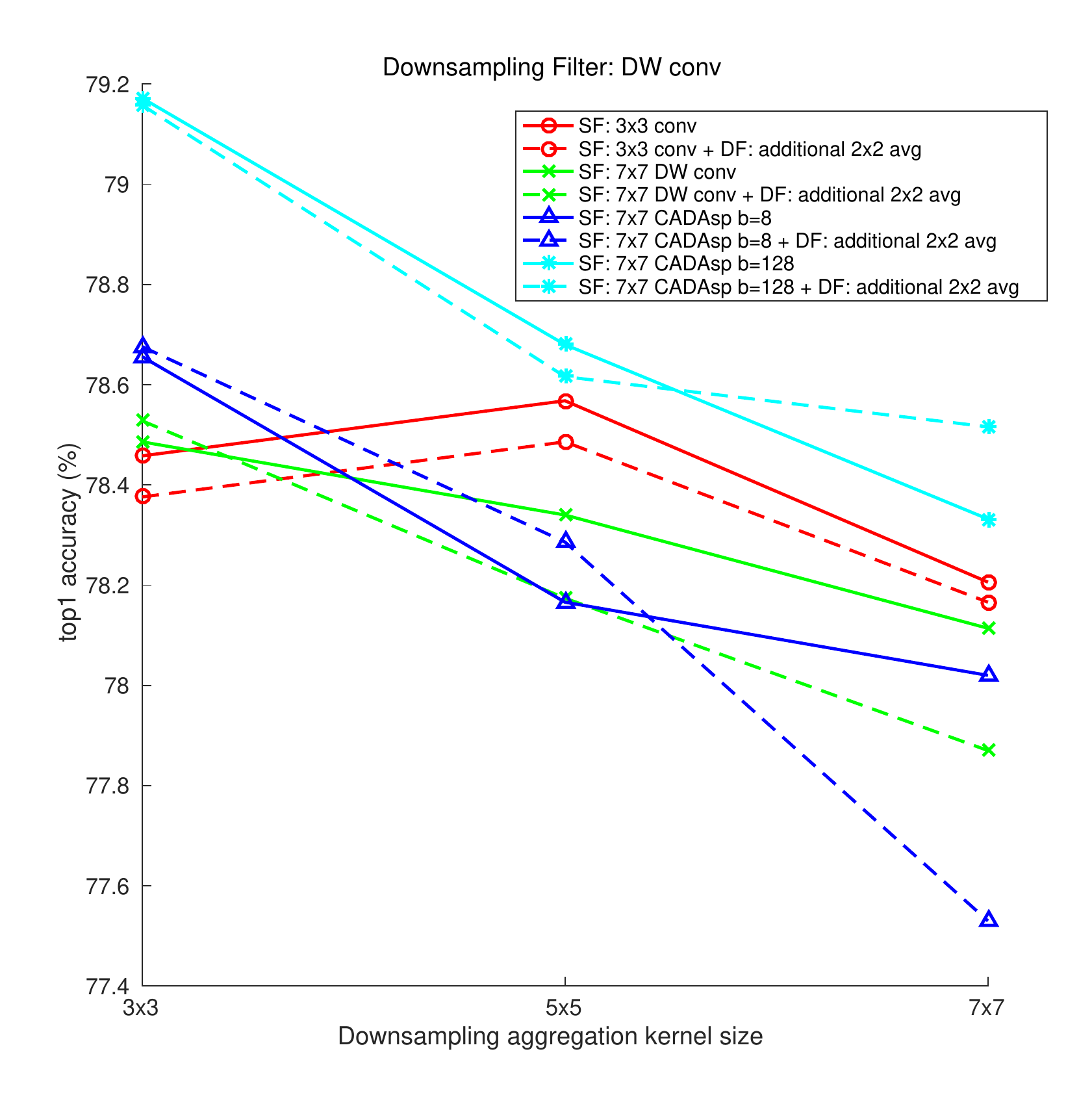}
}{%
  \caption{Accuracy of the ImageNet classification task depends on the aggregation kernel size of the DW convolution downsampling filters (DF) with and without averaging filters.  The CADAsp spatial filters (SF) use $3 \times 3$ CA kernels $7 \times 7$ aggregation kernels and 16 channels in each head.  "$b$" represents the number of base kernels.
  }
  \label{fig:avgDown}
}
\ffigbox[\FBwidth][][]{%
  \centering
    \includegraphics[width=0.9\linewidth]{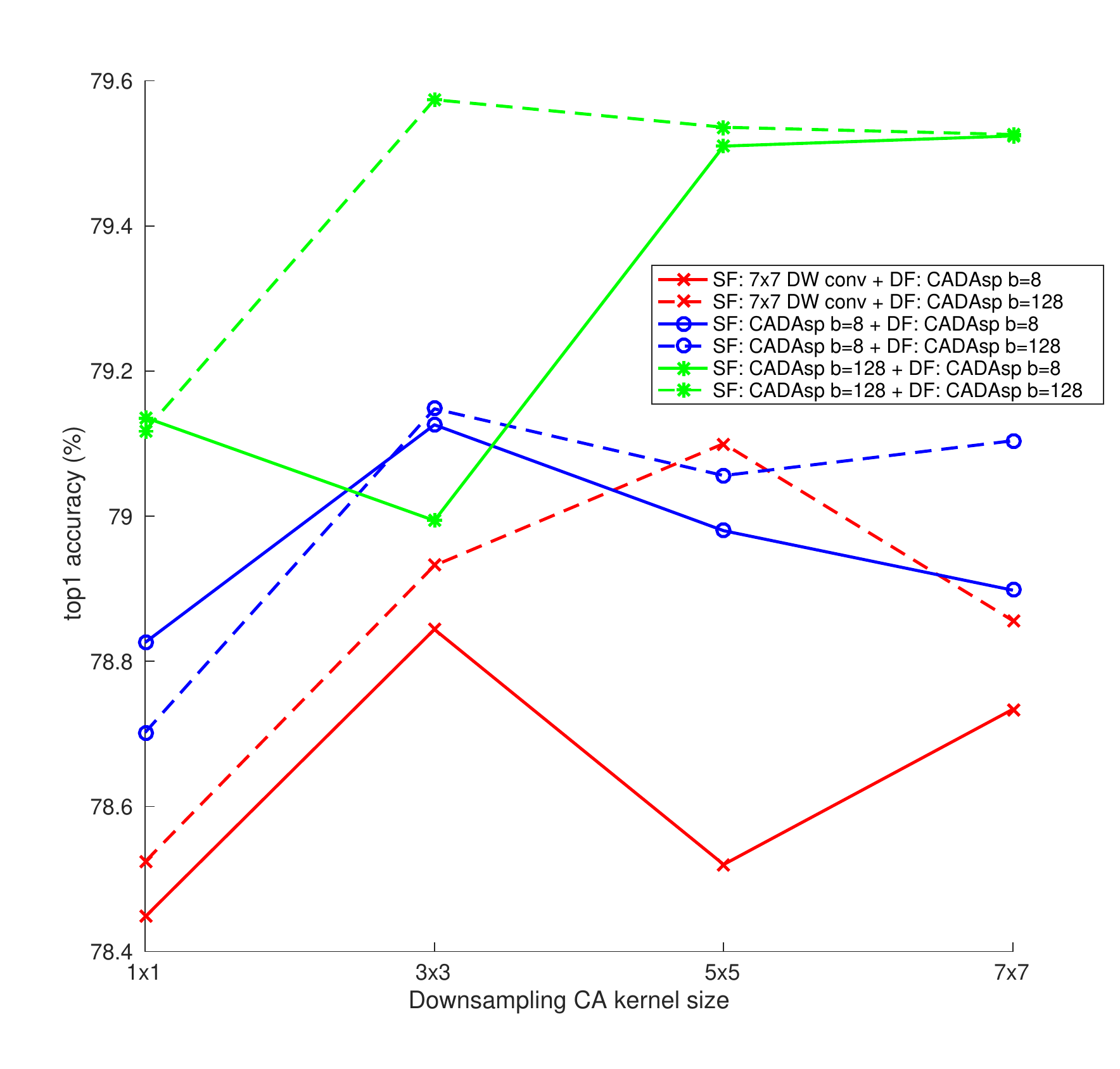}
}{%
  \caption{Accuracy of the ImageNet classification task depends on the CA kernel size of the CADAsp downsampling filters (DF). The CADAsp downsampling filters use $3 \times 3$ aggregation kernels, and one head.  "$b$" represents the number of base kernels.  The CADAsp spatial filters (SF) use $3 \times 3$ CA kernels, $7 \times 7$ aggregation kernels, and 16 channels in each head.  }
  \label{fig:attention}
}
\end{floatrow}
\end{figure}

The DW convolutional downsampling filter requires a strong locality bias, as shown in \cref{fig:PostAct} from Appendix.  No implicit low-pass filter might cause this, so we added $2 \times 2$ averaging filter before DW convolution, which gave no significant difference, as shown in \cref{fig:avgDown}.

First, we tested the locality bias of context awareness in the CA kernels of the CADAsp downsampling filters, using $3 \times 3$ aggregation kernels, one head, and 8 or 128 base kernels, as shown in \cref{fig:attention}.  We experimented with the spatial filters of $7 \times 7$ DW convolution and CADAsp using $3 \times 3$ CA kernels, $7 \times 7$ aggregation kernels, 16 channels in each head, and 8 or 128 base kernels.  Similar to \cref{fig:CAkernel}, the accuracy is quickly saturated, so we only require local information for context awareness. 

\begin{figure}[t]
\captionsetup[subfigure]{justification=centering}
  \centering
  \begin{subfigure}[b]{0.24\linewidth}
    \includegraphics[width=\linewidth]{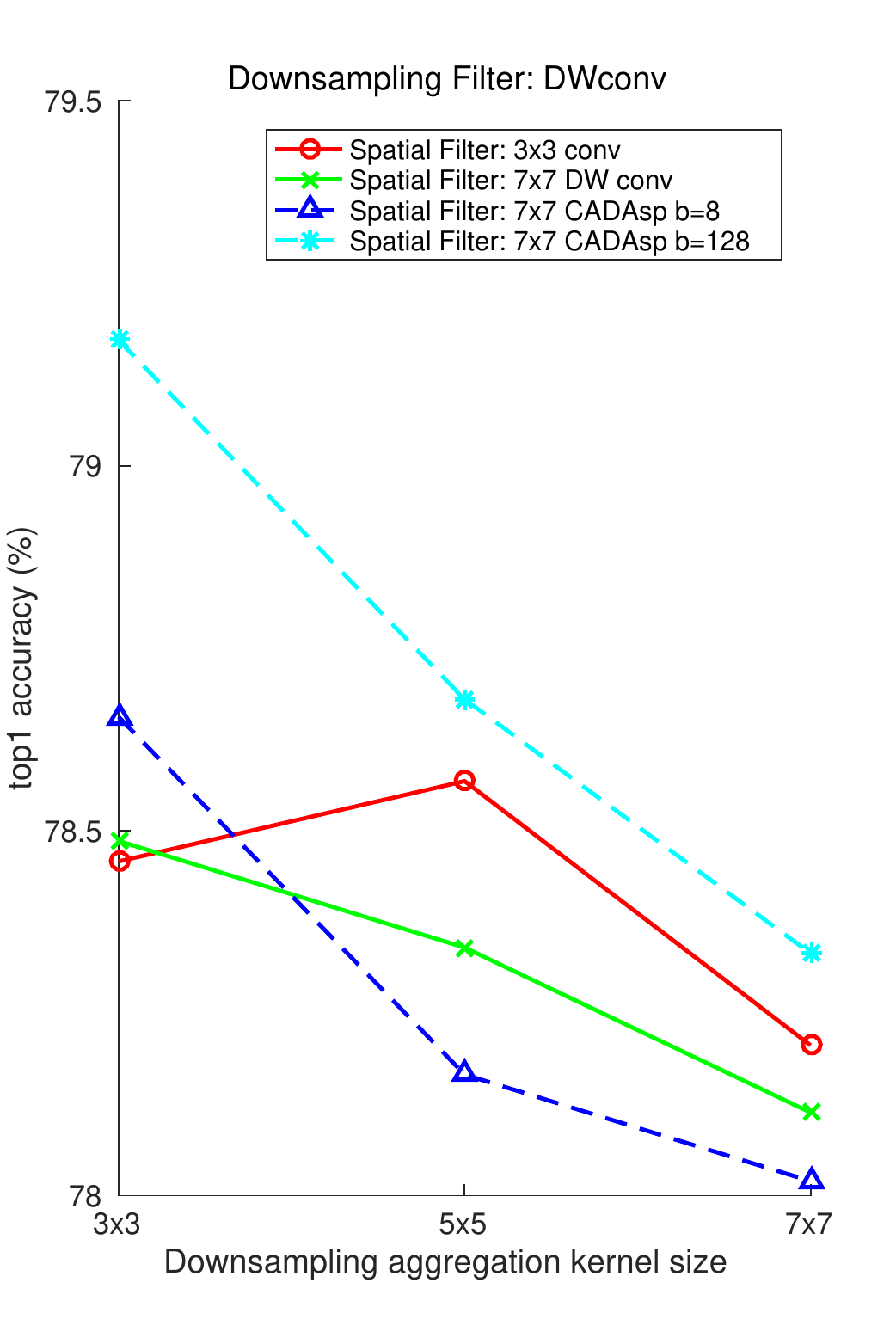}
    \caption{DF: DW conv}\label{SF_DF_DWconv}
  \end{subfigure}
  \begin{subfigure}[b]{0.24\linewidth}
    \includegraphics[width=\linewidth]{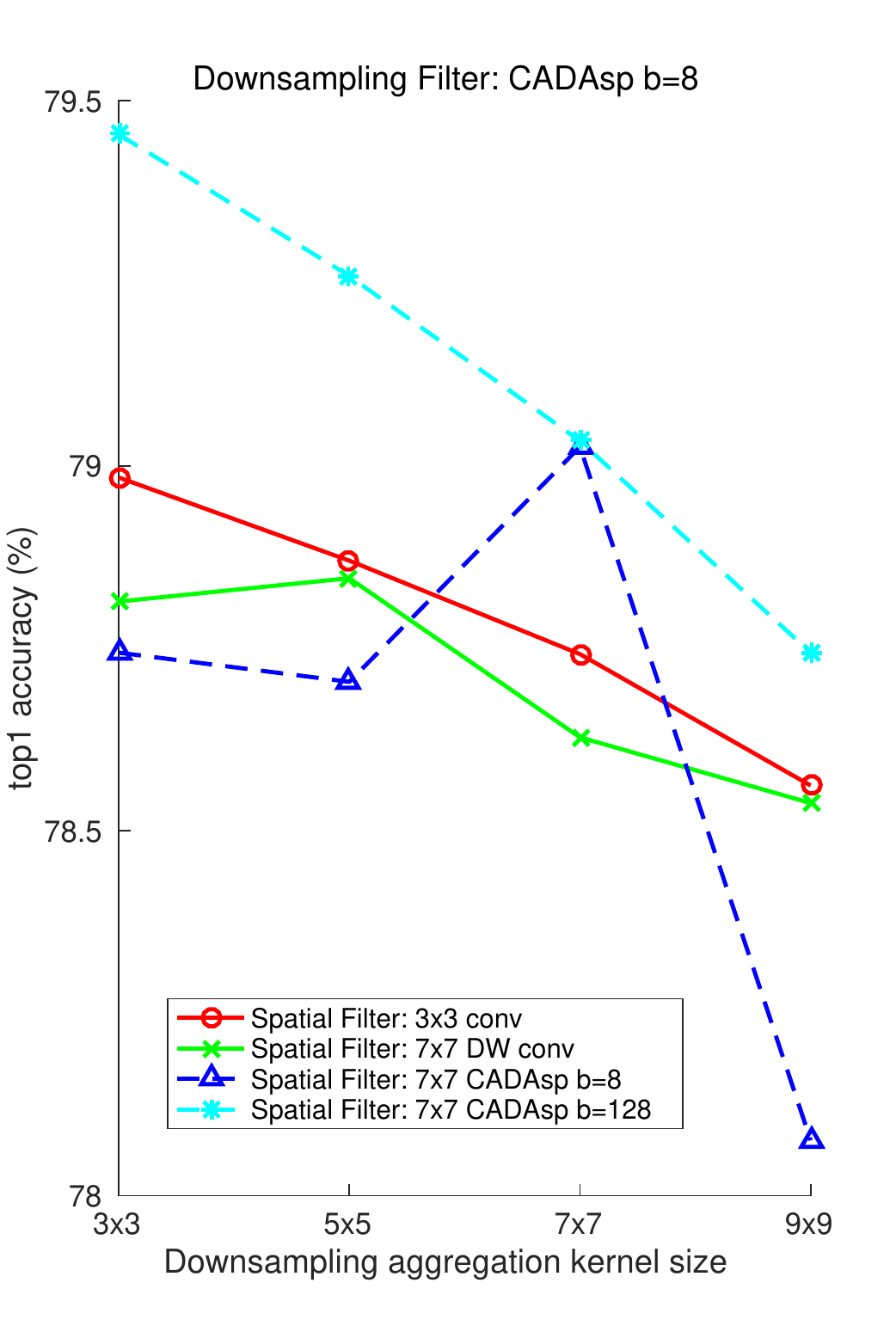}
    \caption{DF: CADAsp b=8} \label{SF_DF_CADA8}
  \end{subfigure}
  \begin{subfigure}[b]{0.24\linewidth}
    \includegraphics[width=\linewidth]{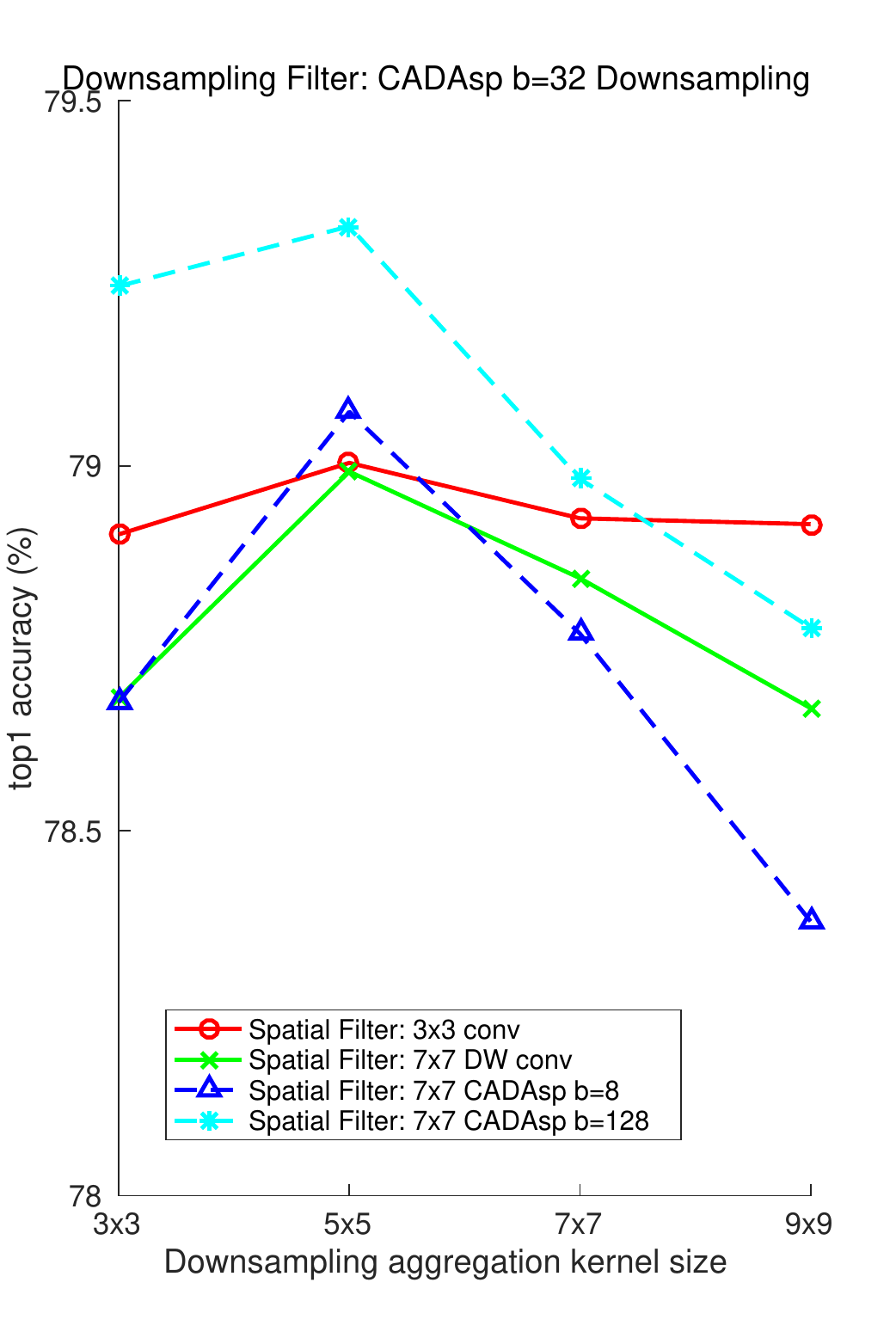}
    \caption{DF: CADAsp b=32} \label{SF_DF_CADA32}
  \end{subfigure}
  \begin{subfigure}[b]{0.24\linewidth}
    \includegraphics[width=\linewidth]{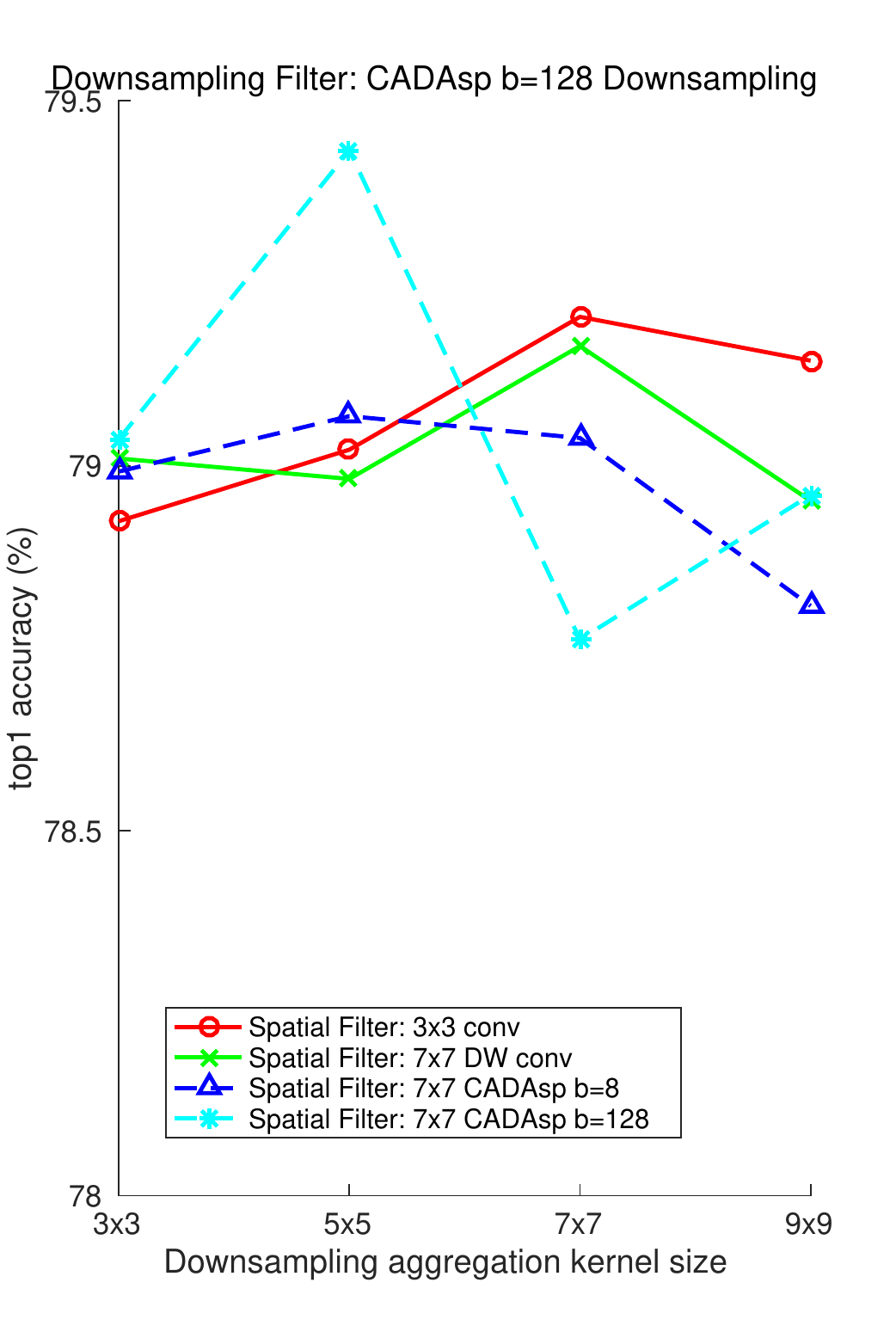}
    \caption{DF: CADAsp b=128} \label{SF_DF_CADA128}
  \end{subfigure}
  \caption{Accuracy of the ImageNet classification task depends on the aggregation kernel size of the DW convolution and CADAsp downsampling filters (DF).  Each subfigure shows a different DF.  CADAsp DF use $3 \times 3$ CA kernels and 16 channels in each head.  "$b$" represents the number of base kernels.   The CADAsp spatial filters use $3 \times 3$ CA kernels, $7 \times 7$ aggregation kernels, and 16 channels in each head.  }
  \label{fig:SF_DF2}
\end{figure}

Second, we tested locality bias in the aggregation kernels of the downsampling filters.  \cref{fig:SF_DF2} shows the same results as \cref{fig:SF_DF} with different groupings.  \cref{SF_DF_DWconv} shows the accuracy of the DW convolutional downsampling filter.  As with  \cref{fig:PostAct} in the Appendix, locality bias is crucial.  \cref{SF_DF_CADA8}, \cref{SF_DF_CADA32}, and \cref{SF_DF_CADA128} show the accuracy of the CADAsp downsampling filters with different base kernels.  As spatial invariance is relaxed, the locality bias is also relaxed; however, larger kernels do not improve accuracy, contrary to the locality bias of the spatial filters, as shown in \cref{fig:aggregation}.  

\begin{figure}[t]
\captionsetup[subfigure]{justification=centering}
  \centering
  \begin{subfigure}[b]{0.325\linewidth}
    \includegraphics[width=\linewidth]{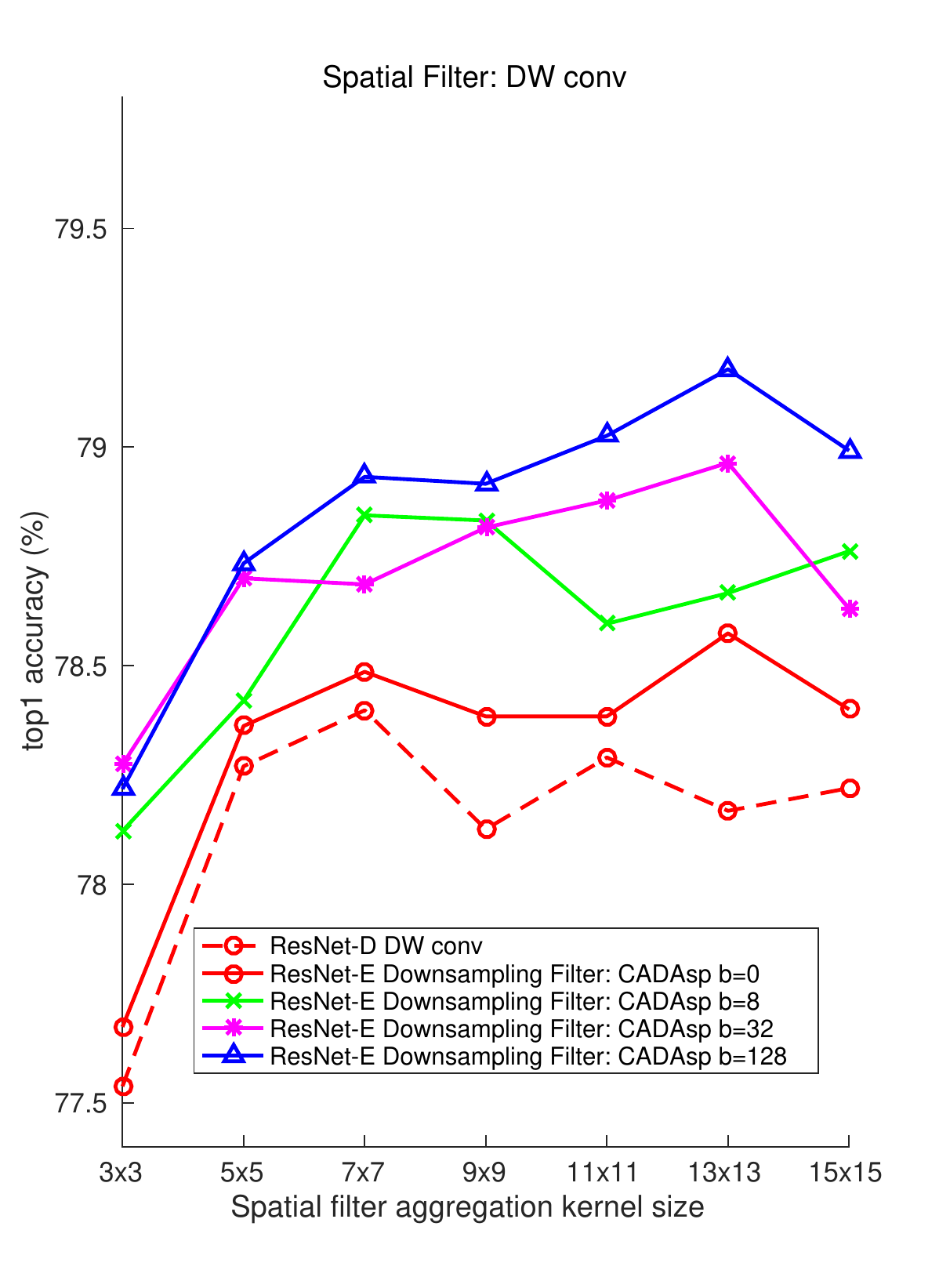}
    \caption{SF: DW convolution}\label{fig:SF_DF3_1}
  \end{subfigure}
  \begin{subfigure}[b]{0.325\linewidth}
    \includegraphics[width=\linewidth]{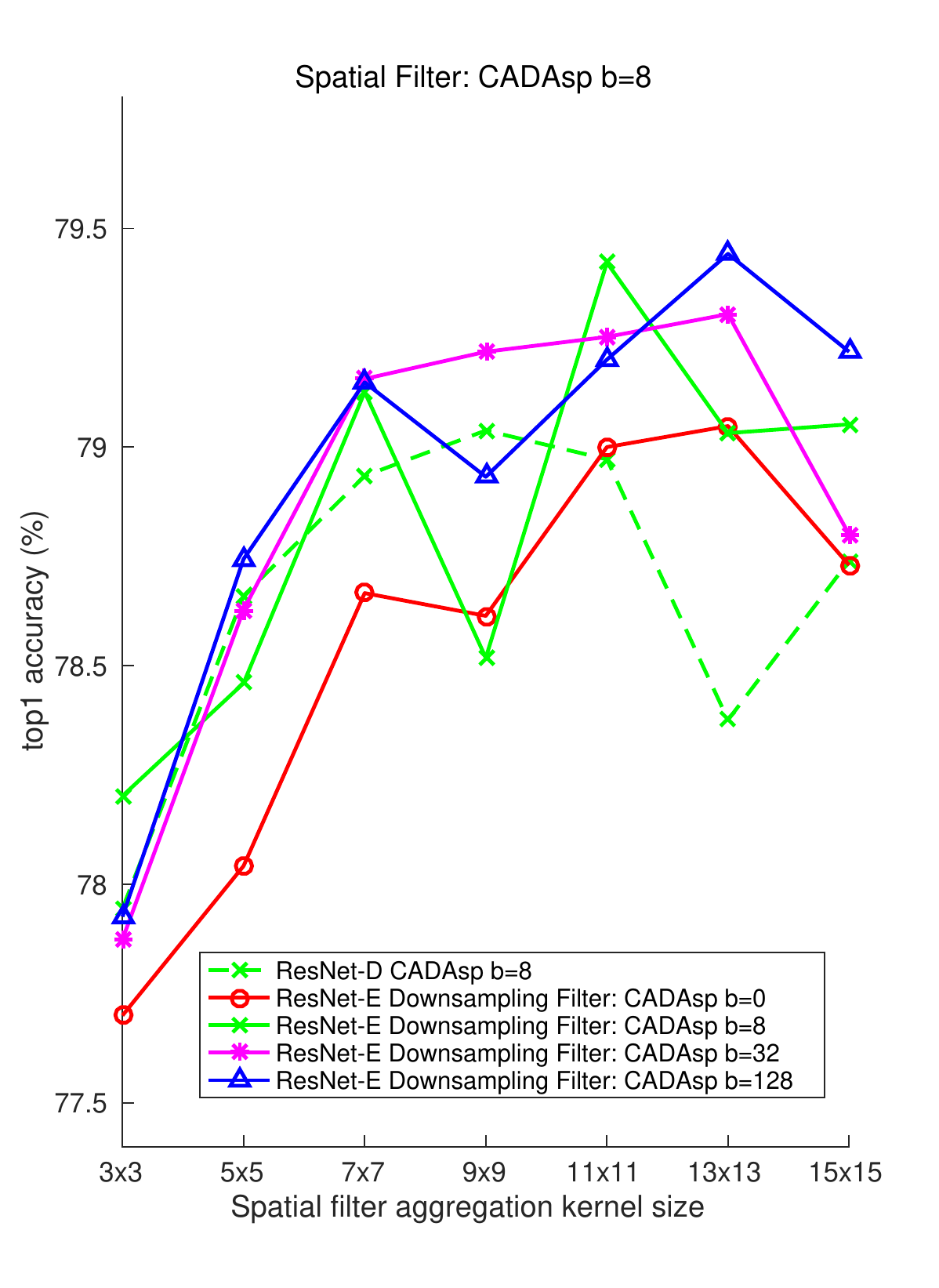}
    \caption{SF: CADAsp b=8} \label{fig:SF_DF3_2}
  \end{subfigure}
  \begin{subfigure}[b]{0.325\linewidth}
    \includegraphics[width=\linewidth]{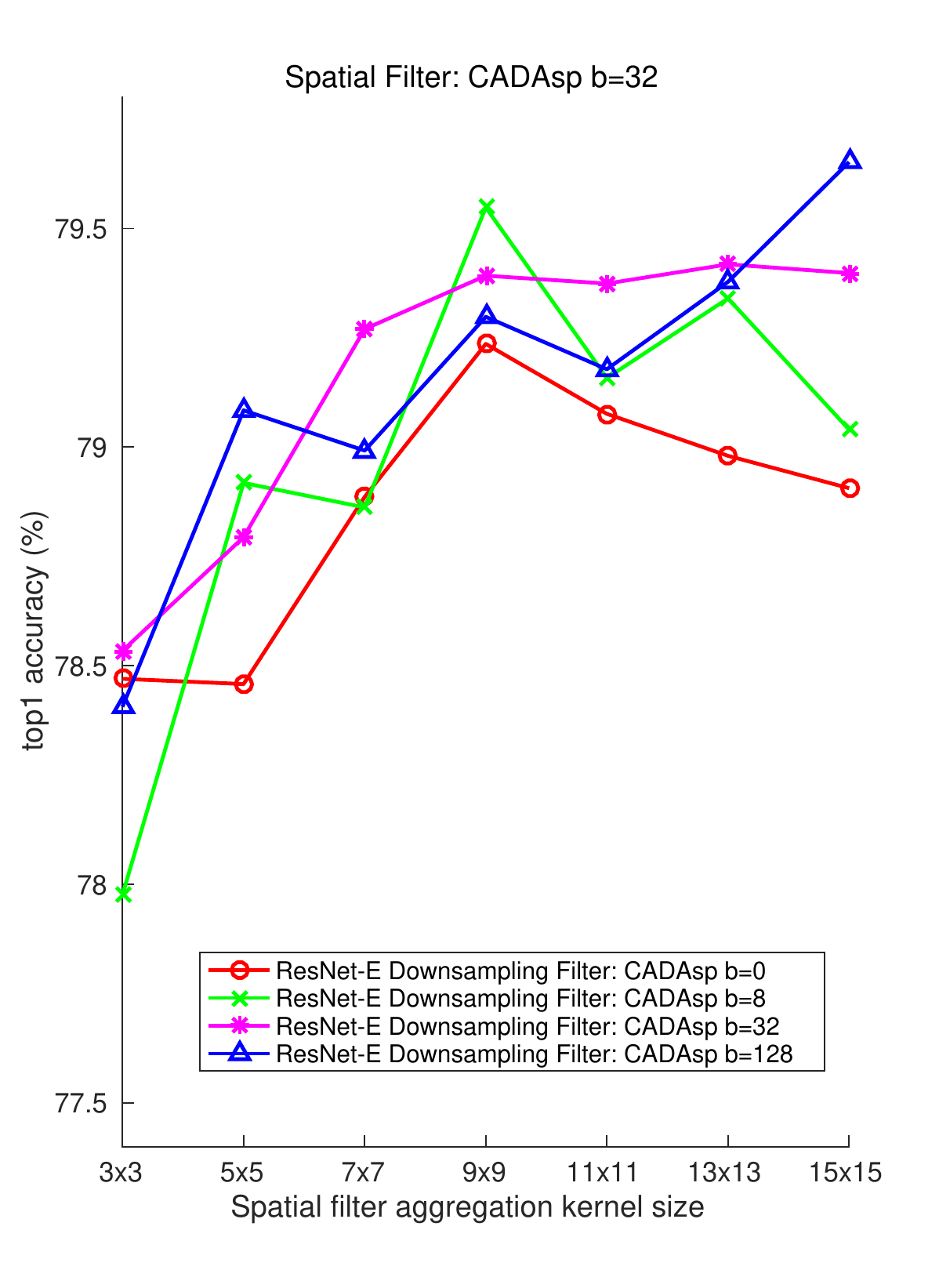}
    \caption{SF: CADAsp b=128} 
  \end{subfigure}
  \caption{Accuracy of the ImageNet classification task depends on the aggregation kernel size of the spatial filters (SF).  The CADAsp SF use $3 \times 3$ CA kernels and 16 channels in each head.  "b" represents the number of base kernels.  The CADAsp downsampling filters use $3 \times 3$ CA kernels, $3 \times 3$ aggregation kernels, and one head.  }
  \label{fig:SF_DF3}
\end{figure}

Finally, we checked whether setting a CADAsp downsampling filter to a $3 \times 3$ aggregation kernel could change the nature of the locality bias of the spatial filter.  \cref{fig:SF_DF3} shows the accuracy of the ImageNet classification task depending on the size of the spatial filter's aggregation kernel.  The CADAsp spatial filters use $3 \times 3$ CA kernels and 16 channels in each head.  The CADAsp downsampling filters use $3 \times 3$ CA kernel, $3 \times 3$ aggregate kernel, and one head.  \cref{fig:SF_DF3_1} and \cref{fig:SF_DF3_2} also show the result of ResNet-D from \cref{fig:aggregation} for references.  Even with a fixed downsampling filter size, the accuracy saturates as the aggregation kernel of the spatial filter increases.

\section{Conclusion}

We introduced CADA to study inductive biases of locality and spatial invariance in locally-connected structures and determine which part of inductive biases in the local SA networks contributes to the remarkable success compared with CNNs.  We found that context awareness is an important property, but large CA kernels do not provide additional information for the classification task.  We also showed that more relaxed spatial invariance, the property of SA networks, gives better accuracy.  Even though removing spatial invariance altogether makes training more difficult.  Also, additional strong spatial invariance through relative position encoding is preferable with relaxed spatial invariance.  The locality bias through filter size is essential for downsampling filters but not for spatial filters in CADA and DW convolution.  Relaxed spatial invariance bias can mitigate the locality bias in downsampling filters, but it does not change the property of the locality bias of spatial filters.  The complete preferable setting of CADA is given in \cref{Tab:properties}.  We believe that these insights can help us understand inductive biases in locally-connected structures.




\bibliographystyle{plainnat}
\bibliography{refs}



\appendix

\section{Appendix}
\subsection{Global kernel}\label{sec:global}
\begin{wrapfigure}[7]{l}[0mm]{60mm}
  \centering 
  \vspace{-8mm}
  \includegraphics[keepaspectratio,width=60mm]
  {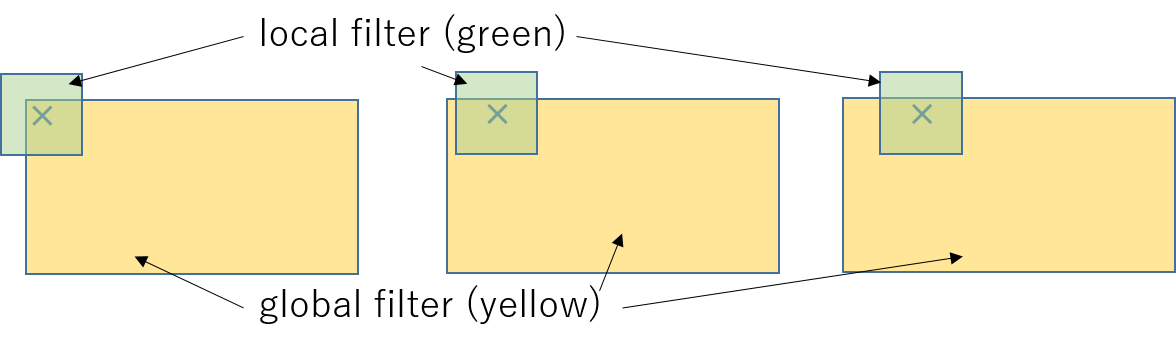}
 
  \caption{Local and global filters surrounding location $\times$}\label{fig:RT}
\end{wrapfigure}

Locally-connected filters, including local attention maps with patch initialization, could be decomposed because we assumed spatial invariant inductive biases in which the networks share local base kernels.   However, if the networks use global filters without sliding,  then the kernel of the global filters is the same everywhere, as shown in \cref{fig:RT}, which means the relative shape of the filter from the location of the interest is different everywhere.  In other words, the base kernels of these filters are also different in each location; hence we cannot study them by CADA structures. 
 
\subsection{Code} \label{sec:code}

We followed implementation from the patchwise SAN \cite{zhao2020exploring}.  To construct filters using CA network A from \cref{Fig:CAnetwork}, we used Listing \ref{code:CADA} for CADA and Listing \ref{code:CADAsp} for CADAsp.  

\begin{lstlisting}[language=Python, caption=filter construction using CA network A for CADA, escapechar=!, label={code:CADA}]
# width: input channel
# numHead: number of channels in each head
# numBase: number of base kernels

self.CAnetworkA = nn.Sequential(
    nn.Conv2d(width, (width//numHead)*(numBase), groups=width//numHead, kernel_size=attention_kernel, padding=attention_kernel//2, bias=False, stride=stride),
    norm_layer((width//numHead)*(numBase)),
    nn.ReLU(inplace=True),
    nn.Conv2d((width//numHead)*(numBase), self.aggregation_kernel_size*self.aggregation_kernel_size*(width//numHead), groups=(width//numHead), kernel_size=1))

\end{lstlisting}
\begin{lstlisting}[language=Python, caption=filter construction using CA network A for CADAsp, escapechar=!, label={code:CADAsp}]
self.CAnetworkA = nn.Sequential(
    nn.Conv2d(width, numBase, kernel_size=attention_kernel, padding=attention_kernel//2, bias=False, stride=stride),
    norm_layer(numBase),
    nn.ReLU(inplace=True),
    nn.Conv2d(numBase, self.aggregation_kernel_size*self.aggregation_kernel_size*(width//numHead), kernel_size=1))
\end{lstlisting}

For aggregation, we used Listing \ref{code:aggregation}.  

\begin{lstlisting}[language=Python, caption=aggregation, escapechar=!, label={code:aggregation}]
self.aggregation = Aggregation(kernel_size=self.aggregation_kernel_size, stride=stride, padding=self.aggregation_kernel_size//2, dilation=1, pad_mode=0)
\end{lstlisting}

To use the aggregation function, we needed to change the order of computation for CADA.  CADAsp does not need to modify the function, but it can also use the modified version.
Three lines in Listing \ref{code:aggregation2} from the original code in "aggregation\_zeropad.py" are replaced with Listing \ref{code:aggregation3}, respectively. 

\begin{lstlisting}[language=Python, caption=original code in "aggregation\_zeropad.py" to be modified for CADA, escapechar=!, label={code:aggregation2}]
const int offset_weight = ((n * ${weight_channels} + c % ${weight_channels}) * ${kernel_h} * ${kernel_w} + (kh * ${kernel_w} + kw)) * ${top_height} * ${top_width} + h * ${top_width} + w;

const int offset_weight = ((n * ${weight_channels} + c % ${weight_channels}) * ${kernel_h} * ${kernel_w} + (kh * ${kernel_w} + kw)) * ${top_height} * ${top_width} + h_out * ${top_width} + w_out;

for (int cc = c; cc < ${input_channels}; cc += ${weight_channels}) {
\end{lstlisting}

\begin{lstlisting}[language=Python, caption=modified code in "aggregation\_zeropad.py" for CADA, escapechar=!, label={code:aggregation3}]
const int offset_weight = ((n * ${weight_channels} + c * ${weight_channels}/${input_channels}) * ${kernel_h} * ${kernel_w} + (kh * ${kernel_w} + kw)) * ${top_height} * ${top_width} + h * ${top_width} + w;

const int offset_weight = ((n * ${weight_channels} + c * ${weight_channels}/${input_channels}) * ${kernel_h} * ${kernel_w} + (kh * ${kernel_w} + kw)) * ${top_height} * ${top_width} + h_out * ${top_width} + w_out;

for (int cc = c*(${input_channels}/${weight_channels}); cc < (c+1)*(${input_channels}/${weight_channels}); cc++) {
\end{lstlisting}

\subsection{ResNet-E} \label{sec:ResNet-E}
\begin{figure*}[t]
\captionsetup[subfigure]{justification=centering}
  \centering
  \hspace*{\fill}%
  \begin{subfigure}[b]{0.19\linewidth}
    \includegraphics[width=\linewidth]{ResNetOriginal.jpg}
    \caption{original ResNet \\ \cite{he2016deep}\hphantom{(b)}}\label{fig:PostOrig_2}
  \end{subfigure}
  \begin{subfigure}[b]{0.19\linewidth}
    \includegraphics[width=\linewidth]{ResNet-B.jpg}
    \caption{ResNet-B \cite{gross2016training}\\ \hphantom{(b)} (TorchVision)} \label{fig:PostResNet-B_2}
  \end{subfigure}
  \begin{subfigure}[b]{0.19\linewidth}
    \includegraphics[width=\linewidth]{ResNet-D.jpg}
    \caption{ResNet-D \cite{he2019bag}\\ \hphantom{(b)}} \label{fig:PostResNet-D_2}
  \end{subfigure}
  \begin{subfigure}[b]{0.19\linewidth}
    \includegraphics[width=\linewidth]{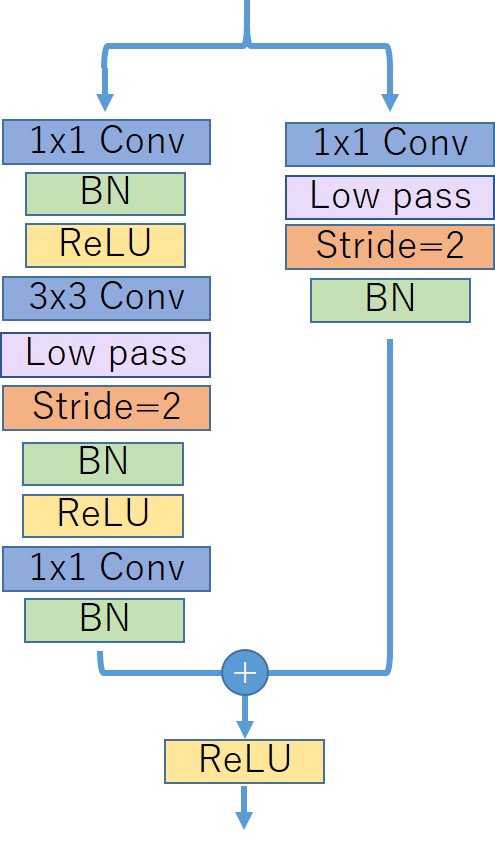}
    \caption{post filter \cite{vasconcelos2021impact}\\ \hphantom{(b)}} \label{fig:PostPostFilter_2}
  \end{subfigure}
  \begin{subfigure}[b]{0.19\linewidth}
    \includegraphics[width=\linewidth]{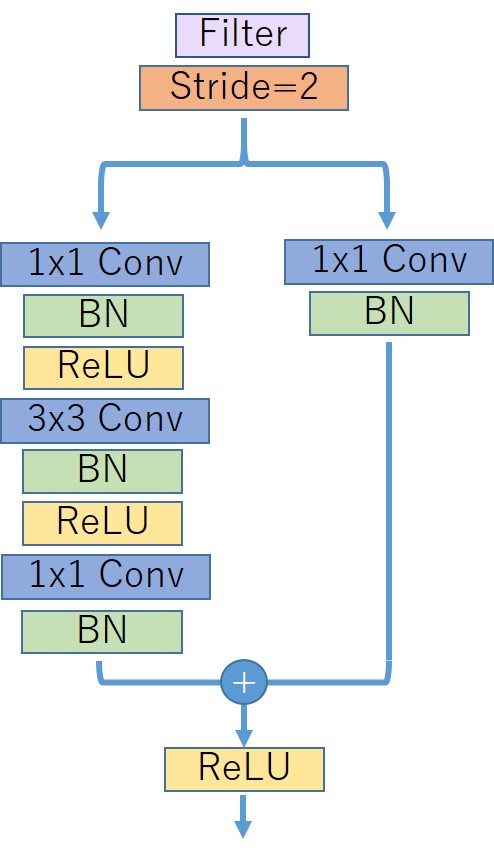}
    \caption{Our filter  \\ \hphantom{(b)}location} \label{fig:PostOurs_2}
  \end{subfigure}
  \hspace*{\fill}%
  \caption{Different modifications of the ResNet regarding downsampling.}
  \label{fig:PostAct_2}
\end{figure*}

There are several versions of ResNets, as shown in \cref{fig:PostAct_2}.   The original ResNet \cite{he2016deep} has subsampling at the beginning of each stage, as shown in \cref{fig:PostOrig_2}.  TorchVision implementation \cite{gross2016training, vryniotis2021how} moved the location of the subsampling within the residual block, as shown in \cref{fig:PostResNet-B_2}.    This implementation is adopted in Tensorflow\footnote{\url{https://github.com/tensorflow/models/blob/master/official/legacy/image_classification/resnet/resnet_model.py}} used by \cite{vasconcelos2021impact}.  It is named ResNet-B in \cite{he2019bag}, which further tweaked ResNet by adding $2 \times 2$ average pooling and named ResNet-D, as shown in \cref{fig:PostResNet-D_2}.  \cite{vasconcelos2021impact} introduced a post-filter before subsampling, as shown in \cref{fig:PostPostFilter_2}, which showed that removing anti-aliasing makes the networks more robust and stable.  We added filters before subsampling in the original ResNet, as shown in \cref{fig:PostOurs_2}.  We compared them using the same STEM from the original ResNet.

\subsubsection{Downsampling filters}
\begin{figure}[t]
\captionsetup[subfigure]{justification=centering}
  \centering
  \begin{subfigure}[1b]{0.15\linewidth}\centering
    \includegraphics[width=\linewidth]{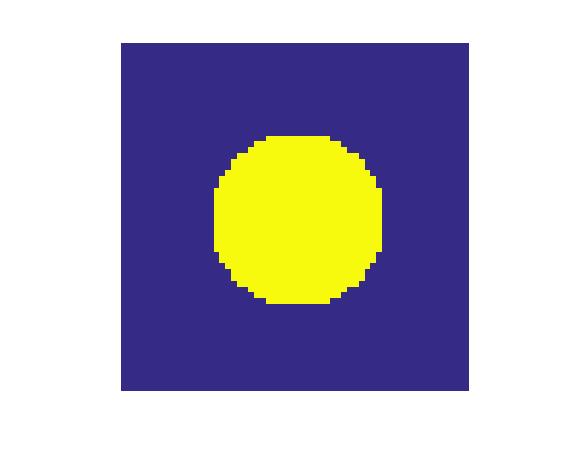}
    \caption{ideal\\ \hphantom{(b)}}\label{fig:FilterIdeal}
  \end{subfigure}
  \begin{subfigure}[1b]{0.15\linewidth}\centering
    \includegraphics[width=\linewidth]{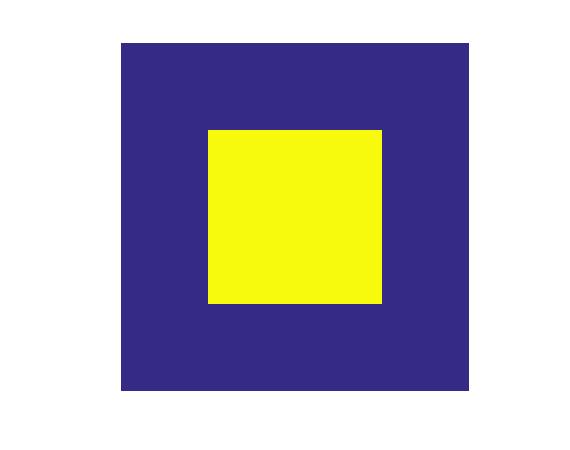}
    \caption{box filter\\ \hphantom{(b)}}\label{fig:box}
  \end{subfigure}
  \begin{subfigure}[2b]{0.15\linewidth}\centering
    \includegraphics[width=\linewidth]{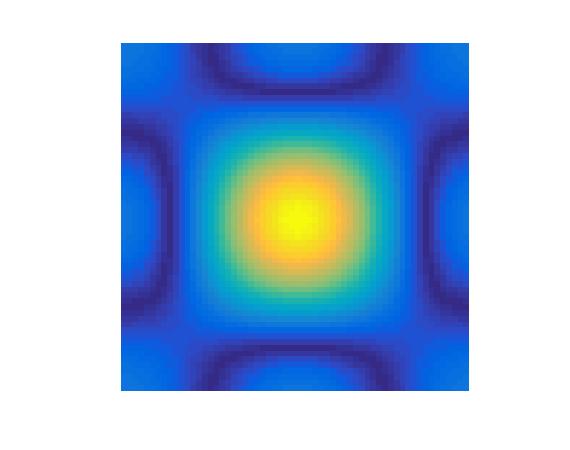}
    \caption{$3 \times 3$\\ \hphantom{(b)} binomial}\label{fig:binomial}
  \end{subfigure}
  \begin{subfigure}[3b]{0.15\linewidth}\centering
    \includegraphics[width=\linewidth]{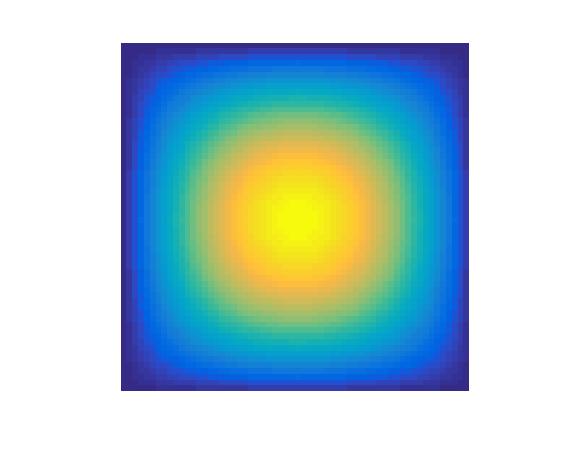}
    \caption{$2 \times 2$\\ \hphantom{(b)}AvgPool}\label{fig:2Avg}
  \end{subfigure}
  \begin{subfigure}[3b]{0.15\linewidth}\centering
    \includegraphics[width=\linewidth]{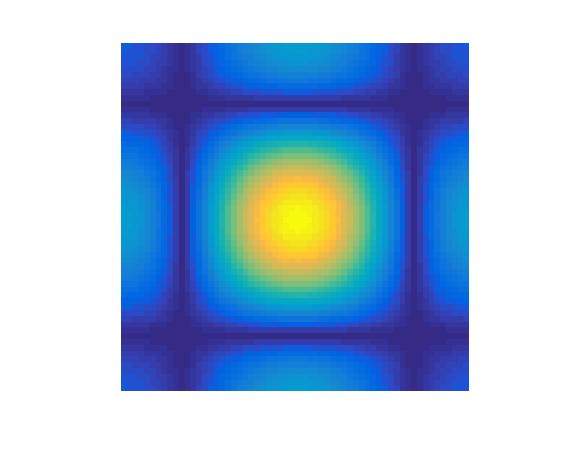}
    \caption{$3 \times 3$\\ \hphantom{(b)}AvgPool}\label{fig:3Avg}
  \end{subfigure}
  \begin{subfigure}[4b]{0.15\linewidth}\centering
    \includegraphics[width=\linewidth]{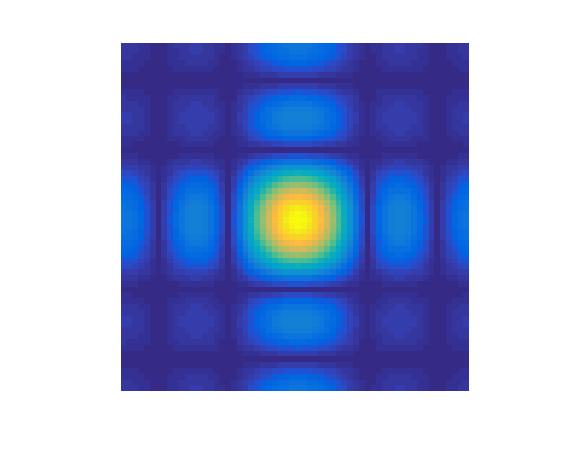}
    \caption{$5 \times 5$\\ \hphantom{(b)} AvgPool}\label{fig:5Avg}
  \end{subfigure}
  \vspace{-0.5em}
  \caption{Different Filters in the frequency domain.}
  
  \label{fig:Filter}
\end{figure}

To avoid aliasing, we should remove the high-pass band from the signal \cite{gonzalez2009digital, szeliski2010computer}.  Since ResNets and many other CNNs use strides  of 2, we should remove half of the signal to satisfy the Nyquist sampling rate.  

\cref{fig:Filter} shows different filters in the frequency domain.
The ideal filter is shown in \cref{fig:FilterIdeal}.  If the input images are not exactly the same size, then creating masks for the ideal filter is difficult; hence we also used box filters, shown in \cref{fig:box}.  These filters are implemented with FFT, so computational complexity is negligible.  However, many embedded systems do not have FFT accelerators.  Also, they have limited memory and data throughput capacity, making it impossible to operate in the frequency domain.  Hence, approximated low-pass filters are used in the spatial domain in classical image processing.  The binomial filter, shown in \cref{fig:binomial}, is a blur filter commonly used in traditional image processing, and average pooling is popularly used in CNNs.  \cref{fig:2Avg}, \cref{fig:3Avg}, and \cref{fig:5Avg} show average pooling filters with different kernel sizes.  

\begin{figure}[t]
  \centering
    \includegraphics[width=\linewidth]{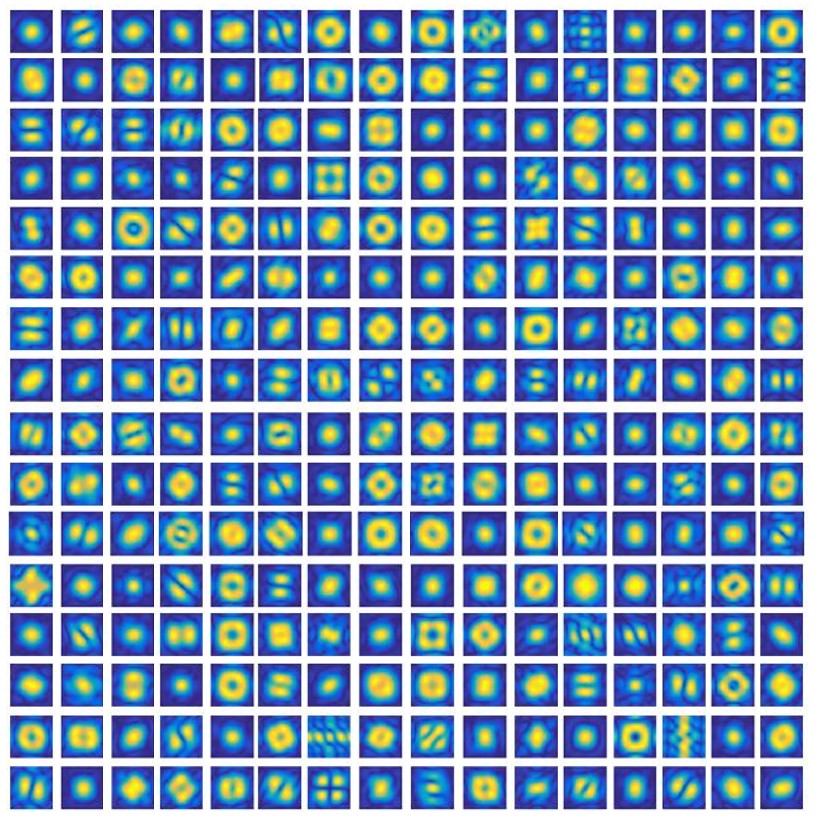}
  \caption{Trained $5 \times 5$ DW convolutional filters before the first downsampled residual block in the frequency domain}
   \vspace{-1em} 
  \label{fig:ConvFilter1}

\end{figure}

We compared the trainable DW convolution with the non-trainable low-pass filters before subsampling.   \cref{fig:ConvFilter1} shows the trained $5 \times 5$ DW convolutions for the first downsampled residual block of ResNet50 in the frequency domain. The trained filters act as feature detectors while removing the high-frequency bands.

\begin{figure}[thb]
  \centering
  \vspace{-1.5em}
  \begin{subfigure}[b]{0.4\linewidth}
    \includegraphics[width=\linewidth]{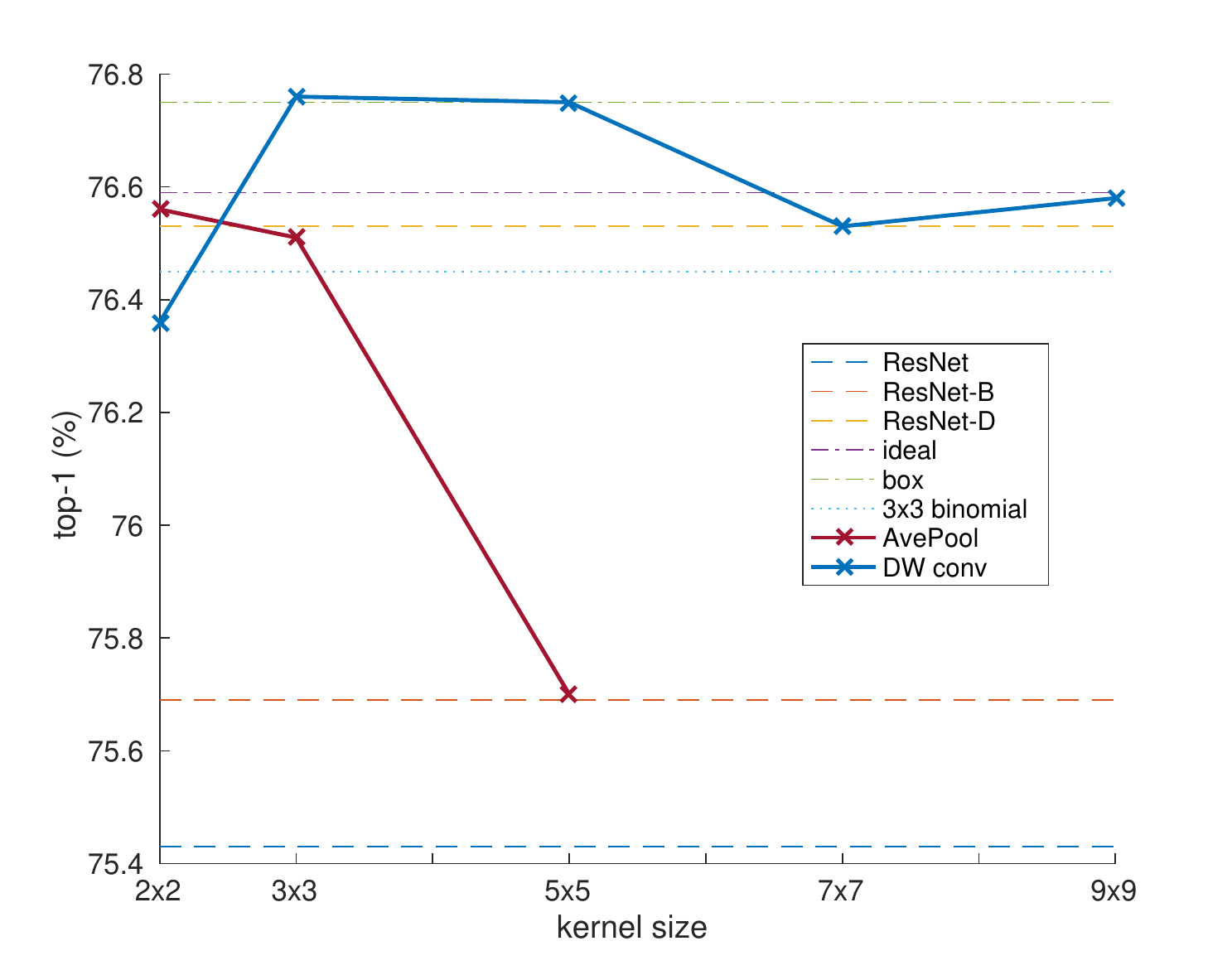}
  \end{subfigure}
  \hspace{-1.5em}
  \begin{subfigure}[b]{0.4\linewidth}
    \includegraphics[width=\linewidth]{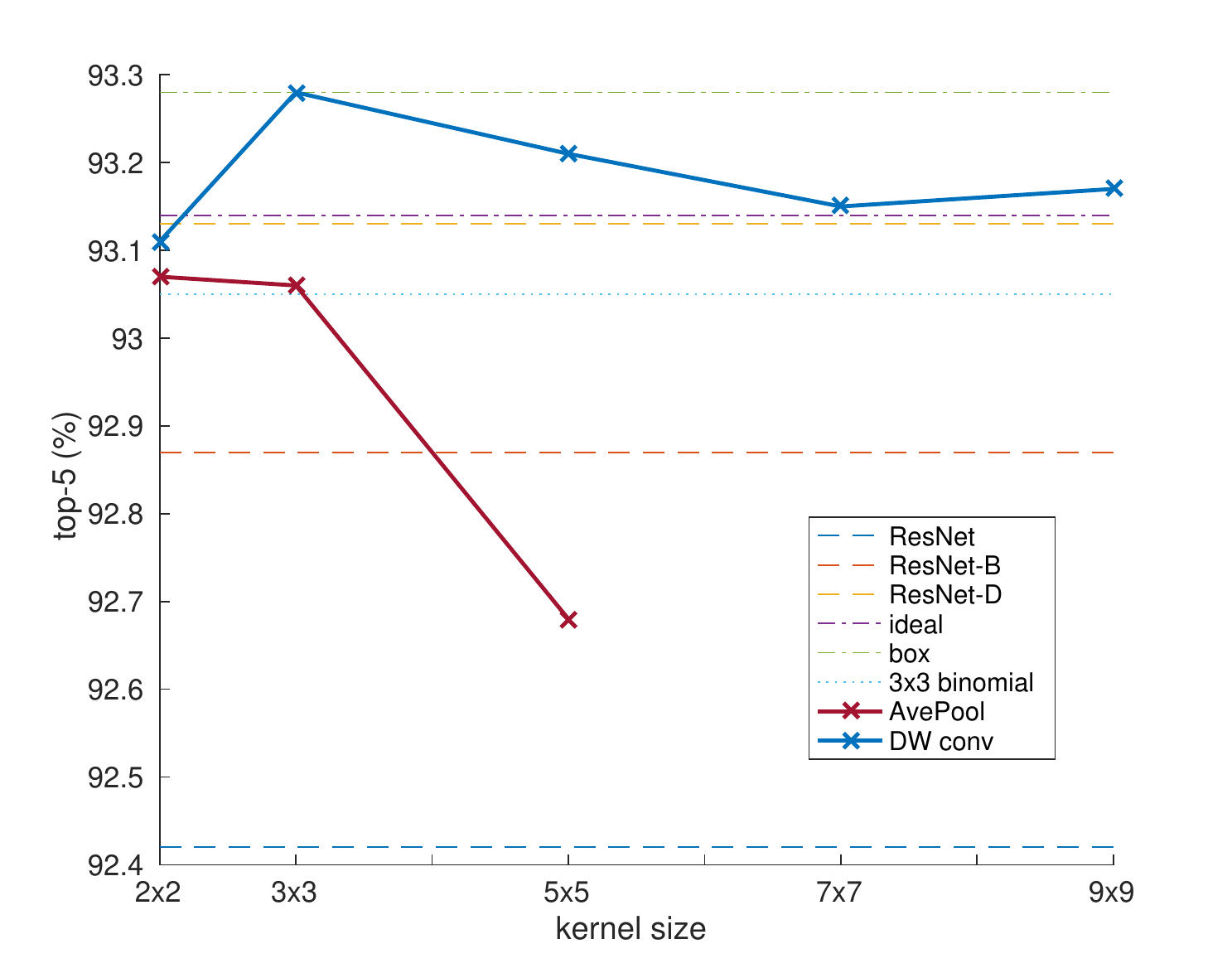}
  \end{subfigure}
  \vspace{-0.5em}
  \caption{Accuracy of the ImageNet classification task depends on the kernel size of the downsampling filters using a classical training scheme.}
  \label{fig:PostAct}
\end{figure}

\subsubsection{Experiments}\label{sec:appendix_exp}
\begin{table}[thb]
  \caption{Accuracy and computational/space complexity of the ImageNet classification task depend on different modifications of ResNet50 using a classical training scheme.}
  \label{tab:post_c}
  \centering
  \begin{tabular}{lllll}
    \toprule
    & Params     &  FLOPs & Top-1 & Top-5\\
    \midrule
    ResNet & 25.6M & 3.9G    & 75.43\% & 92.42\%\\
    ResNet-B & 25.6M & 4.1G & 75.69\% & 92.87\% \\
    ResNet-D & 25.6M & 4.1G & 76.53\% & 93.13\%\\
    post filter & 25.6M & 6.1G & 76.62\% & 93.22\%\\
    \midrule
    \multicolumn{5}{c}{downsampling filters for subsampling in \cref{fig:PostOurs_2}}\\
    \midrule
    ideal & 25.6M & 3.9G & 76.59\% & 93.14\%\\
    box & 25.6M & 3.9G & 76.75\% & 93.28\%\\
    $3\times3$ binomial & 25.6M & 3.9G & 76.45\% & 93.05\%\\
    $3\times3$ Avg & 25.6M & 3.9G & 76.51\% & 93.06\%\\
    $3\times3$ DW Conv & 25.6M & 3.9G & 76.76\% & 93.28\%\\
    
    \bottomrule
  \end{tabular}\label{tab:FLOPs}
\end{table}
We trained the different types of ResNet50 using a classical learning scheme from the default setup of TorchVision.

\cref{fig:PostAct} shows the accuracies over the size of filters before subsampling in \cref{fig:PostOurs_2} using DW convolution and average pooling.  As shown in \cref{fig:5Avg}, $5 \times 5$ average pooling removes too much information, which causes the accuracy decline.  DW convolutional filters are more flexible and can be trained with larger kernels.  However, $3 \times 3$ DW convolution already performs well. 

\cref{tab:FLOPs} lists the accuracies and computational/space complexities.  It clearly shows that the methods with a low-pass filter before subsampling provide better results.  ResNet-B has a low-pass filter, computed through convolution, in only one path; hence it gives accuracy between no low-pass filter and a low-pass filter in both paths.  ResNet-B and ResNet-D are more computationally expensive than the original ResNet because of later subsampling in the networks.  In our method, DW convolution has additional trainable parameters, but compared to the size of the whole network, extra memory usage and FLOPs are negligible.


We also experimented with a modern augmentation and learning scheme from \cite{vryniotis2021how} before any updates; the standard training reference script for TorchVision\footnote{\url{https://github.com/pytorch/vision/tree/main/references/classification}} is given in Listing \ref{lst:modernTraining}.  \cref{Tab:longImageNet} shows similar results as \cref{tab:FLOPs}, where the DW convolutional downsampling filter can provide good accuracy compared to non-trainable low-pass filters.

\begin{table}[thb]
\begin{threeparttable}[htb]
  \caption{Accuracy of the ImageNet classification task depends on the different modifications of ResNet50 using a modern augmentation and training scheme shown in Listing \ref{lst:modernTraining}. }
  \label{Tab:longImageNet}
  \centering
  \begin{tabular}[htp]{l|ll}
    \toprule
    & Top-1\hphantom{(b)} & Top-5\hphantom{(b)} \\
    \midrule
    ResNet & 79.87\% & 94.99\% \\
    ResNet-B\cite{gross2016training, vryniotis2021how}\tnote{*}\hphantom{(b)} & 80.67\% & 95.17\%\\
    \midrule
    \multicolumn{3}{c}{downsampling filters for subsampling in \cref{fig:PostOurs_2}}\\
    \midrule
    ideal & 80.51\% & 95.16\%\\
    box & 80.80\% & 95.42\% \\
    $2\times2$ Avg Pool & 80.05\% & 95.03\% \\
    $3\times3$ Avg Pool & 80.38\% & 95.17\% \\
    $3\times3$ DW Conv & 80.94\% & 95.32\% \\
    
    \bottomrule
  \end{tabular}

\begin{tablenotes}
\item[*] Data is taken from \cite{vryniotis2021how}. Note, official TorchVision implementation is not the original ResNet.
\end{tablenotes}
\end{threeparttable}
\end{table}

\minipage{\linewidth} 
\begin{lstlisting}[language=Python, caption=reference script of the modern training scheme \cite{vryniotis2021how}, escapechar=!, label={lst:modernTraining}]
torchrun --nproc_per_node=8 train.py --model resnet50 --batch-size 128 --lr 0.5 --lr-scheduler cosineannealinglr --lr-warmup-epochs 5 --lr-warmup-method linear --auto-augment ta_wide --epochs 600 --random-erase 0.1 --weight-decay 0.00002 --norm-weight-decay 0.0 --label-smoothing 0.1 --mixup-alpha 0.2 --cutmix-alpha 1.0 --train-crop-size 176 --model-ema --val-resize-size 232
\end{lstlisting}
\endminipage

\subsection{Modern training scheme}\label{sec:modern}

Recently, strong regularizations have provided better results in both traditional CNNs \cite{wightman2021resnet, bello2021revisiting} and SA networks \cite{touvron2020training}.  These methods require long epochs, hence not feasible for our ablation studies.  In this section, we experimented with three CADAsp on ResNet-D with a modern augmentation and learning scheme \cite{vryniotis2021how}, as shown in \cref{tab:Long}.  Listing \ref{code:training} shows the training scheme using the standard training reference script for TorchVision.
Because of cropping in the augmentation, we gave $3 \times 3$ average pooling with zero padding instead of the original $2 \times 2$ average pooling without padding for all networks.  Even with the strong augmentation, CADAsp provided better accuracy than DW convolution and was on par with convolution.  We should note that the augmentations and training schemes are specifically tuned for the ResNet.

\begin{table}[thb]
  \caption{Accuracy of the ImageNet classification task using the modern training method for ResNet on ResNet-D with convolution, DW convolution, and CADAsp using $3\times 3$ CA kernels.  "$C_h$" represents the number of channels in each head, and "$b$" represents the number of base kernels in each stage.\vspace{-0.5cm}}
  \centering
  \setlength{\tabcolsep}{5pt} 
  \begin{tabular}{llll}
    \toprule
    Architecture     &  Top-1   &   Params & FLOPs  \\
    \midrule
    $3\times 3$ conv (ResNet-D)     &  81.17\% & 25.58M & 4.37G     \\
    $7\times 7$ DW conv (no head)  & 80.54\% & 14.44M & 2.58G  \\
    $9\times 9$ CADAsp $b=(8,8,8,8)$, $C_h=16$ & 81.15\% & 14.64M & 2.76G   \\
    $7\times 7$ CADAsp $b=(16,16,16,16)$, $C_h=4$ & 81.42\% & 15.61M & 3.06G   \\
    $9\times 9$ CADAsp $b=(128,128,128,128)$, $C_h=16$ & 81.45\% & 21.10M & 5.12G   \\
    \bottomrule
  \end{tabular}
  \label{tab:Long}
\end{table}

\minipage{\linewidth} 
\begin{lstlisting}[language=Python, caption=reference script of the modern training scheme \cite{vryniotis2021how}, escapechar=!, label={code:training}]
torchrun --nproc_per_node=8 train.py --model resnet50 --batch-size 128 --lr 0.5 --lr-scheduler cosineannealinglr --lr-warmup-epochs 5 --lr-warmup-method linear --auto-augment ta_wide --epochs 600 --random-erase 0.1 --weight-decay 0.00002 --norm-weight-decay 0.0 --label-smoothing 0.1 --mixup-alpha 0.2 --cutmix-alpha 1.0 --train-crop-size 176 --model-ema --val-resize-size 232 --ra-sampler --ra-reps 4
\end{lstlisting}
\endminipage

\subsection{Trained base kernels} \label{sec:kernels}
\begin{figure}[h]
  \centering
   \includegraphics[width=0.6\linewidth]{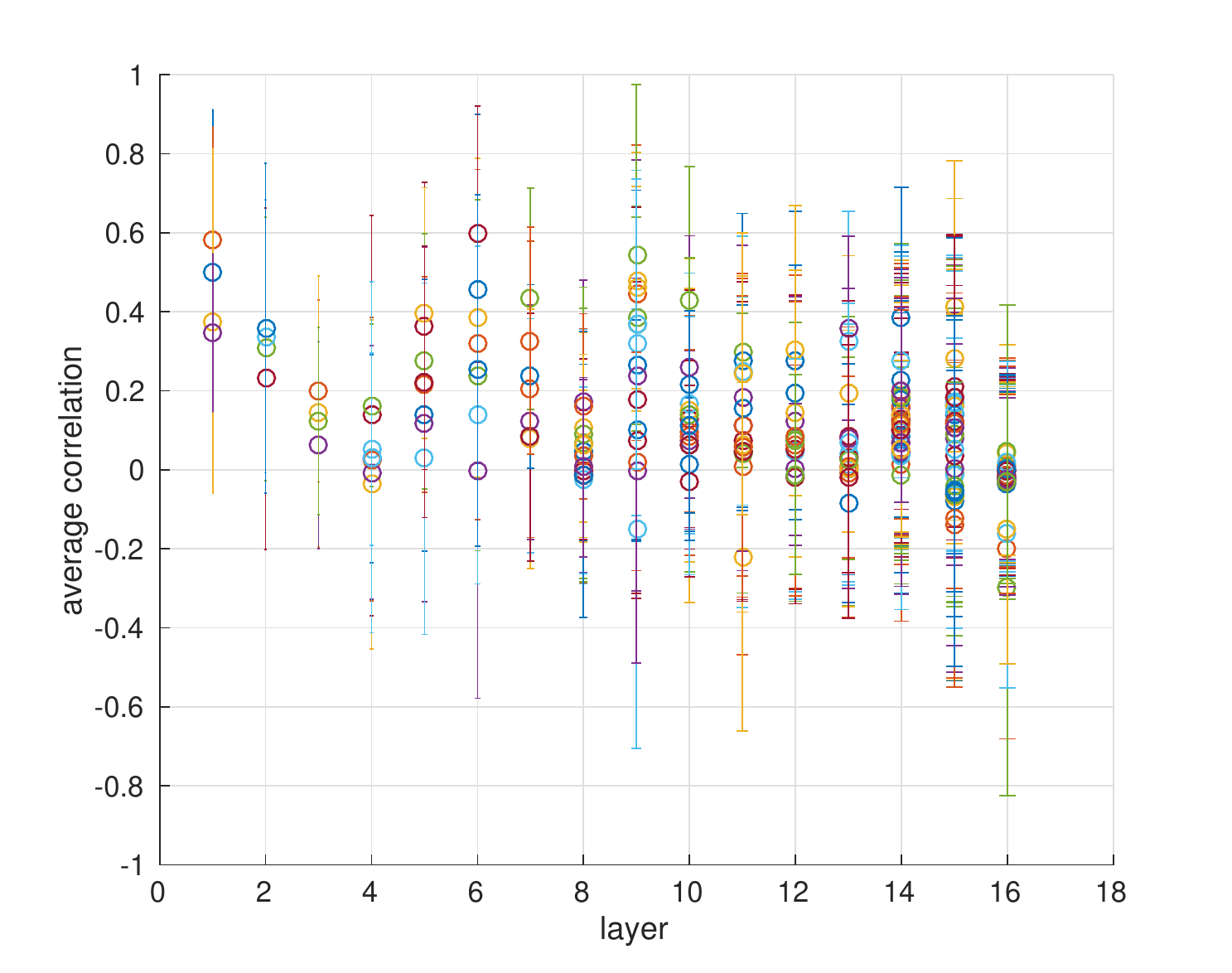}
   \vspace{-0.0cm}
   \caption{Average correlation with base kernels and position encoding in the same head after pruning using CADAsp with 128 base kernels for all layers, $3 \times 3$ CA kernels, $7 \times 7$ aggregation kernels, and 16 channels in each head.  
   }
   \label{fig:KernelCorr2}
\end{figure}

\begin{figure}[t]
  \centering
  \includegraphics[width=1.0\linewidth]{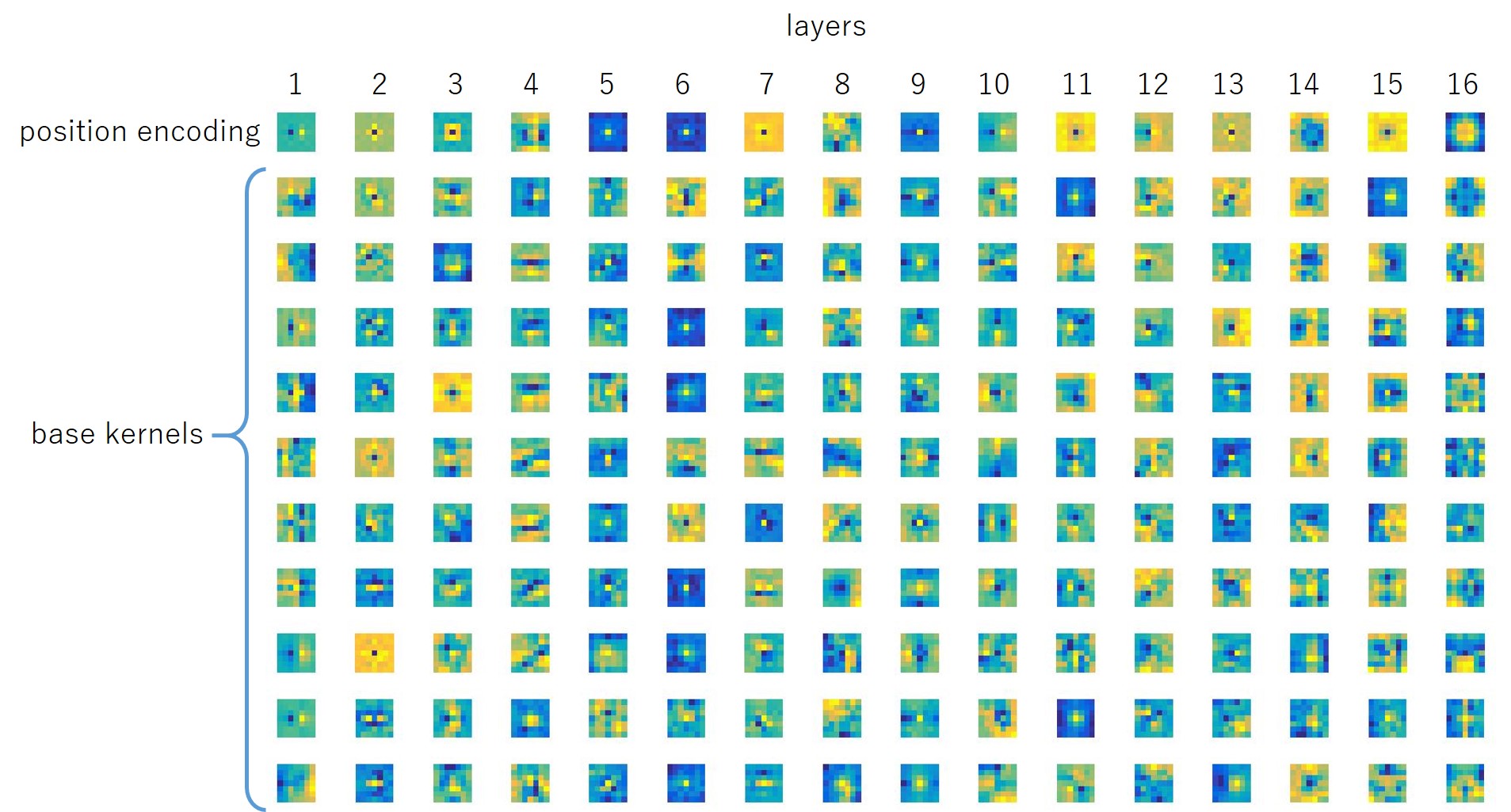}

   \includegraphics[width=1.0\linewidth]{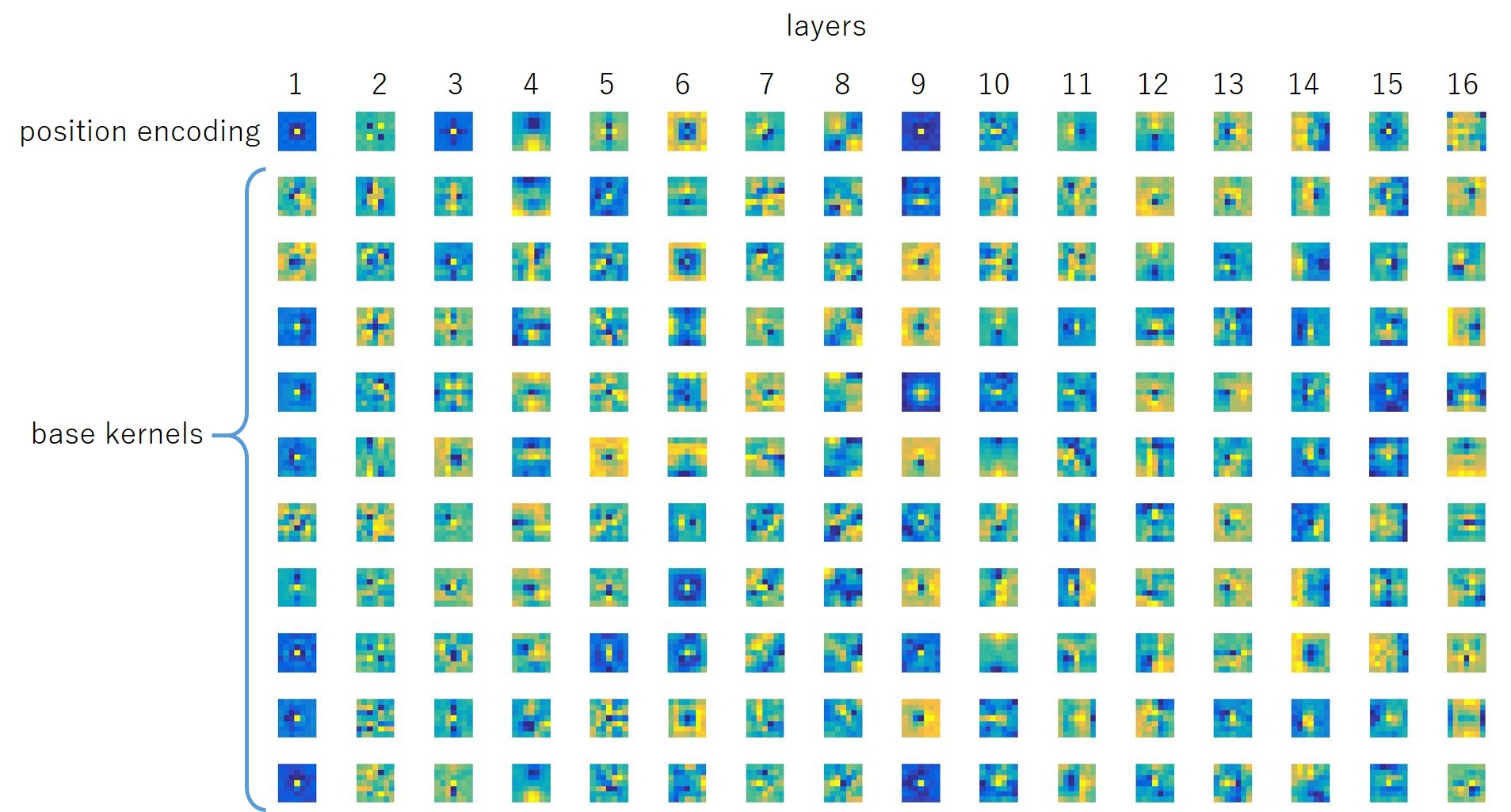}
   \caption{Two sets of trained kernels in the same head after pruning using CADAsp with 128 base kernels for all layers, $3 \times 3$ CA kernels, $7 \times 7$ aggregation kernels, and 16 channels in each head.  Ten randomly chosen base kernels are shown.
   }
   \label{fig:ExampleKernel}
\end{figure}

\begin{figure}[t]
  \centering
  \includegraphics[width=1.0\linewidth]{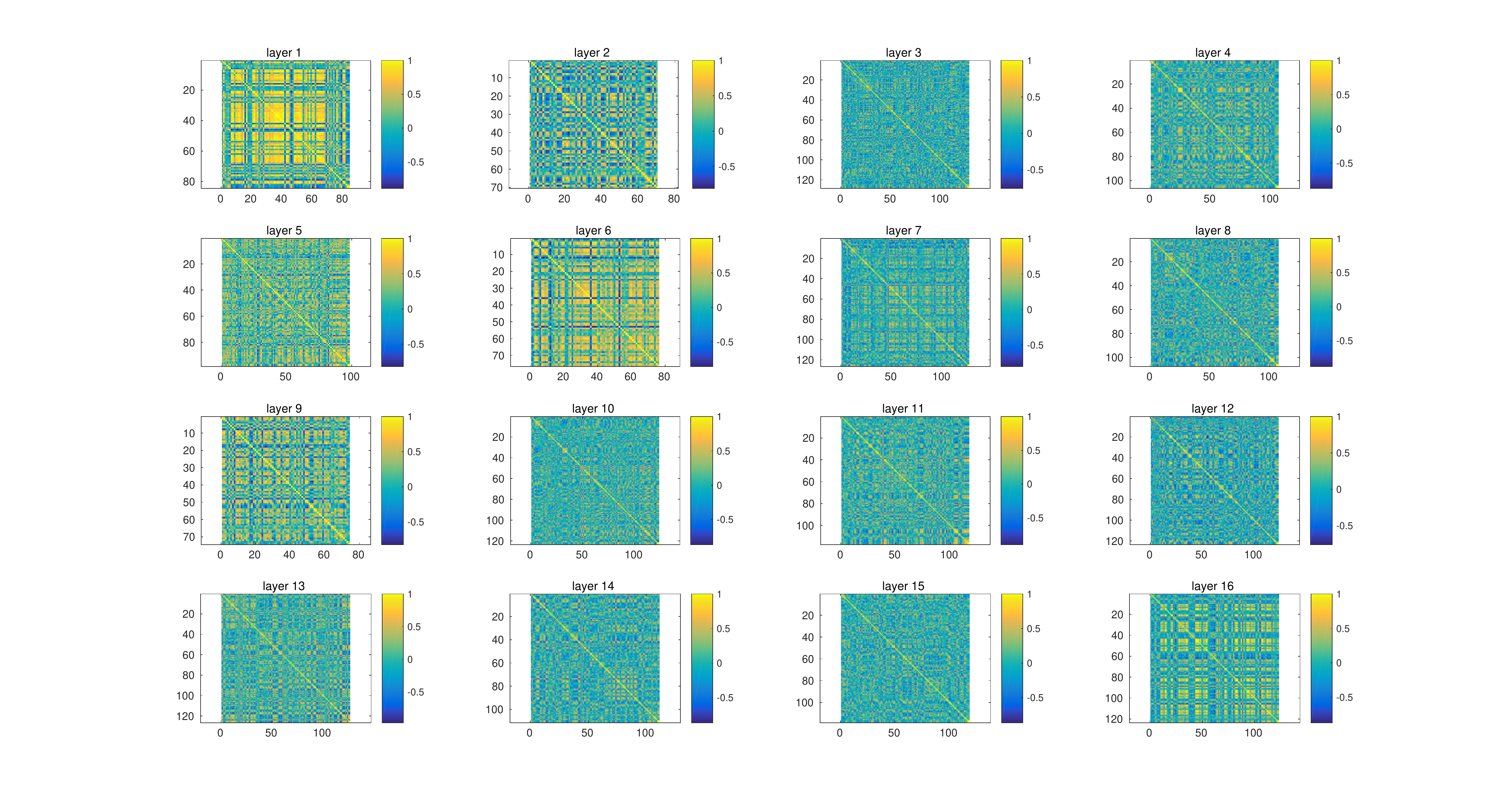}

   \includegraphics[width=1.0\linewidth]{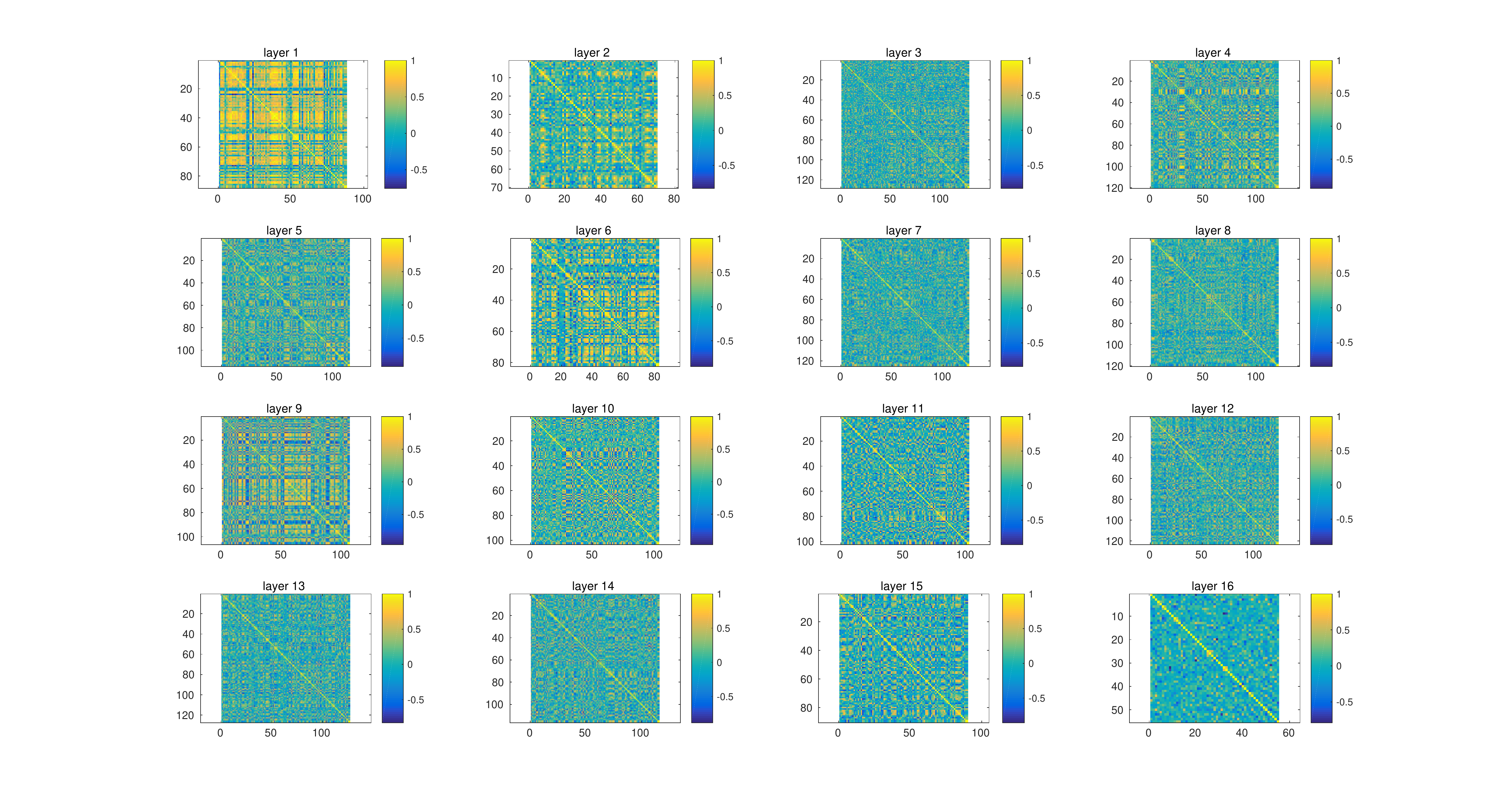}
   \vspace{-0.2cm}
   \caption{Two sets of trained base kernels' correlation in the same head after pruning using CADAsp with 128 base kernels for all layers, $3 \times 3$ CA kernels, $7 \times 7$ aggregation kernels, and 16 channels in each head.  
   }
   \label{fig:KernelCorr}
\end{figure}

In \cref{sec:spatial}, we discussed trained base kernels.  In this section, we give examples of trained kernels  after pruning of CADAsp, using $3 \times 3$ CA kernels, $7 \times 7$ aggregation kernels, 128 base kernels for all layers, and 16 channels in each head ($C_h = 16$).  \cref{fig:ExampleKernel} shows the trained relative position encoding kernels and base kernels in the same head.  Some of the base kernels are correlated, as shown in \cref{fig:KernelCorr}.  Also, some base kernels are correlated with the relative position encoding kernel in the same head, as shown in \cref{fig:KernelCorr2}.


\end{document}